\documentclass{article}

\usepackage[accepted]{configuration/icml2023}
\usepackage[hyperindex,breaklinks,colorlinks,citecolor=darkblue]{hyperref} % allow link break (for arxiv) 

\usepackage{stfloats}
\usepackage[utf8]{inputenc}
\usepackage[T1]{fontenc}
\usepackage{booktabs}
\usepackage{amsfonts}       % blackboard math symbols
\usepackage{nicefrac}       % compact symbols for 1/2, etc.
\usepackage{microtype}      % microtypography
\usepackage{xcolor}
\usepackage[flushleft]{threeparttable}
\usepackage{color}
\usepackage{mathtools}

\usepackage{url}
\usepackage{float}
\usepackage{subfigure}
\usepackage{graphicx}
\usepackage{multirow}
\usepackage{xspace}
\usepackage{natbib}
\usepackage{enumitem}
\usepackage[font=small]{caption}
\usepackage{diagbox}
\usepackage{wrapfig}
\usepackage[toc, page, header]{appendix}
\usepackage{cleveref}
\crefformat{section}{\S#2#1#3} % see manual of cleveref, section 8.2.1
\crefformat{subsection}{\S#2#1#3}
\crefformat{subsubsection}{\S#2#1#3}

\usepackage{amssymb}% http://ctan.org/pkg/amssymb
\usepackage{pifont}% http://ctan.org/pkg/pifont
\newcommand{\cmark}{\ding{51}}%
\newcommand{\xmark}{\ding{55}}%

\usepackage[most]{tcolorbox}
\newtcbtheorem{Summary}{}{enhanced,
  coltitle=black,
  top=0.14in,
  attach boxed title to top left=
  {xshift=0em,yshift=-\tcboxedtitleheight/2},
  boxed title style={size=small,colback=blue!10}
}{summary}

\usepackage{tikz}
\newcommand*\circled[1]{\tikz[baseline=(char.base)]{\node[shape=circle,draw,inner sep=0pt,minimum size=9.pt] (char) {#1};}}

\usepackage{amsmath,amsthm,amssymb}

\DeclareMathOperator*{\argmax}{arg\,max}

\renewcommand{\aa}{\mathbf{a}}

\providecommand{\xx}{\mathbf{x}}

\providecommand{\mtheta}{\boldsymbol{\theta}}

\providecommand{\cD}{\mathcal{D}}

\providecommand{\cF}{\mathcal{F}}
\providecommand{\cG}{\mathcal{G}}

\providecommand{\cJ}{\mathcal{J}}

\providecommand{\cP}{\mathcal{P}}

\providecommand{\cS}{\mathcal{S}}
\providecommand{\cT}{\mathcal{T}}

\providecommand{\cX}{\mathcal{X}}
\providecommand{\cY}{\mathcal{Y}}

\newenvironment{talign*}
{\csname align*\endcsname}
{\endalign}

\usepackage{xspace}
 
\newcommand{\resetmodel}{episodic\xspace} 
 
\newcommand{\notresetmodel}{online\xspace} 

\usepackage{algorithm}
\usepackage[noend]{algpseudocode}

\errorcontextlines\maxdimen

\makeatletter
\newcommand*{\algrule}[1][\algorithmicindent]{\makebox[#1][l]{\hspace*{.5em}\thealgruleextra\vrule height \thealgruleheight depth \thealgruledepth}}%
\newcommand*{\thealgruleextra}{}
\newcommand*{\thealgruleheight}{.75\baselineskip}
\newcommand*{\thealgruledepth}{.25\baselineskip}

\newcount\ALG@printindent@tempcnta
\def\ALG@printindent{%
	\ifnum \theALG@nested>0% is there anything to print
	\ifx\ALG@text\ALG@x@notext% is this an end group without any text?
	\else
		\unskip
		\addvspace{-1pt}% FUDGE to make the rules line up
		\ALG@printindent@tempcnta=1
		\loop
		\algrule[\csname ALG@ind@\the\ALG@printindent@tempcnta\endcsname]%
		\advance \ALG@printindent@tempcnta 1
		\ifnum \ALG@printindent@tempcnta<\numexpr\theALG@nested+1\relax% can't do <=, so add one to RHS and use < instead
			\repeat
		\fi
	\fi
}%
\usepackage{etoolbox}
\patchcmd{\ALG@doentity}{\noindent\hskip\ALG@tlm}{\ALG@printindent}{}{\errmessage{failed to patch}}
\makeatother

\newbox\statebox
\newcommand{\myState}[1]{%
	\setbox\statebox=\vbox{#1}%
	\edef\thealgruleheight{\dimexpr \the\ht\statebox+1pt\relax}%
	\edef\thealgruledepth{\dimexpr \the\dp\statebox+1pt\relax}%
	\ifdim\thealgruleheight<.75\baselineskip
		\def\thealgruleheight{\dimexpr .75\baselineskip+1pt\relax}%
	\fi
	\ifdim\thealgruledepth<.25\baselineskip
		\def\thealgruledepth{\dimexpr .25\baselineskip+1pt\relax}%
	\fi
	\State #1%
	\def\thealgruleheight{\dimexpr .75\baselineskip+1pt\relax}%
	\def\thealgruledepth{\dimexpr .25\baselineskip+1pt\relax}%
}
\usepackage[textwidth=5cm]{todonotes}

\usepackage{xspace}  
\usepackage{bm}

\definecolor{darkblue}{rgb}{0.0, 0.0, 0.55}
\definecolor{myblue}{rgb}{0,0.45,0.74}
\definecolor{myred}{rgb}{0.85,0.33,0.1}

\icmltitlerunning{On Pitfalls of Test-Time Adaptation}

\begin{document}

\twocolumn[\icmltitle{On Pitfalls of Test-Time Adaptation}


	\icmlsetsymbol{equal}{*}

	\begin{icmlauthorlist}
		\icmlauthor{Hao Zhao}{EPFL,equal}
		\icmlauthor{Yuejiang Liu}{EPFL,equal}
		\icmlauthor{Alexandre Alahi}{EPFL}
		\icmlauthor{Tao Lin}{WestlakeRCIF,WestlakeSoE}
	\end{icmlauthorlist}

	\icmlaffiliation{WestlakeRCIF}{Research Center for Industries of the Future, Westlake University}
	\icmlaffiliation{WestlakeSoE}{School of Engineering, Westlake University}
	\icmlaffiliation{EPFL}{École Polytechnique Fédérale de Lausanne (EPFL)}
	\icmlcorrespondingauthor{Tao Lin}{lintao@westlake.edu.cn}

	\icmlkeywords{Machine Learning, ICML}

	\vskip 0.3in]



\printAffiliationsAndNotice{\icmlEqualContribution} % otherwise use the standard text.

% \setlength{\parskip}{1.5mm}  %save some paragraph spacing

% spacing
\setlength{\abovecaptionskip}{6pt plus 2pt minus 3pt}
\setlength{\belowcaptionskip}{6pt plus 2pt minus 3pt}
\captionsetup{belowskip=2.0pt}

\makeatletter
\renewcommand{\paragraph}{%
	\@startsection{paragraph}{4}%
	{\z@}{1.0ex \@plus 1.0ex \@minus .2ex}{-1em}%
	{\normalfont\normalsize\bfseries}%
}
\makeatother

% !TeX root = icml2023_ttab.tex

\begin{abstract}
	Test-Time Adaptation (TTA) has recently emerged as a promising approach for tackling the robustness challenge under distribution shifts.
	However, the lack of consistent settings and systematic studies in prior literature hinders thorough assessments of existing methods.
	To address this issue, we present TTAB, a test-time adaptation benchmark that encompasses ten state-of-the-art algorithms, a diverse array of distribution shifts, and two evaluation protocols.
	Through extensive experiments, our benchmark reveals three common pitfalls in prior efforts. First, selecting appropriate hyper-parameters, especially for model selection, is exceedingly difficult due to online batch dependency. Second, the effectiveness of TTA varies greatly depending on the quality and properties of the model being adapted. Third, even under optimal algorithmic conditions, none of the existing methods are capable of addressing all common types of distribution shifts.
	Our findings underscore the need for future research in the field to conduct rigorous evaluations on a broader set of models and shifts, and to re-examine the assumptions behind the empirical success of TTA.
	Our code is available at \url{https://github.com/lins-lab/ttab}.
\end{abstract}

\setlength{\parskip}{1.5pt plus1pt minus0pt}

\begingroup
\setlength{\parskip}{1.5pt plus1pt minus0pt}

\section{Introduction} \label{sec:introduction}
Tackling the robustness issue under distribution shifts is one of the most pressing challenges in machine learning~\citep{koh2021wilds}.
Among existing approaches, Test-Time Adaptation (TTA)---in which neural network models are adapted to new distributions by making use of unlabeled examples at test time---has emerged as a promising paradigm of growing popularity~\citep{lee2022confidence,kundu2022balancing,gong2022note,chen2022contrastive,goyal2022test,sinha2023test}.
Compared to other approaches, TTA offers two key advantages:
(i) {\it generality}: TTA does not rest on strong assumptions regarding the structures of distribution shifts, which is often the case with Domain Generalization (DG) methods~\citep{gulrajani2021in};
(ii) {\it flexibility}: TTA does not require the co-existence of training and test data, a prerequisite of the Domain Adaptation (DA) approach~\citep{ganin2015unsupervised}.
At the core of TTA is to define a proxy objective used at test time to adapt the model in an unsupervised manner.
Recent works have proposed a broad array of proxy objectives, ranging from entropy minimization~\citep{wang2021tent} and self-supervised learning~\citep{sun2020test} to pseudo-labeling~\citep{liang2020we} and feature alignment~\citep{liu2021ttt++}.
Nevertheless, the efficacy of TTA in practice is often called into question due to restricted and inconsistent experimental conditions in prior literature~\citep{boudiaf2022parameter,su2022revisiting}.
% \looseness=-1

The goal of this work is to gain a thorough understanding of the current state of TTA methods while setting the stage for critical problems to be worked on.
To this end, we present TTAB, an open-sourced \underline{T}est-\underline{T}ime \underline{A}daptation \underline{B}enchmark featuring rigorous evaluations, comprehensive analyses as well as extensive baselines.
Our benchmark carefully examines ten state-of-the-art TTA algorithms on a wide range of distribution shifts using two evaluation protocols.
Specifically, we place a strong emphasis on subtle yet crucial experimental settings that have been largely overlooked in previous works.
Our analyses unveil three common pitfalls in prior TTA methods:

\emph{Pitfall 1}: Hyperparameters have a strong influence on the effectiveness of TTA, and yet they are exceedingly difficult to choose in practice without prior knowledge of distribution shifts. Our results show that the common practice of hyperparameter choice for TTA methods does not necessarily improve test accuracy and may instead lead to detrimental effects.
Moreover, we find that even given the labels of test examples, selecting TTA hyperparameters remains challenging, primarily due to the batch dependency that arises during online adaptation.

\emph{Pitfall 2}: The effectiveness of TTA may vary greatly across different models.
In particular, not only the model accuracy in the source domain but also its feature properties have a strong influence on the result post-adaptation.
Crucially, we find that good practice in data augmentations~\citep{hendrycks2019augmix,hendrycks2022robustness} for out-of-distribution generalization leads to adverse effects for TTA.

\emph{Pitfall 3}: Even under ideal conditions where optimal hyperparameters are used in conjunction with suitable pre-trained models, existing methods still perform poorly on certain families of distribution shifts, such as correlation shifts~\citep{sagawa2019distributionally} and label shifts~\citep{sun2022beyond}), which are infrequently considered in the realm of TTA but widely used in domain adaptation and domain generalization. This observation, together with the previously mentioned issues, raises questions about the potential of TTA in addressing unconstrained distribution shifts in nature that are beyond our control.

Aside from these empirical results, our TTAB benchmark is designed as an expandable package that standardizes experimental settings and eases the integration of new algorithmic implementations.
We hope our benchmark library will not only facilitate rigorous evaluations of TTA algorithms across a broader range of base models and distribution shifts, but also stimulate further research into the assumptions that underpin the viability of TTA in challenging scenarios.

\endgroup

\section{Related Work} \label{sec:related_work}

Early methods of test-time adaptation involve updating the statistics and/or parameters associated with the batch normalization layers~\citep{schneider2020improving,wang2021tent}.
This approach has shown promising results in mitigating image corruptions~\citep{hendrycks2018benchmarking}, but its efficacy is often limited to a narrow set of distribution shifts due to the restricted adaptation capacity~\citep{burns2021limitations}.
To effectively update more parameters, e.g., the whole feature extractor, using unlabeled test examples, prior works have explored a wide array of proxy objectives.

One line of works designs TTA objectives by exploiting common properties of classification problems, e.g., entropy minimization~\citep{liang2020we,fleuret2021uncertainty,zhou2021bayesian}, class prototypes~\citep{li2020model,su2022revisiting,yang2022attracting}, pseudo labels~\citep{rusak2022if,li2021free}, and invariance to augmentation~\citep{zhang2022memo,kundu2022balancing}.
These techniques are restricted to the cross-entropy loss of the main tasks, and hence inherently inapplicable to regression problems, e.g., pose estimation~\citep{li2021test}.

Another line of research seeks more general proxies through self-supervised learning, e.g., rotation prediction~\citep{sun2020test}, contrastive learning~\citep{liu2021ttt++,chen2022contrastive}, and masked auto-encoder~\citep{gandelsmantest}.
While these methods are task-generic, they typically require modifications of the training process to accommodate an auxiliary self-supervised task, which can be non-trivial.

Some recent works draw inspiration from related areas for robust test-time adaptation, such as feature alignment~\citep{liu2021ttt++,eastwoodsource,jiang2023test}, style transfer~\citep{gao2022back}, and meta-learning~\citep{zhang2021adaptive}.
Unfortunately, the absence of standardized experimental settings in the previous literature has made it difficult to compare existing methods.
Instead of introducing yet another new method, our work revisits the limitations of prior methods through a large-scale empirical benchmark.

Closely related to ours, \citet{boudiaf2022parameter} has recently shown that hyperparameters of TTA methods often need to be adjusted depending on the specific test scenario.
Our results corroborate their observations and go one step further by taking an in-depth analysis of the online TTA setting.
Our findings not only shed light on the challenge of model selection arising from batch dependency but also identify other prevalent pitfalls associated with the quality of pre-train models and the variety of distribution shifts.

\section{TTA Settings and Benchmark} \label{sec:eval_limits_in_tta}

Despite the growing number of TTA methods summarized in~\cref{sec:related_work}, their strengths and limitations are not well understood yet due to the lack of systematic and consistent evaluations.
In this section, we will first revisit the concrete settings of prior efforts, highlighting a few factors that vary greatly across different methods.
We will then propose an open-source TTA benchmark, with a particular emphasis on three aspects: standardization of hyper-parameter tuning, quality of pre-trained models, and variety of distribution shifts.

% In this section, we will first revisit the issue arising from improper evaluation in prior work, highlighting the difficulty of model selection.
% We will subsequently propose evaluation protocols to estimate the optimal performance of TTA methods, where we surprisingly find it is non-trivial:
% 1) the oracle model selection does not indicate an optimal TTA performance due to the batch dependency\hao{As per request of reviewer Dyj6, we should elaborate more on the concept of batch dependency.} (see~\cref{sec:fair_evaluation}),
% and 2) even if hyperparameters are optimally selected, the effectiveness of TTA methods varies based on pre-trained model qualities (see~\cref{sec:model_quality}).

% There have been a number of approaches proposed to bridge the distribution shift gap between the source and target domains, and they showcased the comparison results across different methods in their work.
% But these comparison results may be biased and inaccurate due to using inconsistent adaptation setups.
% Additionally, a suitable model selection technique, which can lessen the detrimental effects of hyperparameter sensitivity is still absent in TTA.
% In this section, we sum up the limitations of evaluation for TTA methods and offer a solution for proper evaluation and fair comparison of TTA methods.

\subsection{Preliminary} \label{sec:preliminary}

Let $\cD_{\cS} = \{ \cX_{\cS}, \cY_{\cS} \}$ be the data from the source domain $\cS$ and $\cD_{\cT} = \{ \cX_{\cT}, \cY_{\cT} \}$ be the data from the target domain $\cT$ to adapt to.
Each sample and the corresponding true label pair $(\xx_i, y_i) \in \cX_{\cS} \times \cY_{\cS}$ in the source domain follows a probability distribution $P_{\cS}(\xx, y)$.
Similarly, each test sample from the target domain and the corresponding label at test time $t$, $(\xx^{(t)}, y^{(t)}) \in \cX_{\cT} \times \cY_{\cT}$, follows a probability distribution $P_{\cT}(\xx, y)$ where $y^{(t)}$ is unknown for the learner.
$f_{\mtheta^o}(\cdot)$ is a base model trained on labeled training data $\{(\xx_i,y_i)\}_{i=1}^{N}$, where $\mtheta^o$ denotes the base model parameters.
During the inference time, the pre-trained base model may suffer from a substantial performance drop in the face of out-of-distribution test samples, namely $\xx \sim P_{\cT}\left( \xx \right)$, where $P_{\cT}\left( \xx \right) \neq P_{\cS} \left( \xx \right)$.
Unlike traditional DA that uses $\cD_{\cS}$ and $\cX_{\cT}$ collected beforehand for adaptation, TTA adapts the pre-trained model $f_{\mtheta^o}(\cdot)$ from $\cD_{\cS}$ on the fly by utilizing unlabeled sample $\xx^{(t)}$ obtained at test time $t$.

\begin{table*}[!t]
	\centering
	\caption{\small
		\textbf{Comparison of experimental settings used in prior TTA methods.}
		% We only list some key factors due to the space issue; 
		The inconsistent settings of hyperparameter tuning~(\cref{sec:hyperparam_tuning}), pre-trained models~(\cref{sec:model_quality}), and distribution shifts~(\cref{sec:shift_type}) may yield different observations.
		More details are summarized in~\cref{appendix:messages}.
	}
	\label{tab:unfair_comparison}
	\resizebox{1.\textwidth}{!}{%
		\begin{tabular}{lcccccr}
			\toprule
			\textbf{Methods}                         & \textbf{Venue} & \textbf{Nb. Hyperparameters} & \textbf{Reset Model} & \textbf{Batch-Norm} & \textbf{Adjust Pre-training} & \textbf{Distribution Shifts}      \\ \midrule
			TTT~\citep{sun2020test}                  & ICML 2020      & 6                            & \xmark               & \xmark              & \cmark                       & co-var. \& non-stat. \& natural shifts  \\
			SHOT~\citep{liang2020we}                 & ICML 2020      & 6                            & \xmark               & \xmark              & \xmark                       & domain gen. shifts           \\
			BN\_Adapt~\citep{schneider2020improving} & NeurIPS 2020   & 1                            & \xmark               & \cmark              & \xmark                       & co-var. \& natural shifts    \\
			TENT~\citep{wang2021tent}                & ICLR 2021      & 2                            & \xmark               & \cmark              & \xmark                       & co-var. \& domain gen. shifts\\
			TTT++~\citep{liu2021ttt++}               & NeurIPS 2021   & 6                            & \xmark               & \xmark              & \cmark                       & co-var. \& domain gen. \& natural shifts\\
			T3A~\citep{iwasawa2021test}              & NeurIPS 2021   & 1                            & \xmark               & \xmark              & \xmark                       & domain gen. shifts           \\
			EATA~\citep{niu2022efficient}            & ICML 2022      & 6                            & \xmark               & \cmark              & \xmark                       & co-var. \& non-stat. shifts  \\
			Conjugate PL~\citep{goyal2022test}       & NeurIPS 2022   & 3                            & \xmark               & \cmark              & \xmark                       & co-var. \& domain gen. shifts\\
			MEMO~\citep{zhang2022memo}               & NeurIPS 2022   & 4                            & \cmark               & \xmark              & \xmark                       & co-var. \& natural shifts    \\
			NOTE~\citep{gong2022note}                & NeurIPS 2022   & 6                            & \xmark               & \cmark              & \cmark                       & co-var. \& non-stat. shifts  \\
			SAR~\citep{niu2023towards}               & ICLR 2023      & 4                            & \cmark               & \xmark              & \xmark                       & co-var. \& label shifts      \\
			\bottomrule
		\end{tabular}
	}
	\label{tab:setups}
	\vspace{-10pt}
\end{table*}

% List of significant hyperparameters used in benchmarked TTA methods:
% TTT: lr, num. of steps, the entry point of the auxiliary head, the number of augmentation ops per image, confidence threshold, and rotation type
% SHOT: lr, num. of steps, confidence threshold, ent_par, cls_par, num. of epochs for offline adaptation
% BN_Adapt: adapt_prior (the ratio of training set statistics.)
% TENT: lr, num. of steps
% TTT++: lr, num. of steps, batch size for alignment, queue size, scale_ext, scale_ssh
% T3A: top_M (select top M largest entropy of the support set)
% EATA: lr, num. of steps, threshold for reliable minimization, threshold for filtering redundant samples, fisher_size, fisher_alpha
% Conjugate PL: lr, num. of steps, temperature_scaling
% MEMO: lr, num. of steps, the number of augmentation ops per image, adapt_prior
% NOTE: lr, num. of steps, memory size, number of target samples used for every update, iabn_k, confidence threshold
% SAR: lr, num. of steps, the threshold for reliable minimization, threshold e_m for model recovery scheme, 

% \todo{update table content, e.g., new colums}

\subsection{Inconsistent Settings in Prior Work} \label{sec:improper_eval}

To gain a comprehensive understanding of the experimental settings used in previous studies, we outline in~\autoref{tab:unfair_comparison} some key factors that characterize the adaptation procedure.
We observe that, despite a restricted selection of factors, existing TTA methods still exhibit substantial variation in the following three aspects:

\paragraph{Hyperparameter.}
TTA methods typically require the specification of hyperparameters such as the learning rate, the number of adaptation steps, as well as other method-specific choices.
However, prior research often lacks detailed discussions on how these hyperparameters were tuned.
In fact, there is no consensus on even simple hyperparameters, such as whether to reset the model during adaptation.
Some TTA methods are \emph{\resetmodel}, performing adaptations on the base model $\mtheta^o$ for every adaptation step. Conversely, some other TTA methods adapt models $\mtheta^\star$ in an \emph{\notresetmodel} manner, leading to stronger dependency across batches and thereby further amplifying the importance of hyperparameter tuning, which we will elaborate in~\cref{sec:hyperparam_tuning}.

\paragraph{Pre-trained Model.}
The choice of pre-trained models constitutes another prominent source of inconsistency in prior research.
Earlier TTA methods often hinge on models with BatchNorm (BN) layers, while more recent ones start to incorporate modern architectures, such as GroupNorm (GN) layers and Vision Transformers.
Besides model architectures, the pre-training procedure in the source domain also varies significantly due to the use of auxiliary training objectives and data augmentation techniques, among other factors.
These variations not only affect the capacity and quality of the pre-trained model, but may also lead to different efficacies of TTA methods, as discussed in~\cref{sec:model_quality}.

\paragraph{Distribution Shift.}
The most compelling property of TTA is, arguably, its potential to handle various distribution shifts depending on the encountered test examples.
However, prior work often considers a narrow selection of distribution shifts biased toward the designed method.
For instance, some methods~\citep{iwasawa2021test} undergo extensive evaluations on domain generalization benchmarks, while a few others~\citep{sun2020test,wang2021tent} concentrate more on image corruption.
As such, the efficacy of existing TTA methods under a wide spectrum of distribution shifts remains contentious, which we will further investigate in~\cref{sec:shift_type}

\subsection{Our Proposed TTA Benchmark}

In order to address the aforementioned inconsistencies and unify the evaluation of TTA methods, we present an open-source Test-Time Adaptation Benchmark, dubbed TTAB.
Our TTAB features standardized experimental settings, extensive baseline methods as well as comprehensive evaluation protocols that enable rigorous comparisons of different methods.

\begin{figure}[!t]
	\subfigure[\small generic formulation]{
		\label{fig:test_time_distribution_shift_formulation_overall}
		\resizebox{!}{0.1\textheight}{%
			\begin{tikzpicture}[x=1cm, y=1cm, z=-0.6cm]
				\draw [->] (0,0,0) -- (2,0,0) node [at end, right] {time};
				\draw [->] (0,0,0) -- (0,2,0) node [at end, left] {sampling};
				\draw [->] (0,0,0) -- (0,0,2) node [at end, left] {$\cP(\aa^{1: K})$};
			\end{tikzpicture}%
		}
	}
	\hfill
	\subfigure[\small $\cP(\aa^{1:K})$ example]{
		\label{fig:test_time_distribution_shift_formulation_attribute_distribution}
		\resizebox{!}{0.1\textheight}{%
			\begin{tikzpicture}[x=1cm, y=1cm, z=-0.6cm]
				\draw [->] (0,0,0) -- (2,0,0) node [at end, right, text width=2.5cm] {an unseen \\ style attribute $a^2$};
				\draw [->] (0,0,0) -- (0,2,0) node [at end, left, text width=2.5cm] {a seen \\ style attribute $a^1$};
				\draw [->] (0,0,0) -- (0,0,2) node [at end, left] {label $a^3$};
			\end{tikzpicture}%
		}
	}
	\caption{
		\small A generic formulation of distribution shifts, where $\cP(a^{1:K})$ is characterized by some attributes, for instance, two image styles and one target label.
	}
	\label{fig:test_time_distribution_shift_formulation}
	\vspace{-10pt}
\end{figure}
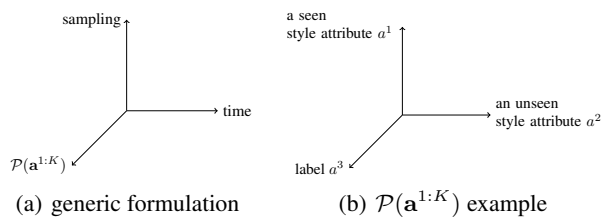

\paragraph{Standardized Settings.}
To streamline standardized evaluations of TTA methods, we first equip the benchmark library with shared data loaders for a set of common datasets, including CIFAR10-C~\cite{hendrycks2018benchmarking}, CIFAR10.1~\cite{recht2018cifar10.1}, ImageNet-C~\cite{hendrycks2018benchmarking}, OfficeHome~\cite{venkateswara2017deep}, PACS~\cite{li2017deeper}, ColoredMNIST~\cite{arjovsky2019invariant}, and Waterbirds~\cite{sagawa2019distributionally}.
These datasets allow us to examine each TTA method under various shifts, ranging from common image corruptions and natural style shifts that are widely used in prior literature to time-varying shifts and spurious correlation shifts that remain under-explored in the field, as detailed in \cref{appendix:datasets}.

% To enable greater flexibility and extendibility, we further introduce a fine-grained formulation of distribution shifts.
% Specifically, we generalize the notations in \cref{sec:preliminary} and decompose data into an underlying set of factors of variations, i.e., we assume a joint distribution $\cP$ of (i) inputs $\xx$ and (ii) corresponding attributes $\aa^{1:K} := \{ a^1, \ldots, a^k, \ldots, a^K \}$, where the values of attribute $a^k$ are sampled from a finite set.
% \yuejiang{add a sentence summarizing the unique advantage of this formulation? @Tao}

% A fine-grained formulation for test-time distribution shifts is formulated in \autoref{fig:test_time_distribution_shift_formulation}.
% Figure~\autoref{fig:test_time_distribution_shift_formulation_overall} decipts an overall framework, by considering \circled{1} the underlying distribution of attribute values $\cP(\aa^{1:K})$, \circled{2} sampling operators (e.g., \# of sampling trials and sampling distribution), and \circled{3} the concatenation of sampled data over time-slots.

To enable greater flexibility and extensibility that can go beyond existing settings, we further introduce a fine-grained formulation to capture a wide spectrum of empirical data distribution shifts.
Specifically, we generalize the notations in \cref{sec:preliminary} and decompose data into an underlying set of factors of variations, i.e., we assume a joint distribution $\cP$ of (i) inputs $\xx$ and (ii) corresponding attributes $\aa^{1:K} := \{ a^1, \ldots, a^k, \ldots, a^K \}$, where the values of attribute $a^k$ are sampled from a finite set.
As shown in Figure~\autoref{fig:test_time_distribution_shift_formulation_overall}, the empirical data distribution is characterized by \circled{1} the underlying distribution of attribute values $\cP(\aa^{1:K})$, \circled{2} sampling operators (e.g., \# of sampling trials and sampling distribution), and \circled{3} the concatenation of sampled data over time-slots.
Figure~\autoref{fig:test_time_distribution_shift_formulation_attribute_distribution} exemplifies the distribution of data $\cP(\aa^{1:K})$ through three attributes.

This formulation encompasses several kinds of distribution shifts, wherein the test data $\cP_{\cT}$ deviates from the training data $\cP_{\cS}$ across all time slots:
\begin{enumerate}[nosep,leftmargin=12pt]
	\item \emph{attribute-relationship shift} (a.k.a. spurious correlation): attributes are correlated differently between $\cP_{\cS}$ and $\cP_{\cT}$.
	\item \emph{attribute-values shift}: the distribution of attribute values under $\cP_{\cS}$ are differ from that of $\cP_{\cT}$.
	      Its extreme case generalizes to the shift that some attribute values are unseen under $\cP_{\cS}$ but are under $\cP_{\cT}$.
\end{enumerate}

\paragraph{Extendable Baselines.}

Given the rich set of distribution shifts described above, we benchmark $11$ TTA methods: Batch Normalization Test-time Adaptation (BN\_Adapt~\citep{schneider2020improving}), Test-time Entropy Minimization (TENT~\citep{wang2021tent}), Test-time Template Adjuster (T3A~\citep{iwasawa2021test}), Source Hypothesis Transfer (SHOT~\citep{liang2020we}), Test-time Training (TTT~\citep{sun2020test}), Marginal Entropy Minimization (MEMO~\citep{zhang2022memo}), Non-i.i.d.\ Test-time Adaptation (NOTE~\citep{gong2022note}), Continual Test-time Adaptation (CoTTA~\citep{wang2022continual}), Conjugate Pseudo-Labels (Conjugate PL~\cite{goyal2022test}), Sharpness-aware Entropy Minimization (SAR~\cite{niu2023towards}), and Fisher Regularizer~\citep{niu2022efficient}.
These algorithms are implemented in a modular manner to support the seamless integration of other components, such as different model selection strategies. More implementation details of the TTAB can be found in~\cref{appendix:ttab_implementation_details}.

\section{Batch Dependency Obstructs TTA Tuning} \label{sec:hyperparam_tuning}

As summarized in~\autoref{tab:setups}, TTA methods often come with a number of hyper-parameters, ranging from at least one up to six.
Yet, the influence of these hyper-parameters on adaptation outcomes, as well as the optimal strategies for tuning them, remains poorly understood.
In this section, we will first shed light on these issues by examining the sensitivity of previous methods to hyperparameter tuning.
We will further investigate the underlying challenge by looking into the online adaptation dynamics through the lens of batch dependency.
We will finally propose two evaluation protocols that enable a more objective assessment of TTA methods through upper-bound performance estimates.

\subsection{Sensitivity to Hyperparameter Tuning}

\begin{figure}[!t]
	\centering
	\subfigure[\small TENT]{
		\includegraphics[width=0.22\textwidth]{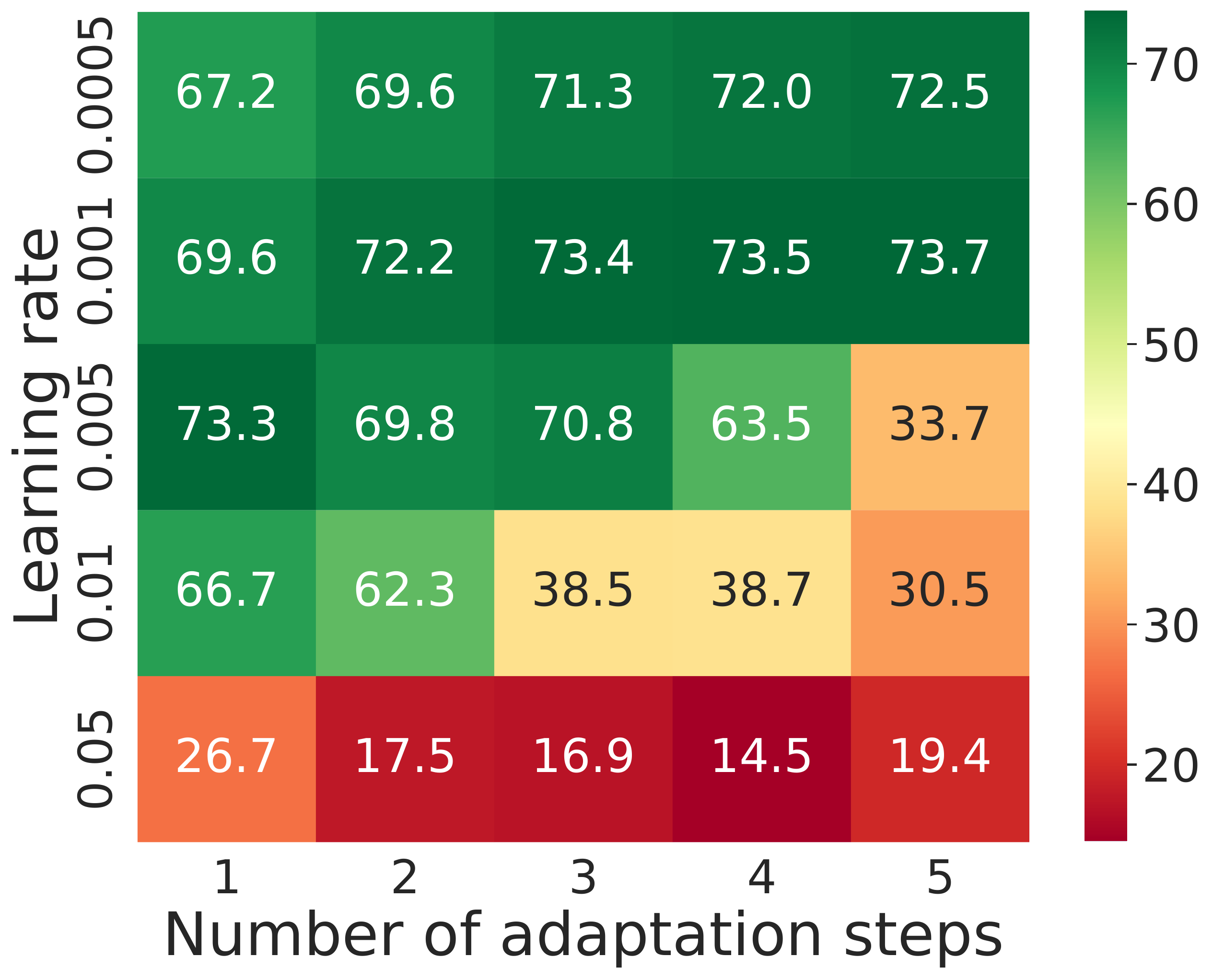}
		\label{fig:hparam_sensitivity_tent}
	}
	\subfigure[\small SHOT]{
		\includegraphics[width=0.22\textwidth]{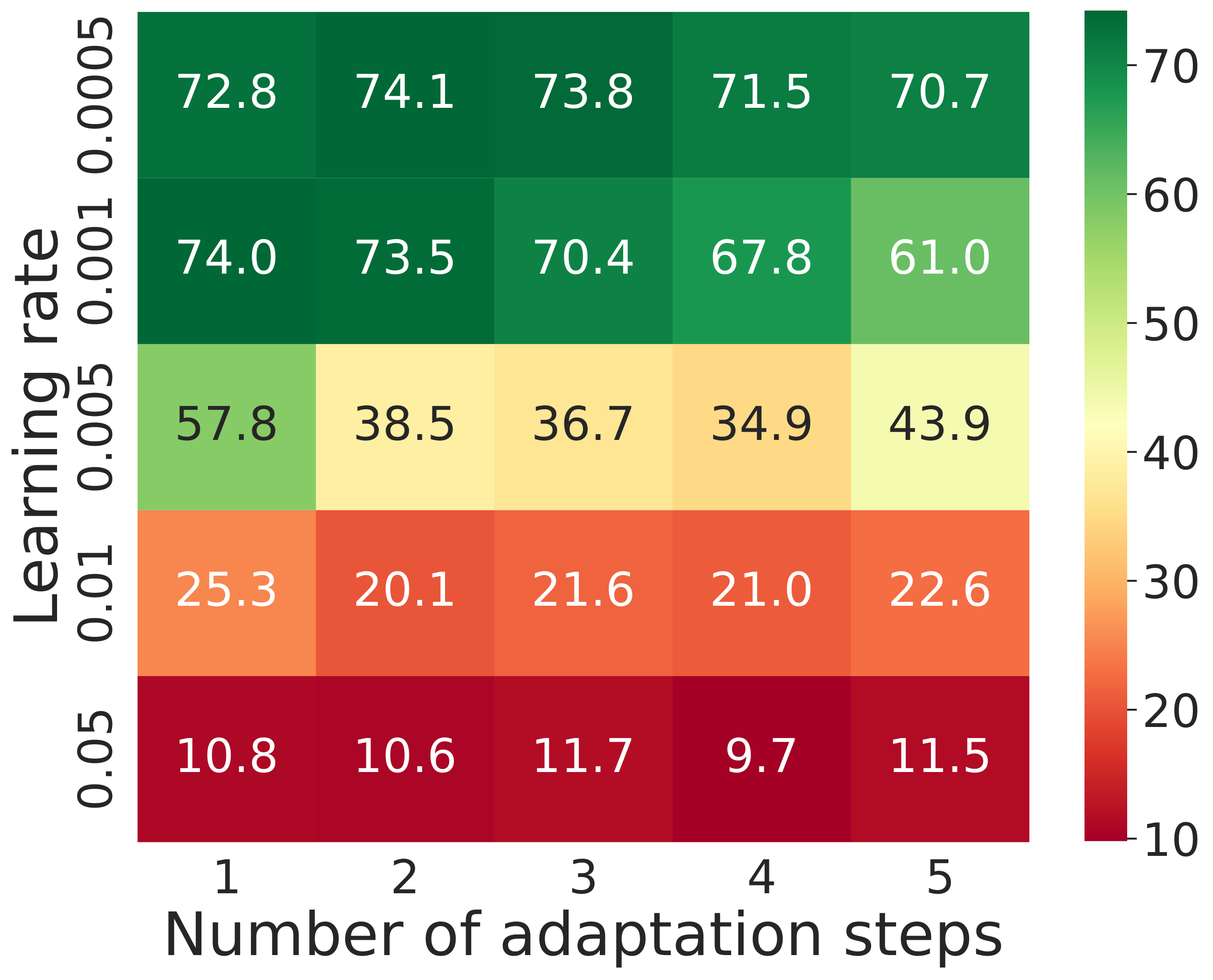}
		\label{fig:hparam_sensitivity_shot}
	}
	\caption{\small
		\textbf{On the hyperparameter sensitivity of TTA methods}, for evaluating the adaptation performance (test accuracy) of TENT and SHOT on CIFAR10-C (gaussian noise), under the combinations of learning rate and \# of adaptation steps.
		The results indicate that the commonly used practice of selecting hyperparameters, e.g.\ setting the number of adaptation steps to $1$ while slightly varying the learning rate, does not necessarily lead to an improvement in test accuracy (it may even have detrimental effects).
		This phenomenon occurs in all corruption types.
		%\looseness=-1
	}
	\label{fig:hparam_sensitivity}
	\vspace{-10pt}
\end{figure}

% \paragraph{TTA methods are highly sensitive to the choices of hyperparameters.}
% Selecting hyperparameters for TTA methods is challenging in practice, due to the inaccessible knowledge of the distribution shifts presented in the test streams.
% Existing evaluation usually leaves this under-explored, either by reusing existing values in the literature regardless of the testing scenario, or tuning in a limited region (see~\autoref{tab:setups} for a reference).
\paragraph{Empirical Sensitivity.}
To understand the importance of hyperparameter choices, we start by re-evaluating two renowned TTA methods, TENT and SHOT, with hyperparameters deviated away from the default values.
\autoref{fig:hparam_sensitivity} shows the test accuracy on the CIFAR10-C dataset resulting from different learning rates and adaptation steps.
We observe that the effectiveness of TTA methods is highly sensitive to both two considered hyperparameters.
Notably, an improper choice of hyperparameters can significantly deteriorate accuracy, with a decrease of up to 59.2\% for TENT and 64.4\% for SHOT.

% However, \emph{the effectiveness of TTA methods is heavily dependent on the selection of hyperparameters}: we can witness from a comprehensive evaluation in~\autoref{fig:hparam_sensitivity}, an improper choice of hyperparameters can lead to a significant degradation in accuracy, with a decrease of up to 59.2\% for TENT and 64.4\% for SHOT.
% \todo{expand explanations, emphasize variations}

% The issue of improper evaluation in TTA methods motivates us to design strategies for fair comparison among them.
% One approach to achieve fairness is by comparing the optimal test-time performance for each method with every pair of hyperparameters.
% While this is simple for \resetmodel settings, it is non-trivial for the majority of TTA methods under the \notresetmodel case (as shown in~\autoref{tab:unfair_comparison}) due to the batch-dependency issue (see~\cref{sec:batch_dependency}), where an oracle model selection during TTA cannot guarantee an optimal performance (see~\cref{sec:non_trivial_optimal_model_selection}).
% To address this, we propose two evaluation protocols in~\cref{sec:evaluation_protocols} as a way to estimate the performance upper-bound of the TTA method.

\begin{figure*}[!t]
	\centering
	\subfigure[\small Batch dependency exists in the \notresetmodel TTA setting with a single adaptation step.]{
		\includegraphics[width=0.31\textwidth]{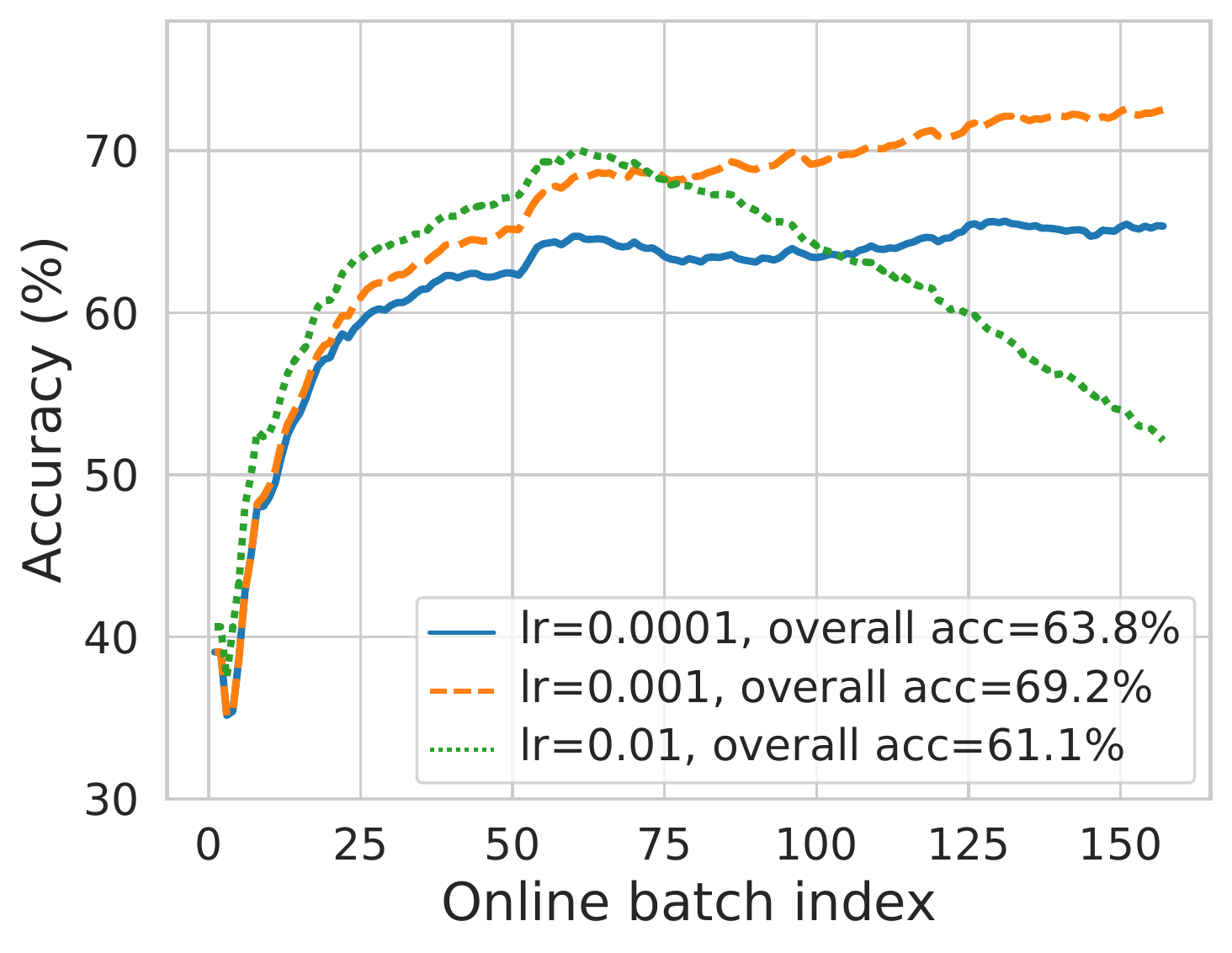}
		\label{fig:batch_dependency_single_step}
	}
	\hfill
	\subfigure[\small Multiple-step improves TTA but still has strong dependency among batches.]{
		\includegraphics[width=0.31\textwidth]{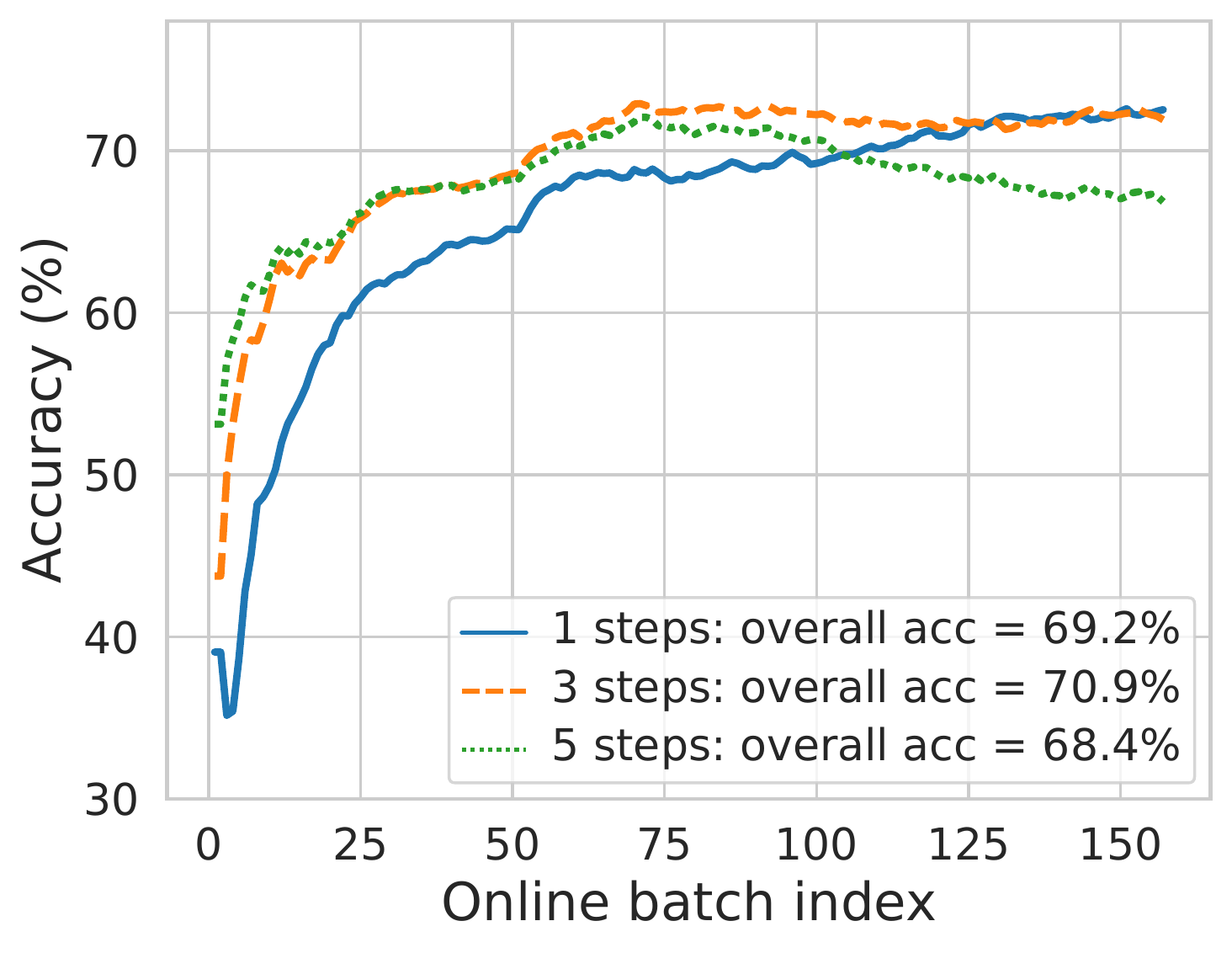}
		\label{fig:batch_dependency_multiple_steps}
	}
	\hfill
	\subfigure[\small Oracle model selection may introduce a more serious dependency problem to TTA.]{
		\includegraphics[width=0.31\textwidth]{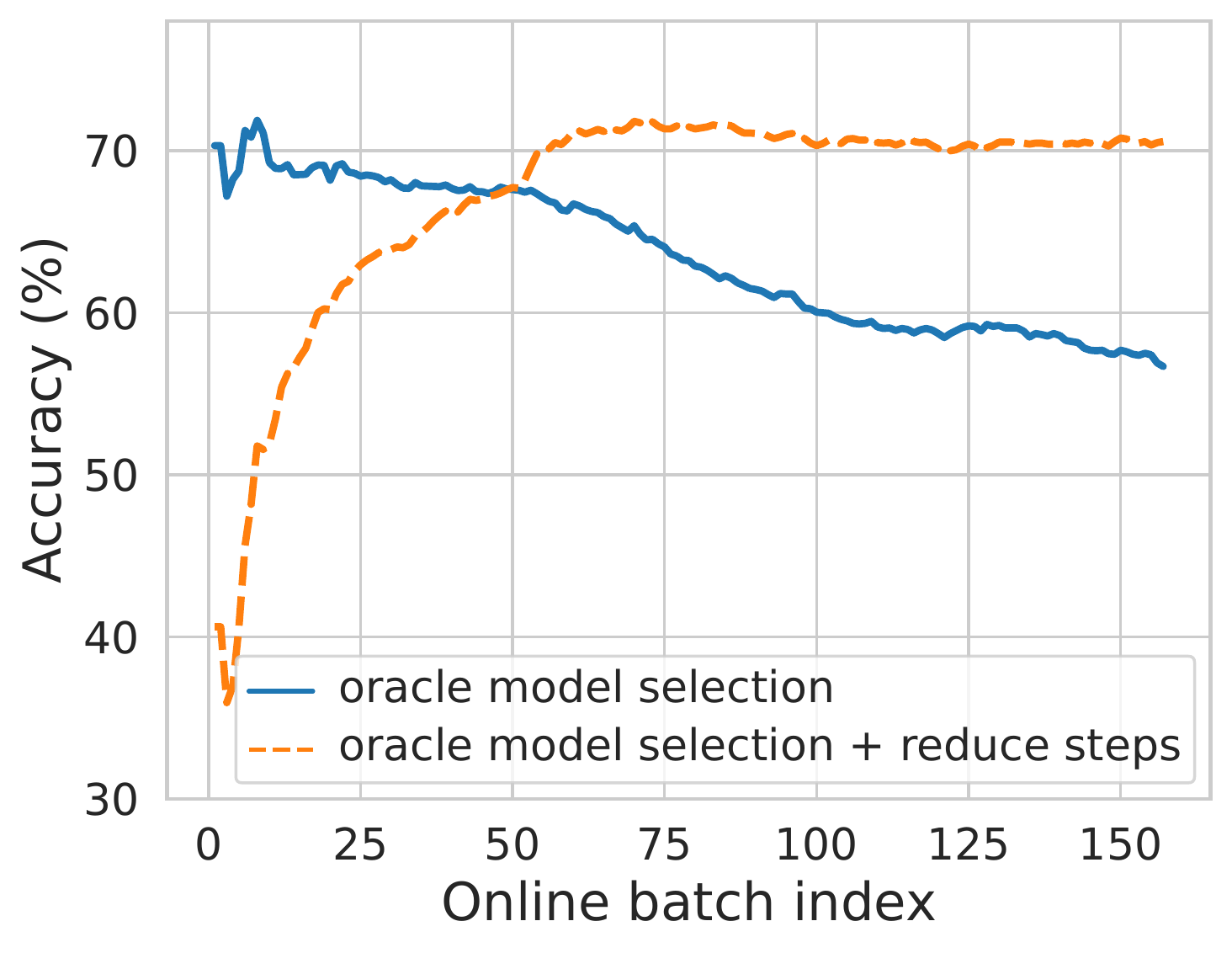}
		\label{fig:batch_dependency_oracle}
	}
	\caption{\small
		\textbf{The batch dependency issue during TTA and non-trivial model selection}, for evaluating SHOT on CIFAR10-C (gaussian noise).
		Similar trends can be found in all corruption types.
		SHOT suffers a significant decline in performance in an \notresetmodel adaptation setting, particularly when improper hyperparameters are chosen.
		Despite efforts to improve adaptation performance through the implementation of multiple adaptation steps, the problem of batch dependency remains unresolved.
		Oracle model selection, while providing reliable label information to guide the adaptation process at test time, ultimately leads to even more severe dependency issues.
		%\looseness=-1
	}
	\label{fig:batch_dependency}
	\vspace{-10pt}
\end{figure*}

% \todo{rewrite: emphasize that batch dependency is double-edged}

% \subsubsection{The Batch Dependency Issues During TTA} \label{sec:batch_dependency}
% \paragraph{The choice of \resetmodel and \notresetmodel in TTA setups.}

% Most existing TTA methods, as identified in~\autoref{tab:unfair_comparison}, tend to leverage distribution knowledge (i.e.\ adaptation history) learned from previous test batches to improve the test-time performance on new samples, in which the choice of \notresetmodel is dominant over that of \resetmodel.
% This design inherently increases the risk of batch dependent during TTA.
%\looseness=-1

\paragraph{Batch Dependency.}

Given that most existing TTA methods tend to leverage distribution knowledge (i.e.\ adaptation history) learned from previous test batches to improve the test-time performance on new samples, we further examine the influence of hyperparameter choices on adaptation dynamics.
Figure~\ref{fig:batch_dependency_single_step} shows the common \notresetmodel setting with a single adaptation step and a range of learning rates, where we observe clear over-adaptation as TTA progresses with a large learning rate.
Moving to the phase of multiple adaptation steps with a relatively small learning rate in Figure~\ref{fig:batch_dependency_multiple_steps}, we observe that adaptation performance increases from 69.2\% to 70.9\%.
However, if we continue to increase the number of adaptation steps, the adaptation performance quickly drops to 68.4\% \emph{due to over-adaptation on previous test batches.}
The risk of over-adaptation raises a practical question: when should we terminate TTA given a stream of test examples?
We next examine the challenge of model selection in the online TTA setting.

% This phenomenon, in which adaptation through entropy minimization in a non-i.i.d. scenario may silently degrade the model, has been reported by~\cite{boudiaf2022parameter}.\hao{We report a degeneration phenomenon in a more general setting: i.i.d. It might be worth mentioning the difference with LAME.}

% Although overall adaptation performance degrades when choosing an improper hyperparameter value, we still observe an upward trend in the accuracy curve in the beginning, indicating that adaptation proxies like entropy minimization can adapt model parameters to distribution shifts, but are vulnerable to the batch dependency problem.

% In conclusion, dependency on adaptation history is a common issue in dominant TTA methods, making adaptation on subsequent test samples difficult, particularly when using inappropriate hyperparameters.
% This phenomenon is particularly pronounced in the \notresetmodel TTA setting with varying hyperparameter combinations.
% In \cref{sec:non_trivial_optimal_model_selection}, we will further elaborate on the amplified negative effects caused by batch dependency and oracle model selection.
% \todo{add transition}

\subsection{Difficulty of TTA Model Selection} \label{sec:fair_evaluation}

% \todo{start with empirical ruglarizers (move the remark to here)}

% \subsubsection{Non-trivial Model Selection During TTA} \label{sec:non_trivial_optimal_model_selection}

% Proper model selection during TTA is crucial for estimating the optimal performance under every choice of hyperparameters.
% Note that DG only considers examining the time-varying scenarios very recently~\citep{yao2022wild}.

Model selection has recently gained great attention in the field of Domain Generalization~\citep{gulrajani2021in} and Domain Adaptation~\citep{you2019towards}.
Yet, its importance and necessity in the context of TTA have been largely unexplored.
We seek to shed light on this by exploring model selection in two paradigms: (i) with oracle information and (ii) with auxiliary regularization.

% The challenges in TTA also differ from those of traditional DG, due to the issues of (i) the lack of validation set and label information during test time; and (ii) batch-dependency issues emerged in the streaming test mini-batches making the oracle model selection method challenging.

% \todo{further extend to oracle labels}
\paragraph{Oracle Information.}
We first consider an oracle setting, where we assume access to true labels and select the optimal model (with early stopping) for each test batch with a sufficient number of adaptation steps.
This approach is expected to achieve the highest possible adaptation performance per adaptation batch.
For the sake of simplicity, we select the method-specific hyperparameters of each TTA method following the prior work (see more details in~\cref{appendix:tta_implementation_details}), while focusing on tuning two key adaptation-specific hyperparameters, namely learning rate and number of adaptation steps, which are highly relevant to the adaptation process detailed in~\cref{sec:improper_eval}.
We choose the maximum steps in Algorithm~\ref{algo:oracle_model_selection} as 50 according to our observation in~\autoref{fig:hparam_sensitivity} and set the maximum steps as 25 in large-scale datasets due to the computational feasibility.
The implementation is detailed in Algorithm~\ref{algo:oracle_model_selection}.

% Additionally, we aim to mitigate the influence of sensitivity to hyperparameters through the use of this rigorous, though unrealistic, model selection method.
%\looseness=-1
% Next, we will use this oracle model selection strategy to evaluate TTA methods.

% \looseness=-1

% \paragraph{Observation: oracle model selection is not optimal in TTA.}
Figure~\ref{fig:batch_dependency_oracle} shows that utilizing an oracle model selection strategy in TTA methods under an online adaptation setup with sufficient adaptation steps initially improves adaptation performance in the first several test batches, compared to Figure~\ref{fig:batch_dependency_single_step} and~\ref{fig:batch_dependency_multiple_steps}.
However, such improvement is short-lived, as the adaptation performance quickly drops in subsequent test batches.
It suggests that \emph{the oracle model selection strategy exacerbates the batch dependency problem when considering its use in isolation}.
This phenomenon is consistent across various choices of learning rates. Additionally, we find the same problem in TENT and NOTE as shown in~\autoref{fig:batch_dependency_in_tent_note} of~\cref{appendix:batch_dependency_in_tent_note}.

% In Figure~\ref{fig:batch_dependency_oracle}, we examined the impact of oracle model selection on TTA methods in an \notresetmodel adaptation setup with ample adaptation steps.
% We observed that adaptation performance increases rapidly in the first few test batches compared to Figure~\ref{fig:batch_dependency_single_step} and~\ref{fig:batch_dependency_multiple_steps}, where we only adapted a limited number of steps.
% However, adaptation performance then drops in subsequent test batches, which illustrates that the oracle model selection strategy exacerbates the batch dependency issue when considering the use of oracle model selection alone.
% This phenomenon can be observed across various choices of learning rates.

% \todo{rephrase}
% Given that no regularization techniques were applied during test time, the failure of oracle model selection may be attributed to the batch dependency and the model's overfitting to test batches it has seen.
% To further investigate this hypothesis, we systematically reduced the number of adaptation steps for each test batch in Figure~\ref{fig:batch_dependency_oracle}.
% Our results indicate that such modification mitigates the deterioration and leads to improved adaptation performance from 61.3\% to 69.8\%.

\paragraph{Auxiliary Regularization.}
Given the suboptimality of the oracle-based model selection, we further investigate the effect of auxiliary regularization on mitigating batch dependency.
Specifically, we consider Fisher regularizer~\citep{niu22efficient} and stochastically restoring~\citep{wang2022continual}, two regularizers originally proposed for non-stationary distribution shifts.
Our results in~\autoref{fig:regularization_techniques} of~\cref{appendix:batch_dependency_regularization} indicate that \emph{while these strategies may alleviate the negative effects of batch dependence to some extent, there is currently no principle to trade-off the adaptation and regularization within a test batch, and leave the challenge of balancing adaptation across batches touched.}
These techniques are infeasible to consider in model selection and cannot provide a fair assessment for TTA methods, due to the increased sensitivity to their hyperparameters; see a significant variance caused by the regularization method across different learning rates and adaptation steps in~\autoref{fig:new_methods_under_multiple_steps} of~\cref{appendix:batch_dependency_regularization}.
%\looseness=-1

% \section{TTAB: A PyTorch Testbed for \underline{T}est-\underline{T}ime \underline{A}daptation \underline{B}enchmark}

\begin{algorithm}[!t]
	\caption{Oracle model selection for \notresetmodel TTA}
	\label{algo:oracle_model_selection}
	\begin{algorithmic}[1]
		\myState{
			\textbf{Input}: model state $\mtheta^o$, test sample $\xx^{(t)}$, true label $y^{(t)}$, maximum adaptation steps $M$, learning rate $\eta$, objective function $\ell$, update rule $\cG$, and model selection metric $\cJ$.
			% \looseness=-1
		}
		% \Procedure{Oracle\_model\_selection}{$f_{\theta}, \xx_t, y_t, M, \eta, \ell, G, \mathcal{J}$}
		\Procedure{Oracle\_model\_selection}{$\mtheta, \ldots$}
		\myState{\textbf{Initialize}: $m \gets 1, \cF \gets \{ \mtheta \}, \mtheta_m \gets \mtheta$}
		\For{$m \in \{1, \cdots, M\}$}
		\myState{Compute loss $\tilde \ell \approx \ell(\mtheta_{m}; x^{(t)})$}
		\myState{Adapt parameters via $\mtheta_{m+1} \gets \cG(\mtheta_{m}, \eta, \tilde \ell)$}
		\myState{$\cF \gets \cF \cup \mtheta_{m+1}$}
		\EndFor
		\myState{Select optimal model $\mtheta^\star \!\gets\! \argmax_{ \tilde{\mtheta} \in \cF} \cJ( \tilde{ \mtheta }, y^{(t)})$}
		\myState{\Return Pass $\mtheta^\star$ to next test sample $\xx^{(t+1)}$}
		\EndProcedure
	\end{algorithmic}
\end{algorithm}

\subsection{Evaluation with Oracle Model Selection} \label{sec:evaluation_protocols}
In light of the aforementioned model selection difficulty, we design two evaluation protocols for estimating the potential of a given TTA method.
The first one resorts to \resetmodel adaptation with oracle model selection.
It fully eliminates the impact of batch dependency, resulting in stable TTA outcomes.
However, the performance gain of this protocol is often limited as it discards the valuable information from the previous batch about test data distribution.

As an alternative, the use of \notresetmodel adaptation empowers a large potential by accumulating historical knowledge.
However, it presents a batch dependency challenge, posing model selection during TTA as a min-max equilibrium optimization problem across time and potentially leading to a significant decline in performance.
To mitigate this issue, we use oracle model selection in conjecture with grid search over the best combinations of learning rates and adaptation steps.
While such a traverse is computationally expensive, it allows for a reliable estimate of the optimal performance of each TTA method.

\begin{figure*}[!t]
	\centering
	\subfigure[\small OOD v.s.\ OOD (BN\_Adapt).]{
		\includegraphics[width=0.22\textwidth]{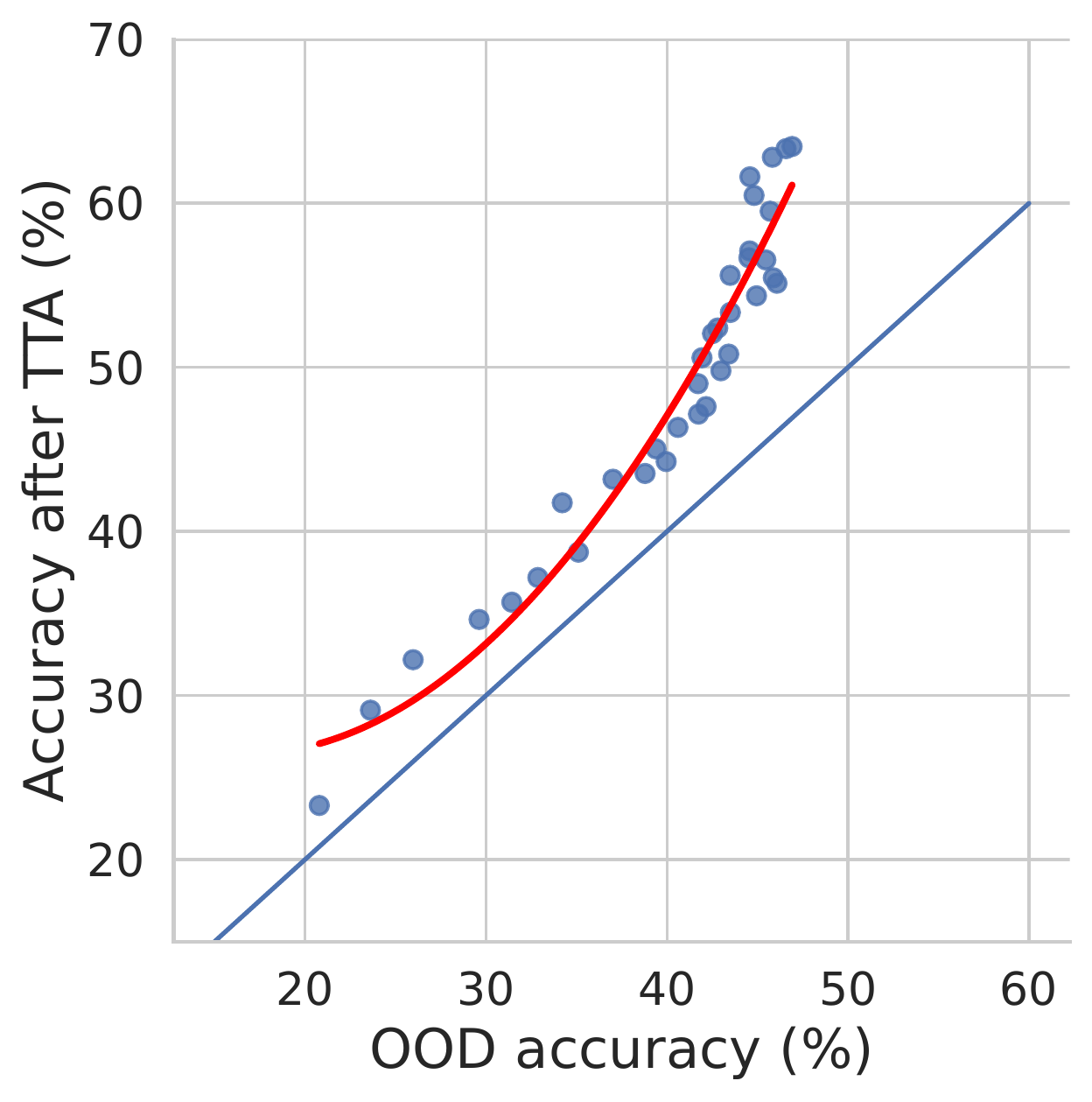}
		\label{fig:bn_adapt_feature_ext}
	}
	\subfigure[\small OOD v.s.\ OOD (TENT).]{
		\includegraphics[width=0.22\textwidth]{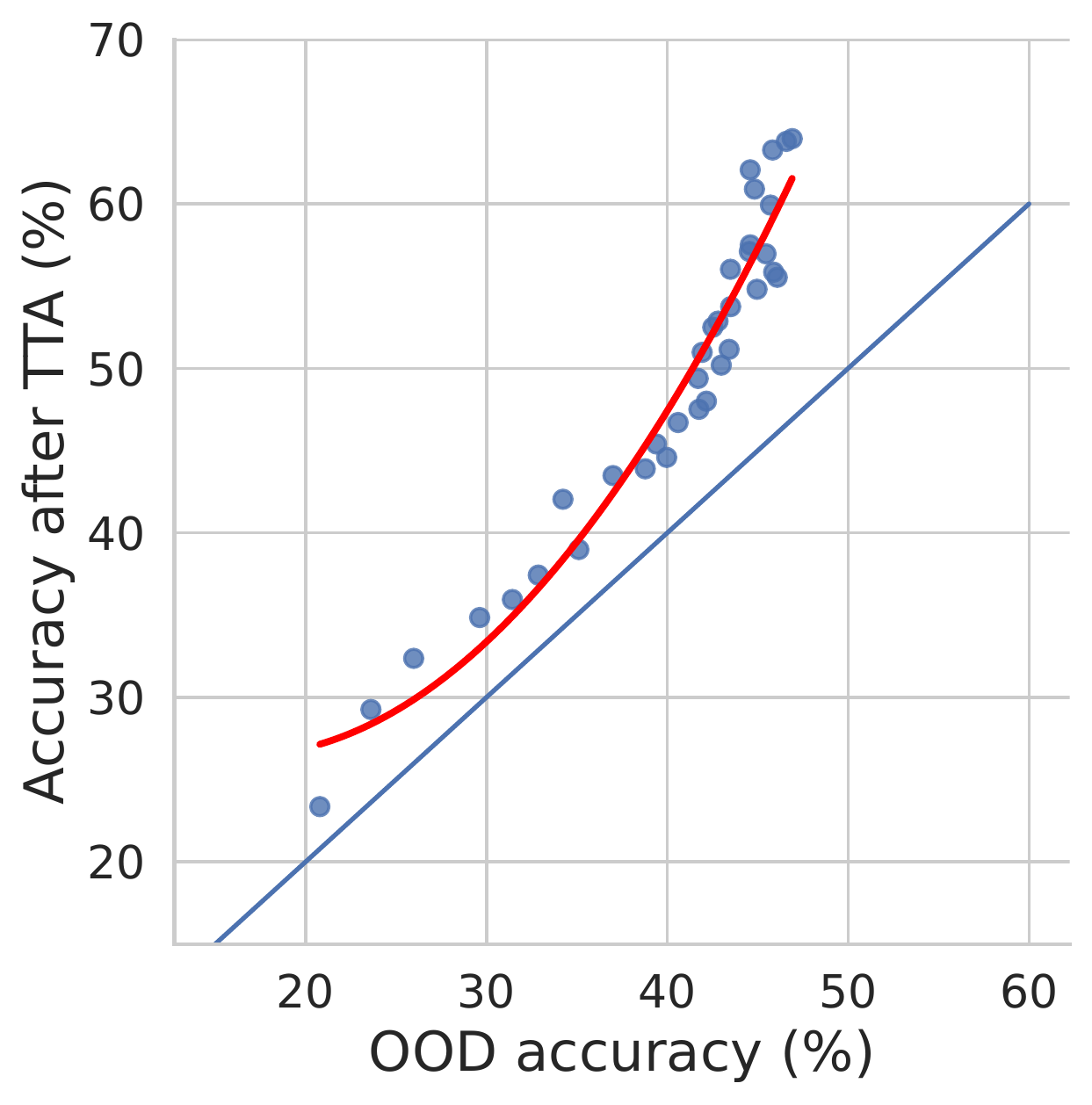}
		\label{fig:tent_adapt_feature_ext}
	}
	\subfigure[\small OOD v.s.\ OOD (SHOT).]{
		\includegraphics[width=0.22\textwidth]{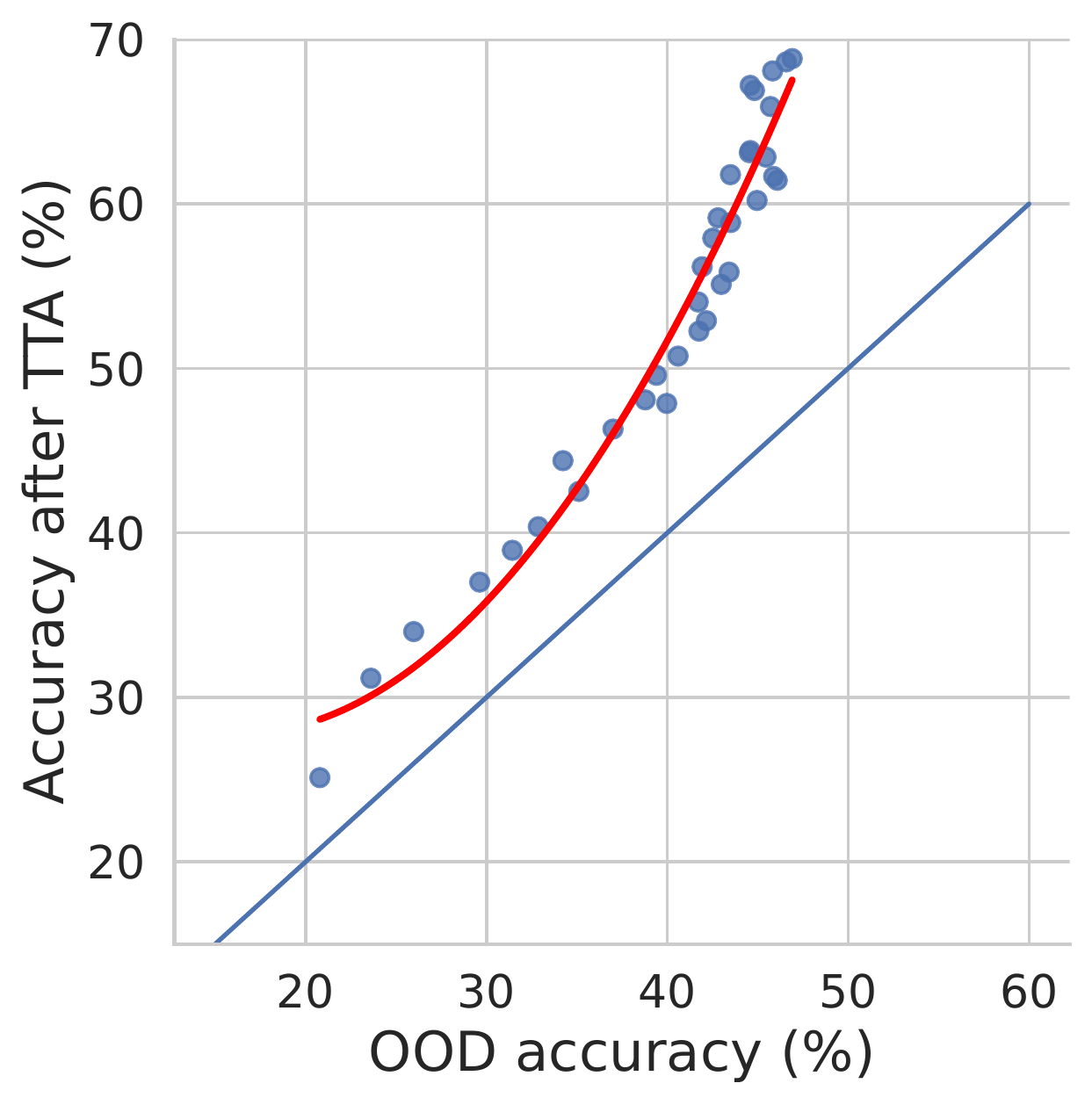}
		\label{fig:shot_adapt_feature_ext}
	}
	\subfigure[\small OOD v.s.\ OOD (T3A).]{
		\includegraphics[width=0.22\textwidth]{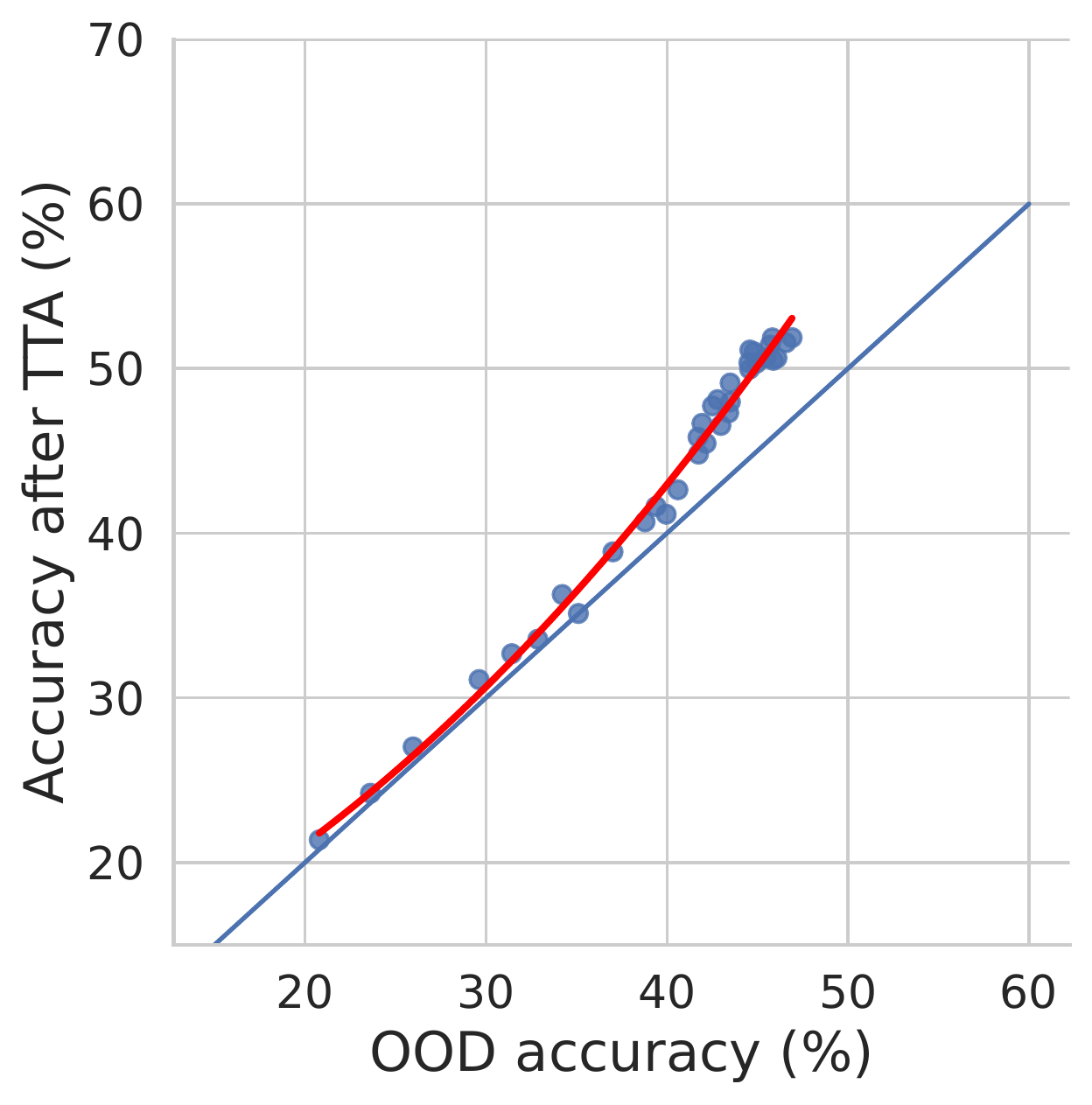}
		\label{fig:t3a_adapt_feature_ext}
	}
	\caption{\small
		\textbf{The impact of model quality on TTA performance, in terms of OOD v.s.\ OOD (TTA)} on CIFAR10-C.
		We save the checkpoints from the pre-training phase of ResNet-26 with standard augmentation and evaluate TTA performance on these checkpoints using oracle model selection.
		The OOD generalization performance has a significant impact on the overall performance (i.e.\ averaged accuracy of all corruption types) of various TTA methods.
		Our analysis reveals a strong correlation between model quality and the effectiveness of TTA methods.
		Furthermore, certain TTA methods, specifically SHOT, may not provide an improvement in performance on OOD datasets and may even result in a decrease in performance when applied to models of low quality.
	}
	\label{fig:feature_ext_quality}
	\vspace{-10pt}
\end{figure*}

\section{Pre-trained Model Bottlenecks TTA Efficacy} \label{sec:model_quality}

Recall that several recent TTA methods outlined in~\autoref{tab:unfair_comparison} necessitate modifications of pre-training, which naturally results in inconsistent model qualities across methods and may deteriorate the test performance even before the TTA.
In this section, we conduct a comprehensive and large-scale evaluation to examine the impact of base model quality on TTA performance across various TTA methods.

% In addition to the unfairness caused by the diverse hyperparameters with improper model selection during TTA, the need of modifying the pre-training phase in~\autoref{tab:unfair_comparison} naturally results in inconsistent model qualities across methods and may deteriorate the test performance even before the TTA.
% In this section, we conduct a comprehensive and large-scale evaluation to examine the impact of base model quality on TTA performance across various TTA methods.

% Prior work uses well-trained models to do Test-time Adaptation, and no work has investigated whether TTA algorithms have similar adaptation performance patterns when using models with poor or limited quality.
% As a result, it is of interest to determine the extent to which TTA methods can benefit from the quality of pre-trained models.
% \citep{miller2021accuracy} have demonstrated that out-of-distribution performance is closely linked to in-distribution generalization performance in many practical scenarios.

\paragraph{Evaluation setups.}
We thoroughly examine the pre-trained model quality from the aspects of (1) disentangled feature extractor and classifier, and (2) data augmentation. %\looseness=-1
\begin{enumerate}[nosep, leftmargin=12pt]
	\item We consider a model with decoupled feature extractor and classifier.
	      We keep the checkpoints with varying performance levels, generated from the pre-training phase using the standard data augmentation technique (mentioned below).
	      We then fine-tune a trainable linear classifier for each frozen feature extractor from the checkpoints, using data with a uniform label distribution, to study the effect of the feature extractors (equivalently full model).
	      To study the effect of the linear classifiers, we freeze a well-trained feature extractor and fine-tune trainable linear classifiers on several non-i.i.d.\ datasets created from a Dirichlet distribution; we further use Dirichlet distribution to create non-i.i.d.\ test data streams. %\looseness=-1
	      % , using the same pretraining process in order to minimize any unexpected influences on model quality.

	\item We consider 5 data augmentation policies: (i) no augmentations, (ii) standard augmentation, i.e.\ random crops and horizontal flips, (iii) MixUp~\citep{zhang2017mixup} combined with standard augmentations, (iv) AugMix~\citep{hendrycks2019augmix}, and (v) PixMix~\citep{hendrycks2022robustness}.
	      For each data augmentation method, we save the checkpoints from the standard supervised pre-training phase to cover a wide range of pre-trained model qualities.
	      %   Here we do not stick to self-supervised learning to pretrain models in order to reduce the confounding effect on model quality. 
	      % \looseness=-1
\end{enumerate}

\begin{figure*}[!t]
	\centering
	\subfigure[\small ID v.s.\ OOD]{
		\includegraphics[width=0.225\textwidth]{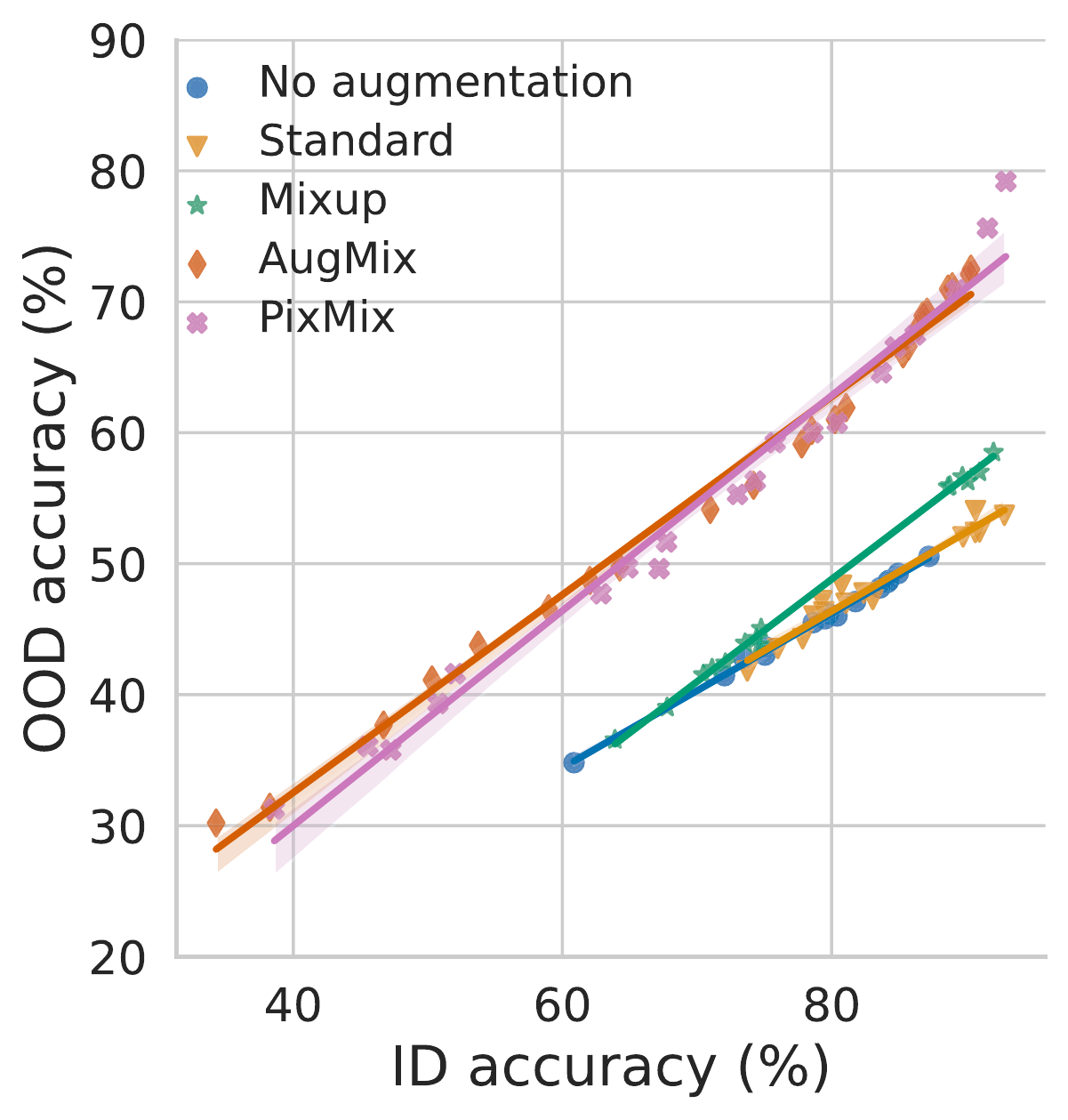}
		\label{fig:data_aug_rn26_id_vs_ood}
	}
	\subfigure[\small OOD v.s.\ OOD (BN\_Adapt)]{
		\includegraphics[width=0.225\textwidth]{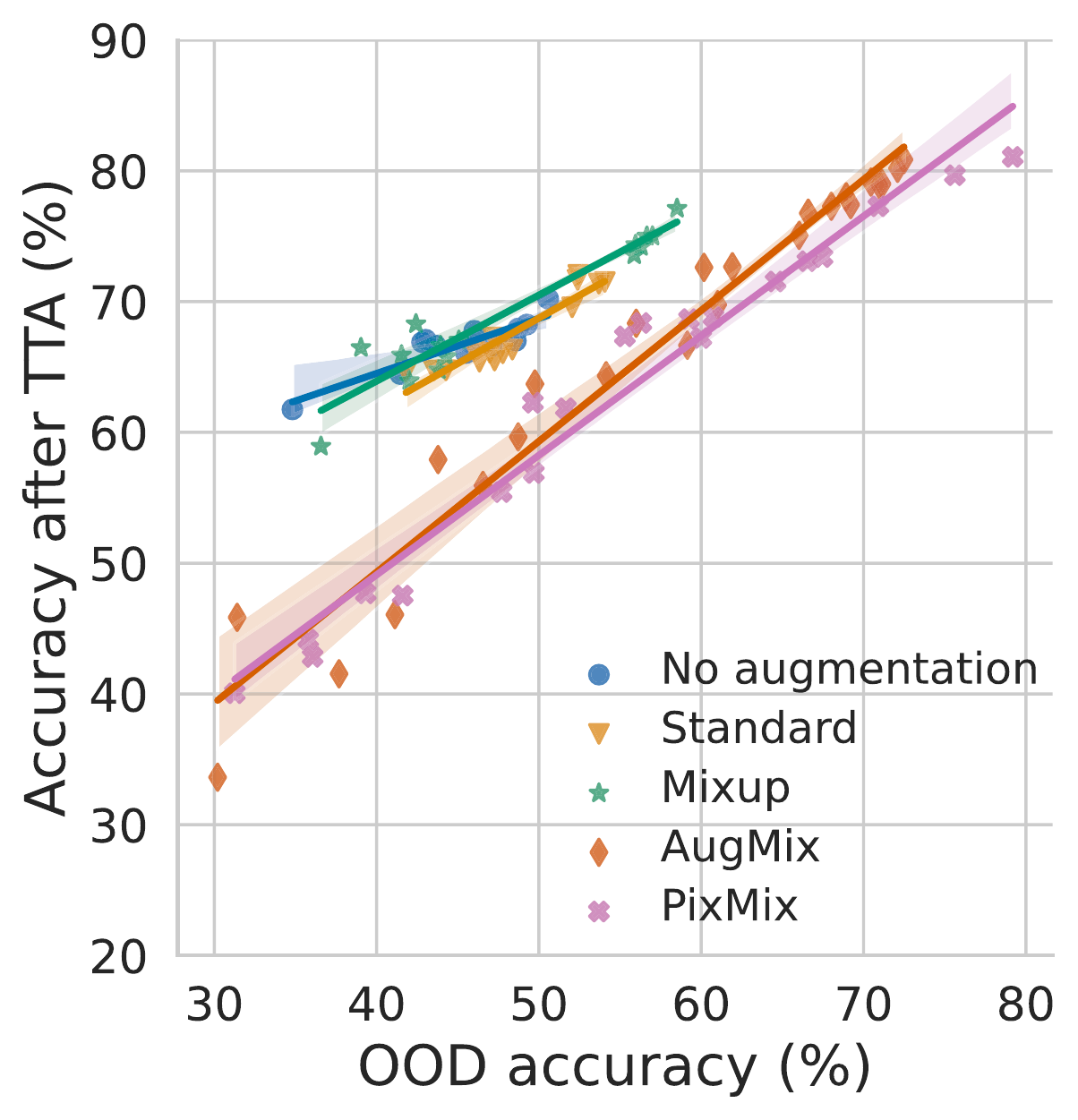}
		\label{fig:data_aug_rn26_bn_adapt}
	}
	\subfigure[\small OOD v.s.\ OOD (TENT)]{
		\includegraphics[width=0.225\textwidth]{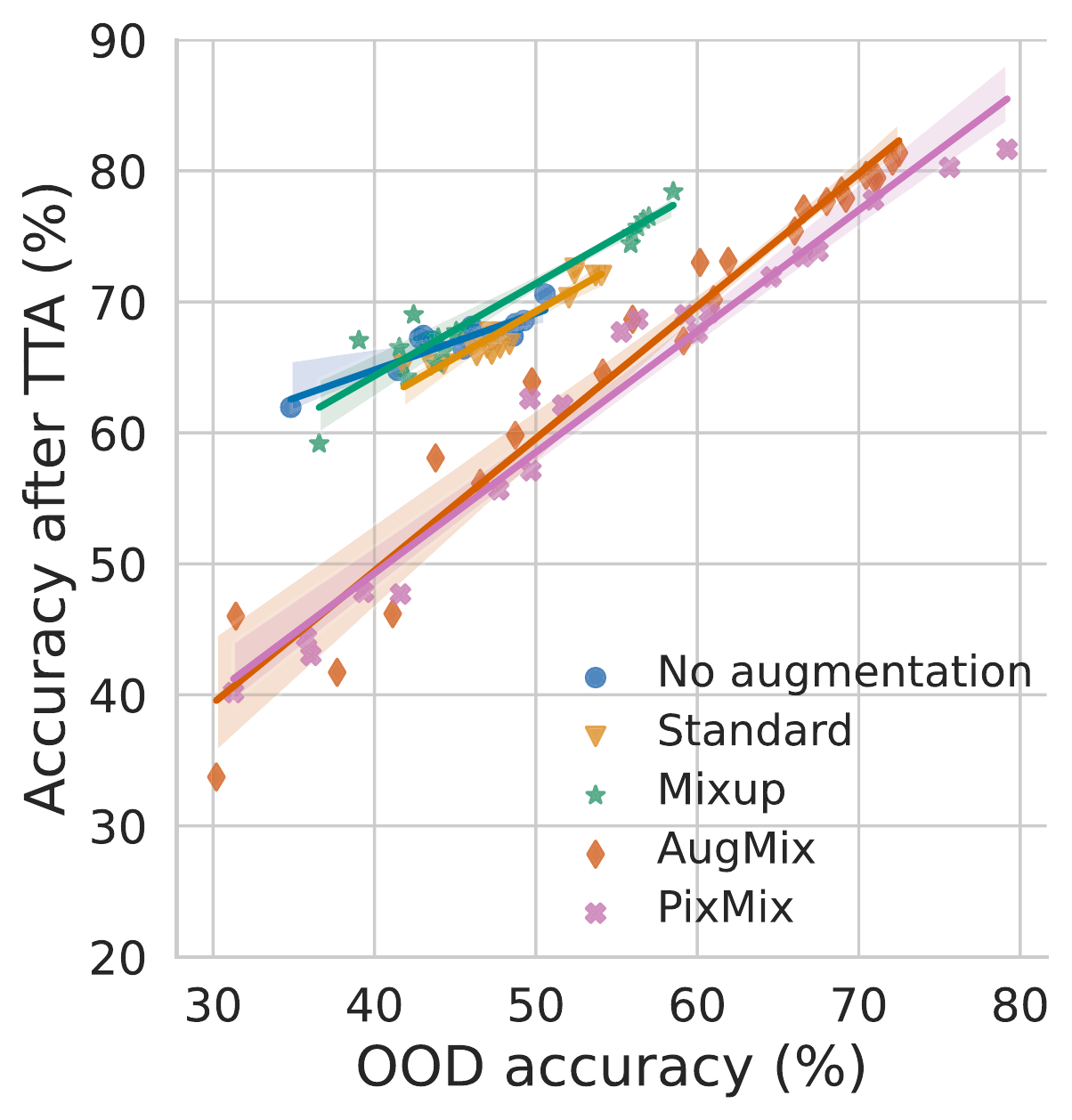}
		\label{fig:data_aug_rn26_tent}
	}
	\subfigure[\small OOD v.s.\ OOD (SHOT)]{
		\includegraphics[width=0.225\textwidth]{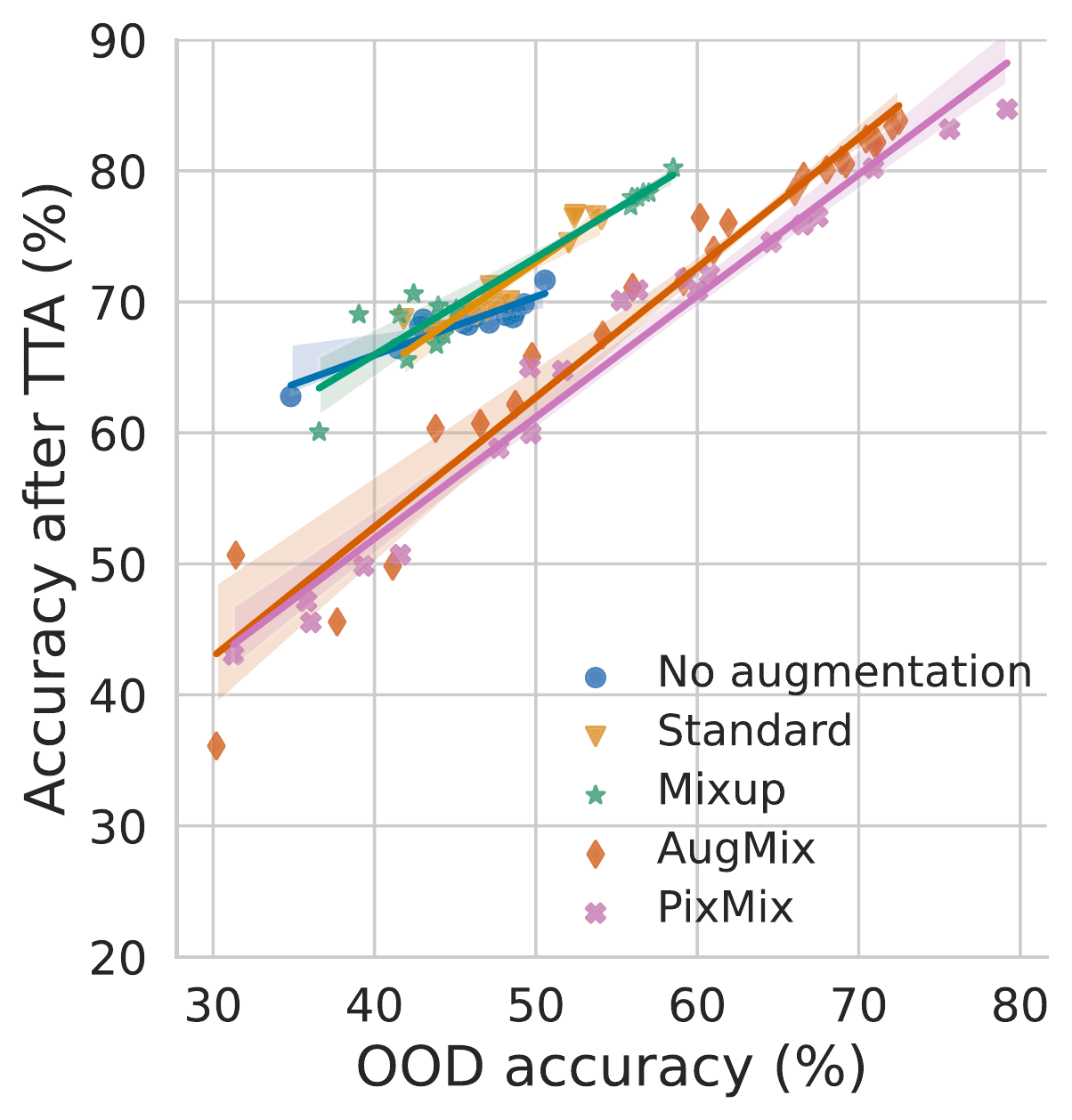}
		\label{fig:data_aug_rn26_shot}
	}
	\vspace{-0.75em}
	\caption{\small
		\textbf{The impact of data augmentation policy on the TTA performance of the target domain.}
		We save various sequences of checkpoints from the pre-training phase of ResNet-26 with five data augmentation policies and fine-tune each sequence to study the impact of data augmentation.
		TENT and SHOT use episodic adaptation with oracle model selection.
		Different data augmentation strategies have different corruption robustness, which causes varying generalization performance on CIFAR10-C.
		However, good practice in data augmentations and architecture designs for out-of-distribution generalization can be bad for test-time adaptation.
	}
	\label{fig:data_augmentations_rn26}
	\vspace{-10pt}
\end{figure*}

\begin{figure}[!t]
	\centering
	\subfigure[\small domain \#0]{
		\includegraphics[width=0.10\textwidth]{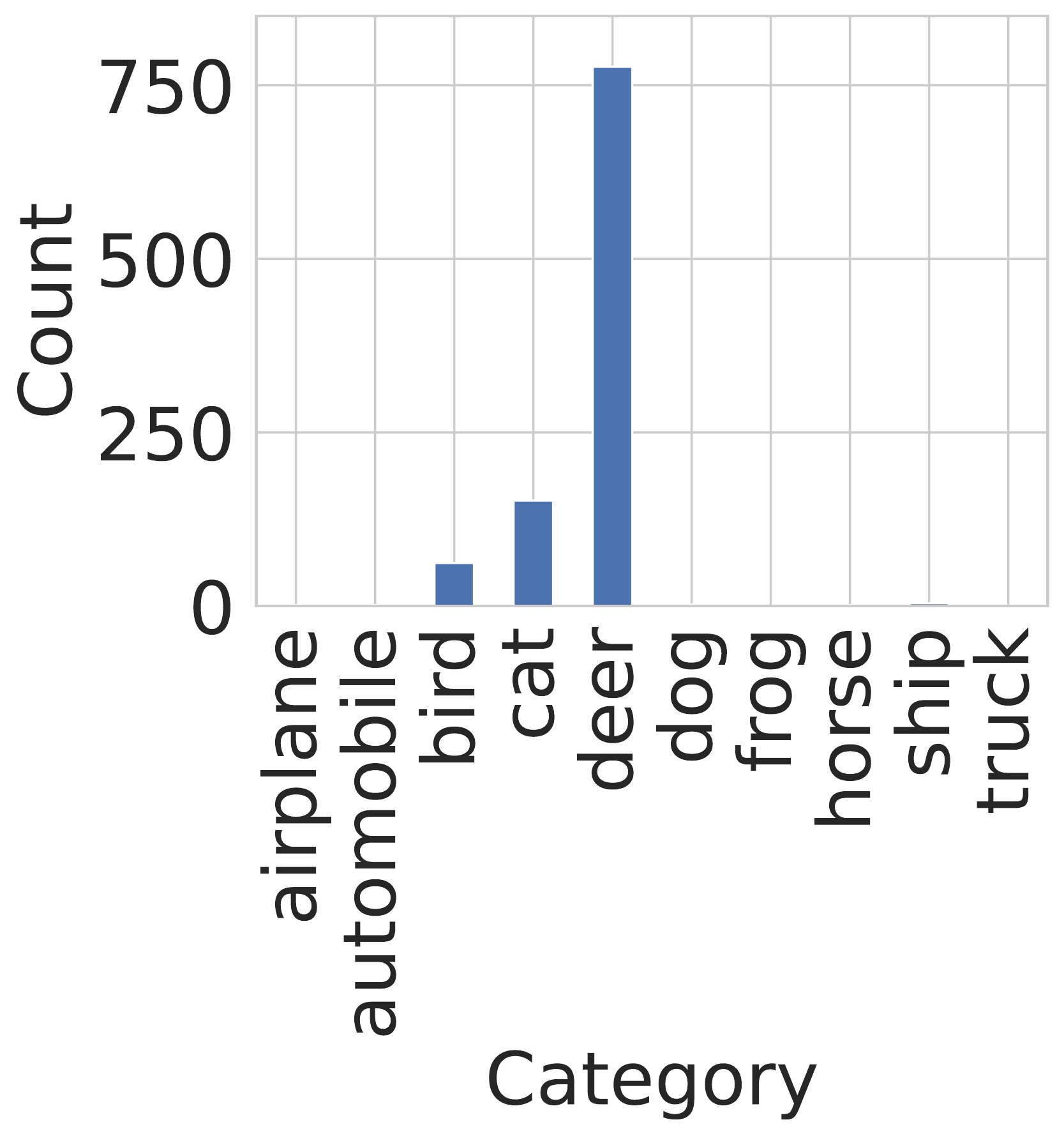}
		\label{fig:train_env}
	}
	\subfigure[\small domain \#1]{
		\includegraphics[width=0.10\textwidth]{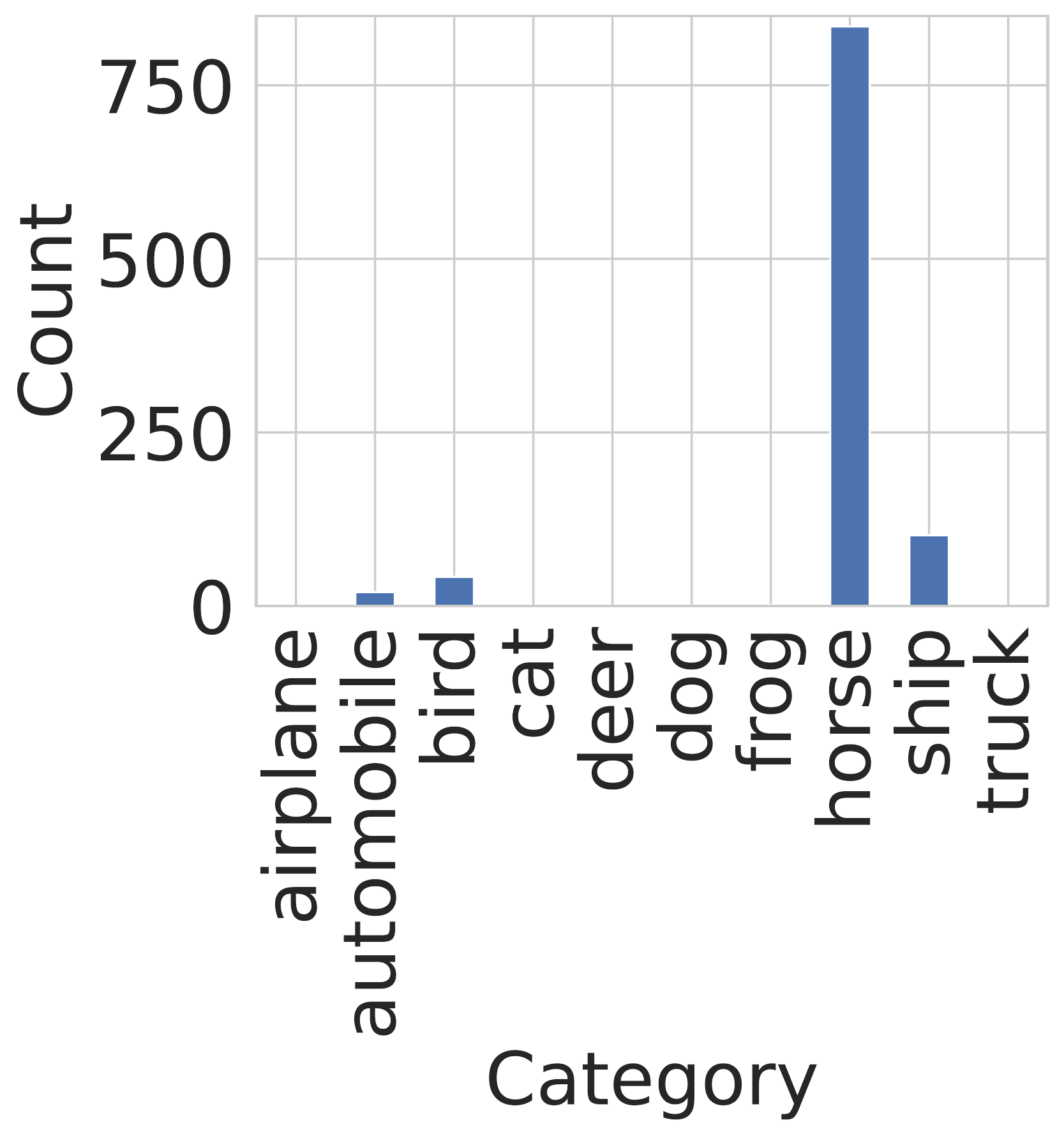}
		\label{fig:test_env0}
	}
	\subfigure[\small domain \#2]{
		\includegraphics[width=0.10\textwidth]{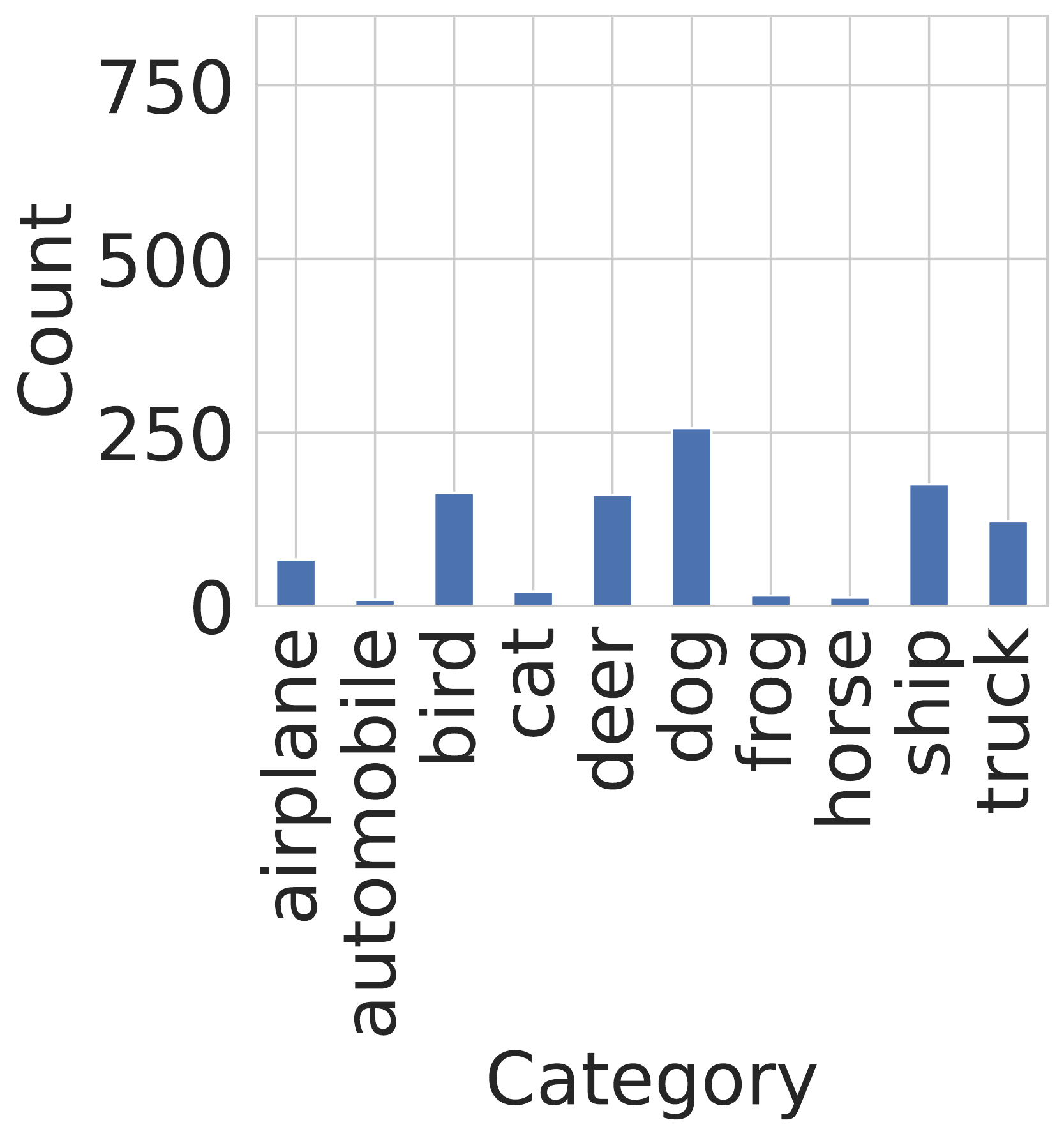}
		\label{fig:test_env1}
	}
	\subfigure[\small domain \#3]{
		\includegraphics[width=0.10\textwidth]{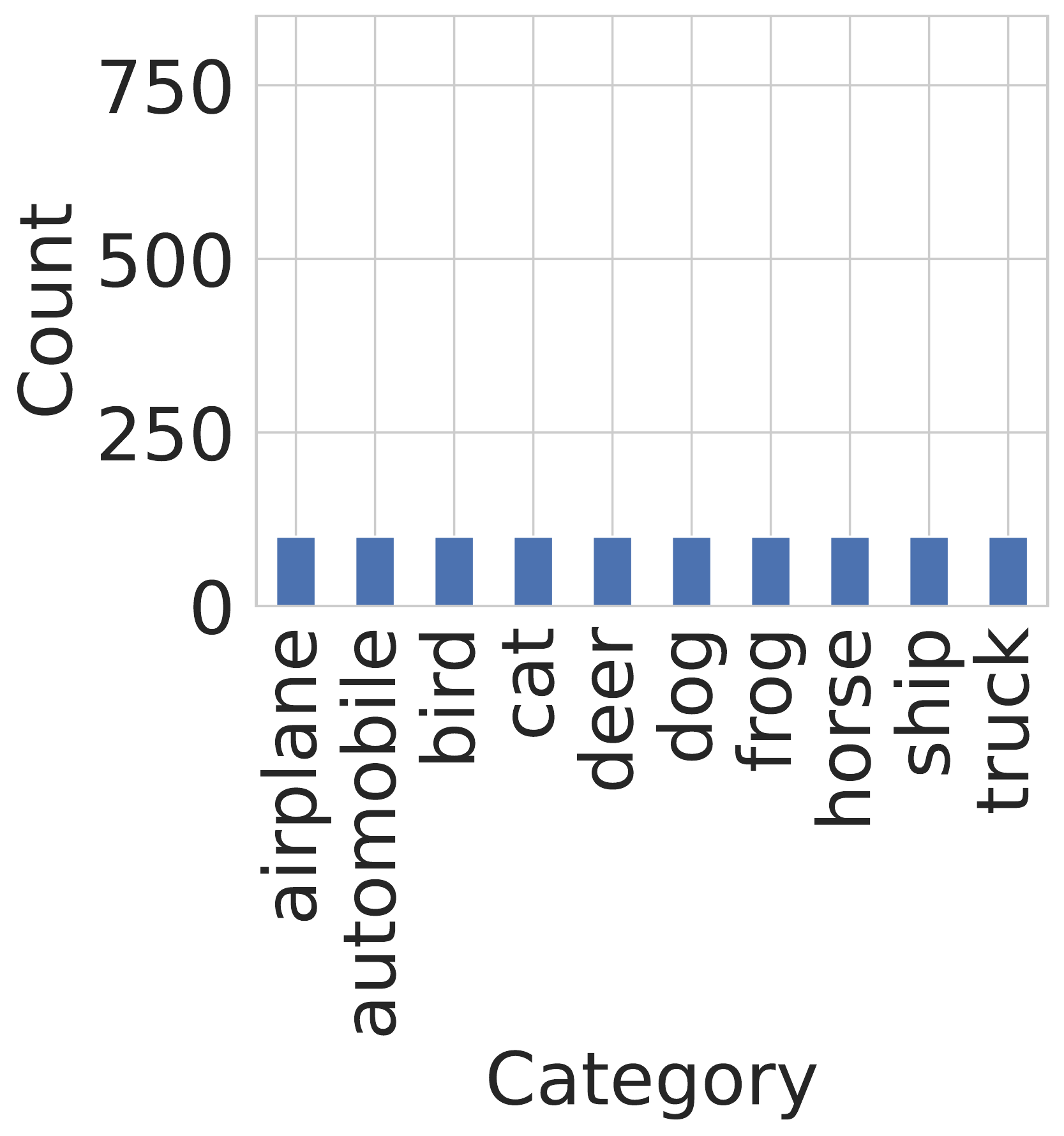}
		\label{fig:test_env2}
	}
	\resizebox{.48\textwidth}{!}{%
		\begin{tabular}{lcccc}
			\toprule
			          & domain \#0 (\%) $\downarrow$ & domain \#1 (\%) $\downarrow$ & domain \#2 (\%) $\downarrow$ & domain \#3 (\%) $\downarrow$ \\
			\midrule
			Baseline  & $12.6$                       & $96.5$                       & $79.9$                       & $76.1$                       \\
			\midrule
			BN\_Adapt & $7.9$                        & $98.4$                       & $85.8$                       & $80.7$                       \\
			\midrule
			T3A       & $22.0$                       & $96.0$                       & $77.5$                       & $74.8$                       \\
			\midrule
			TENT      & $7.0$                        & $98.1$                       & $84.4$                       & $80.1$                       \\
			\midrule
			SHOT      & $\mathbf{5.4}$               & $\mathbf{95.0}$              & $\mathbf{72.1}$              & $\mathbf{67.4}$              \\
			\midrule
			TTT       & $6.3$                        & $96.7$                       & $77.0$                       & $73.5$                       \\
			\midrule
			MEMO      & $10.1$                       & $97.4$                       & $83.7$                       & $80.3$                       \\
			\bottomrule
		\end{tabular}
	}
	\caption{\small
		\textbf{Adaptation performance (error) of TTA methods over CIFAR10-C with different label shifts}.
		(a) test domain \#0: $\alpha=0.1$, same label distribution with training environment.
		(b) test domain \#1: $\alpha=0.1$, different label distribution with training environment.
		(c) test domain \#2: $\alpha=1$.
		(d) test domain \#3: uniformly distributed test stream.
		We investigate the impact of the degree of non-i.i.d.-ness in the fine-tuning dataset on the performance of the linear classifier. Label smoothing~\citep{liang2020we} technique is used to learn higher quality features.
		Our findings reveal that the quality of the linear classifier plays a crucial role in determining the effectiveness of TTA methods, as they can only enhance performance on test data that shares similar i.i.d.-ness and label distribution characteristics.
		Despite utilizing a well-trained feature extractor, the quality of the linear classifier remains a significant determining factor in the overall performance of TTA methods.
	}
	\label{fig:classifier_quality}
	\vspace{-10pt}
\end{figure}

\paragraph{On the influence of the feature extractor (equivalently full model).}
The results of our study, as depicted in~\autoref{fig:feature_ext_quality}, reveal a strong correlation between the performance of test-time augmentation and out-of-distribution generalization on CIFAR10-C.
Our analysis shows that across a wide range of TTA methods, \emph{the OOD generalization performance is highly indicative of TTA performance}.
A quadratic regression line was found to be an appropriate fit for the data, suggesting that \emph{TTA methods are more effective when applied to models of higher (OOD) quality}.
% Our findings are further supported by the results presented in~\autoref{fig:feature_ext_quality}, which demonstrate that models that exhibit superior OOD generalization also tend to perform better under TTA.
% Additional details and measurements can be found in~\autoref{appendix:model_quality}.

\paragraph{On the influence of the linear classifier.}
Our study has revealed that the performance of TTA methods is significantly impacted by the quality of the feature extractor used.
The question then arises, can TTA methods bridge the distribution shift gap when equipped with a high-quality feature extractor and a suboptimal linear classifier?
Our analysis, as shown in Figure~\ref{fig:train_env}-\ref{sub@fig:test_env2}, indicates that most TTA methods on CIFAR10-C are only able to mitigate the distribution shift gap when the label distribution of the target domain is identical to that of the source domain, at which point the classifier is considered optimal.
In this case, SHOT attains a 5.4\% error rate, the best result observed in test domain \#0.
However, it is clear that all TTA methods either perform worse than the baseline in the remaining $3$ test domains or yield only marginal improvements over the baseline.
These findings suggest that \emph{the quality of the classifier plays a crucial role in determining the performance of TTA methods}.

\begin{figure}[!tb]
	\centering
	\subfigure[\small ID v.s.\ OOD]{
		\includegraphics[width=0.22\textwidth]{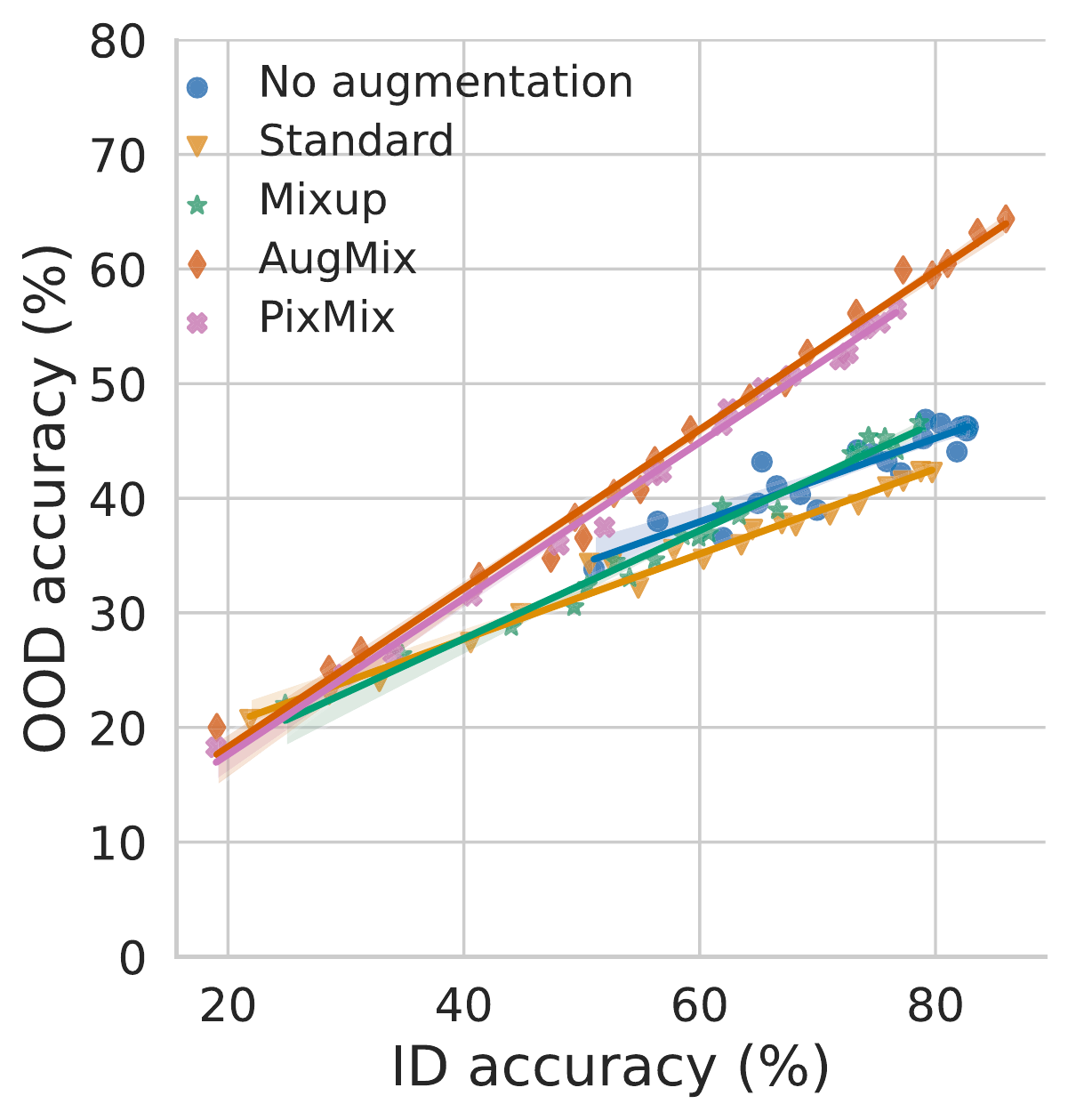}
		\label{fig:data_aug_cct_id_vs_ood}
	}
	\subfigure[\small OOD v.s.\ OOD (SHOT)]{
		\includegraphics[width=0.22\textwidth]{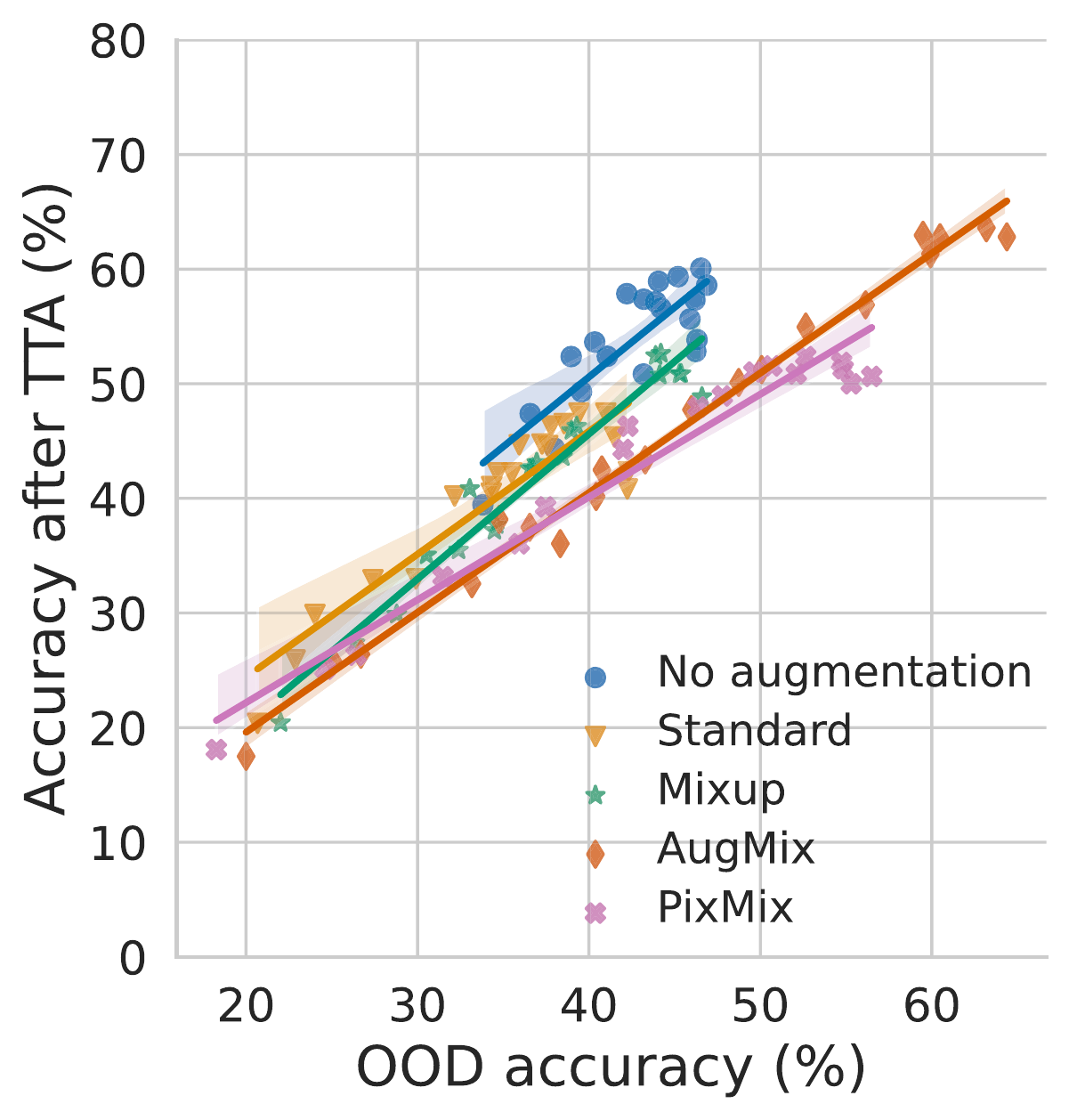}
		\label{fig:data_aug_cct_shot}
	}
	\vspace{-0.75em}
	\caption{\small
		\textbf{Revisit the impact of data augmentation policy on the TTA performance by using CCT.}
		With the same data augmentation policies as in~\autoref{fig:data_augmentations_rn26}, we save 5 sequences of checkpoints with different model quality and investigate the performance of SHOT under oracle model selection on CCT,
		a computationally efficient variant of ViT. The same trend as in ResNet-26 and WideResNet40-2 can be observed from CCT, emphasizing the unfavorable impact of strong data augmentation strategies on TTA performance regardless of the architecture designs.
	}
    \label{fig:data_augmentations_cct}
	\vspace{-10pt}
\end{figure}

\paragraph{On the influence of the data augmentation strategies.}
We investigate the impact of various augmentation policies on the performance of ResNet-26 models trained on the CIFAR10 dataset.
Our experimental results, as depicted in~\autoref{fig:data_augmentations_rn26} (more results in~\autoref{fig:data_augmentations_rn26_details} of~\cref{appendix:data_augmentation}), reveal that models pre-trained with the augmentation techniques like AugMix and PixMix exhibit superior OOD generalization performance on CIFAR10-C compared to models that do not utilize augmentation or only employ standard augmentations.
Interestingly, even though \emph{these robust augmentation strategies} significantly improve the robustness of the base model in the target domain, \emph{they only result in a marginal performance increase when combined with TTA.}
This disparity is particularly pronounced when compared to the performance of models trained with no augmentation or standard augmentations.
However, when all models are fully trained in the source domain, the use of techniques such as AugMix and PixMix still leads to the best adaptation performance on CIFAR10-C, owing to their exceptional OOD generalization capabilities.
We reach the same conclusion across both evaluation protocols and different architectures (e.g., WideResNet40-2) as shown in~\cref{appendix:data_augmentation}. In order to prove the influence of data augmentation strategies on TTA performance,
we also conduct experiments on CCT, a computationally efficient variant of ViT and present experimental results in~\autoref{fig:data_augmentations_cct}.
We highlight that good practice in strengthening the generalization performance of the base model in the target domain will decline its ability to bridge the distribution gap in the test time regardless of architecture designs.
\looseness=-1

% \todo{update title and transition}

\section{No TTA Methods Mitigate All Shifts Yet}  \label{sec:shift_type}

\begin{table*}[!t]
	\centering
	\caption{\small
		\textbf{Adaptation performance (error) of TTA methods over OOD datasets with common distribution shifts.}
		Optimal results in \resetmodel \& \notresetmodel are highlighted by \textbf{bold} and \textcolor{blue}{blue} respectively. \looseness=-1
		% , or \textcolor{blue}{\textbf{bolded blue color}} for neither \resetmodel nor \notresetmodel.
	}
	\vspace{-0.5em}
	\label{tab:common_distribution_shifts}
	\setlength{\tabcolsep}{2pt}
	\def\arraystretch{1.25}
	\resizebox{0.85\textwidth}{!}{
		\begin{tabular}{lcccccc}
			\toprule
			                            & CIFAR10-C (\%) $\downarrow$      & CIFAR100-C (\%) $\downarrow$ & ImageNet-C (\%) $\downarrow$ & CIFAR10.1 (\%) $\downarrow$      & OfficeHome (\%) $\downarrow$     & PACS (\%) $\downarrow$           \\
			\midrule
			Baseline                    & $44.3$                           & $68.7$                       & $82.4$                       & $12.8$                           & $39.2$                           & $39.5$                           \\
			\midrule
			BN\_Adapt                   & $27.5 \pm 0.1$                   & $56.5 \pm 0.1$               & $72.3 \pm 0.1$               & $19.0 \pm 0.4$                   & $39.6 \pm 0.1$                   & $27.6 \pm 0.1$                   \\
			\midrule
			SHOT-\resetmodel            & $21.6 \pm 0.0$                   & $49.2 \pm 0.1$               & $68.0 \pm 0.0$               & $11.8 \pm 0.2$                   & $35.9 \pm 0.0$                   & $\mathbf{22.0 \pm 0.1}$          \\
			SHOT-\notresetmodel         & $21.0 \pm 0.1$                   & $46.8 \pm 0.1$               & $62.4 \pm 0.0$               & $14.8 \pm 0.0$                   & \textcolor{blue}{$35.5 \pm 0.1$} & \textcolor{blue}{$17.8 \pm 0.1$} \\
			\midrule
			TTT-\resetmodel             & $\mathbf{20.9 \pm 0.4}$          & $51.8 \pm 0.2$               & -                            & $12.5 \pm 0.1$                   & $40.2 \pm 0.0$                   & $25.3 \pm 0.1$                   \\
			TTT-\notresetmodel          & \textcolor{blue}{$20.0 \pm 0.1$} & $51.9 \pm 0.1$               & -                            & $13.5 \pm 0.0$                   & $42.2 \pm 0.1$                   & $26.6 \pm 0.1$                   \\
			\midrule
			TENT-\resetmodel            & $26.9 \pm 0.0$                   & $54.6 \pm 0.1$               & $70.3 \pm 0.0$               & $18.6 \pm 0.4$                   & $38.4 \pm 0.0$                   & $26.1 \pm 0.1$                   \\
			TENT-\notresetmodel         & $21.7 \pm 0.1$                   & $49.9 \pm 0.2$               & $61.9 \pm 0.1$               & $17.9 \pm 0.2$                   & $37.6 \pm 0.0$                   & $22.7 \pm 0.2$                   \\
			\midrule
			T3A                         & $40.3 \pm 0.1$                   & $67.6 \pm 0.0$               & $83.1 \pm 0.0$               & \textcolor{blue}{$12.5 \pm 0.1$} & $\mathbf{35.7 \pm 0.1}$          & $31.0 \pm 0.4$                   \\
			\midrule
			CoTTA-\resetmodel           & $25.3 \pm 0.1$                   & $55.3 \pm 0.1$               & $94.0 \pm 0.0$               & $19.1 \pm 0.4$                   & $53.7 \pm 0.0$                   & $28.6 \pm 0.1$                   \\
			CoTTA-\notresetmodel        & $42.5 \pm 0.1$                   & $78.1 \pm 0.1$               & $94.4 \pm 0.1$               & $39.4 \pm 1.2$                   & $52.9 \pm 0.3$                   & $31.7 \pm 0.2$                   \\
			\midrule
			MEMO-\resetmodel            & $38.1 \pm 0.1$                   & $65.3 \pm 0.0$               & $81.3 \pm 0.0$               & $\mathbf{10.8 \pm 0.1}$          & $37.6 \pm 0.0$                   & $39.4 \pm 0.0$                   \\
			MEMO-\notresetmodel         & $85.2 \pm 0.7$                   & $96.3 \pm 0.2$               & $99.4 \pm 0.1$               & $14.2 \pm 1.1$                   & $91.3 \pm 0.1$                   & $75.5 \pm 0.4$                   \\
			\midrule
			NOTE-\resetmodel            & $32.4 \pm 0.0$                   & $60.0 \pm 0.0$               & $80.8 \pm 0.3$               & $12.0 \pm 0.1$                   & $37.9 \pm 0.0$                   & $32.0 \pm 0.1$                   \\
			NOTE-\notresetmodel         & $24.0 \pm 0.1$                   & $54.5 \pm 0.2$               & $69.8 \pm 0.1$               & $12.7 \pm 0.2$                   & $37.9 \pm 0.1$                   & $27.7 \pm 0.0$                   \\
			\midrule
			Conjugate PL-\resetmodel    & $26.9 \pm 0.0$                   & $54.4 \pm 0.1$               & $70.0 \pm 0.1$               & $18.7 \pm 0.3$                   & $38.0 \pm 0.1$                   & $25.3 \pm 0.1$                   \\
			Conjugate PL-\notresetmodel & $22.9 \pm 0.1$                   & $51.0 \pm 0.3$               & $62.2 \pm 0.0$               & $18.3 \pm 0.2$                   & $37.5 \pm 0.1$                   & $21.8 \pm 0.1$                   \\
			\midrule
			SAR-\resetmodel             & $24.5 \pm 0.0$                   & $54.6 \pm 0.1$               & $70.6 \pm 0.1$               & $17.1 \pm 0.2$                   & $38.1 \pm 0.0$                   & $26.2 \pm 0.1$                   \\
			SAR-\notresetmodel          & $21.9 \pm 0.1$                   & $49.7 \pm 0.1$               & $59.1 \pm 0.3$               & $18.0 \pm 0.1$                   & $37.9 \pm 0.0$                   & $22.7 \pm 0.2$                   \\
			\bottomrule
		\end{tabular}
	}
	\vspace{-10pt}
\end{table*}

The efficacy of TTA is contingent upon the nature of distributional variations. Specifically, the advantages demonstrated in previous research in the context of uncorrelated attribute shifts cannot be extrapolated to other forms of distributional shifts, such as shifts in spurious correlation, label shifts, and non-stationary shifts.
In this section, we employ two evaluation protocols previously outlined in~\cref{sec:evaluation_protocols} to re-evaluate commonly used datasets for distributional shifts, as well as benchmarks for distributional shifts that have been infrequently or never evaluated by prevalent TTA methods.
\autoref{tab:common_distribution_shifts} and \autoref{tab:realistic_distribution_shifts} summarize the results of our experiments on all benchmarks for distributional shifts. Details of evaluation setups can be found in~\cref{appendix:tta_implementation_details}.
% \autoref{appendix:additional_results} contains a comprehensive presentation of the results for each dataset and domain.

% \tao{maybe we need to highlight, old methods indeed could be very effective under a fair evaluation, like TTT and SHOT (even in intro).}
\paragraph{Common distribution shifts.}
Here our evaluation of TTA performance primarily focuses on three areas: synthetic co-variate shift (i.e.\ CIFAR10-C), natural shift (i.e.\ CIFAR10.1), and domain generalization (i.e.\ OfficeHome and PACS).
\emph{Except for \notresetmodel MEMO, all methods improve average performance across four common distributional shifts}, although the extent of the adaptation performance gain varies among different TTA methods.
Notably, \notresetmodel MEMO resulted in a significant degradation in adaptation performance, with an average test error of 66.6\%, compared to 31.5\% for episodic MEMO and 34.0\% for the baseline, indicating that MEMO is only effective in episodic adaptation settings.
Additionally, \emph{BN\_Adapt, TENT, and TTT were unable to ensure improvement in adaptation performance on more challenging and realistic distributional shift benchmarks}, such as CIFAR10.1 and OfficeHome.
It should be noted that \emph{no single method consistently outperforms the others across all datasets under our fair evaluation}.
\citet{niu2023towards} shows that batch normalization hinders stable TTA by estimating problematic mean and variance statistics, and prefers to use batch-agnostic norm layers, such as group norm~\citep{wu2018group} and layer norm~\citep{ba2016layer}.
We provide additional benchmark results on architecture designs that utilize the group norm and layer norm in~\cref{appendix:additional_results_on_normalization_layers}.

% \hao{''Beyond Invariance: Test-Time Label-Shift Adaptation for Distributions with “Spurious” Correlations'' compared their method with other UDA methods. They did not provide evaluation results on any dominant TTA methods}

\begin{table}[!t]
	\caption{\small
		\textbf{Adaptation error (in \%) of TTA methods over OOD datasets with two realistic distribution shifts.}
		% We use the same highlighting strategy as in~\autoref{tab:common_distribution_shifts}.
		Dirichlet distribution is used to create non-i.i.d.\ test streams;
		% and different trials will have different samples, resulting in varied baseline performance
		the smaller value of $\alpha$ is, the more severe the label shift will be.
		Optimal results in \resetmodel \& \notresetmodel are highlighted by \textbf{bold} and \textcolor{blue}{blue} respectively.
		%\looseness=-1
	}
	\vspace{-0.5em}
	\label{tab:realistic_distribution_shifts}
	\centering
	\setlength{\tabcolsep}{2pt}
	\def\arraystretch{1.25}
	\resizebox{.5\textwidth}{!}{
		\begin{tabular}{@{}l  c c  cccc }
			\toprule
			\multicolumn{1}{c}{}        & \multicolumn{2}{c}{\textbf{Spurious correlation shifts} $\downarrow$} & \multicolumn{3}{c}{\textbf{Label shifts} on CIFAR10 $\downarrow$}                                                                                                       \\
			\cmidrule(lr){2-3} \cmidrule(lr){4-6} \cmidrule(lr){7-7}
			                            & ColoredMNIST                                                          & Waterbirds                                                        & $\alpha\!=\!0.01$               & $\alpha\!=\!0.1$                & $\alpha\!=\!1$                  \\
			\midrule
			Baseline                    & $85.6$                                                                & $29.1$                                                            & \textcolor{blue}{$7.8 \pm 2.3$} & $5.5 \pm 1.3$                   & $6.5 \pm 0.8$                   \\
			\midrule
			BN\_adapt                   & $83.9 \pm 0.2$                                                        & $38.1 \pm 1.0$                                                    & $77.8 \pm 1.7$                  & $64.5 \pm 7.7$                  & $18.2 \pm 1.0$                  \\
			\midrule
			SHOT-\resetmodel            & $83.0 \pm 0.3$                                                        & $29.4 \pm 0.3$                                                    & $10.1 \pm 2.5$                  & $7.3 \pm 1.0$                   & $6.6 \pm 0.8$                   \\
			SHOT-\notresetmodel         & $89.7 \pm 0.2$                                                        & $27.0 \pm 0.7$                                                    & $39.1 \pm 3.1$                  & $30.0 \pm 3.3$                  & $10.7 \pm 1.0$                  \\
			\midrule
			TTT-\resetmodel             & $78.1 \pm 0.1$                                                        & $28.2 \pm 0.3$                                                    & $11.0 \pm 3.0$                  & $5.8 \pm 1.7$                   & $6.6 \pm 1.6$                   \\
			TTT-\notresetmodel          & \textcolor{blue}{$67.1 \pm 1.3$}                                      & $24.0 \pm 1.9$                                                    & $9.0 \pm 2.3$                   & $6.1 \pm 1.3$                   & $7.2 \pm 1.4$                   \\
			\midrule
			TENT-\resetmodel            & $83.9 \pm 0.2$                                                        & $37.7 \pm 1.0$                                                    & $76.8 \pm 1.9$                  & $63.3 \pm 7.1$                  & $17.6 \pm 0.8$                  \\
			TENT-\notresetmodel         & $84.3 \pm 0.2$                                                        & $24.2 \pm 0.4$                                                    & $76.3 \pm 2.1$                  & $62.2 \pm 6.5$                  & $16.2 \pm 0.4$                  \\
			\midrule
			T3A                         & $88.1 \pm 0.1$                                                        & \textcolor{blue}{$\mathbf{22.3 \pm 0.2}$}                         & $15.9 \pm 3.5$                  & $9.6 \pm 0.7$                   & $7.2 \pm 0.6$                   \\
			\midrule
			CoTTA-\resetmodel           & $\mathbf{72.6 \pm 0.2}$                                               & $31.7 \pm 0.4$                                                    & $74.7 \pm 1.7$                  & $61.1 \pm 7.4$                  & $17.0 \pm 1.2$                  \\
			CoTTA-\notresetmodel        & $87.0 \pm 0.5$                                                        & $25.5 \pm 1.5$                                                    & $80.5 \pm 2.0$                  & $70.6 \pm 5.0$                  & $31.7 \pm 5.3$                  \\
			\midrule
			MEMO-\resetmodel            & $84.9 \pm 0.1$                                                        & $34.3 \pm 0.1$                                                    & $\mathbf{0.1 \pm 0.0}$          & $\mathbf{1.2 \pm 0.9}$          & $\mathbf{4.5 \pm 0.6}$          \\
			\midrule
			NOTE-\resetmodel            & $83.5 \pm 0.1$                                                        & $30.0 \pm 0.4$                                                    & $7.9 \pm 2.3$                   & \textcolor{blue}{$5.4 \pm 1.1$} & $5.7 \pm 0.7$                   \\
			NOTE-\notresetmodel         & $83.4 \pm 0.4$                                                        & $43.3 \pm 6.3$                                                    & $9.0 \pm 2.1$                   & $6.2 \pm 1.1$                   & \textcolor{blue}{$6.4 \pm 0.7$} \\
			\midrule
			Conjugate PL-\resetmodel    & $83.9 \pm 0.2$                                                        & $37.9 \pm 0.9$                                                    & $76.9 \pm 1.9$                  & $63.8 \pm 7.4$                  & $17.6 \pm 0.8$                  \\
			Conjugate PL-\notresetmodel & $87.3 \pm 0.3$                                                        & $23.7 \pm 2.9$                                                    & $72.2 \pm 0.3$                  & $59.5 \pm 7.0$                  & $16.0 \pm 0.1$                  \\
			\midrule
			SAR-\resetmodel             & $83.9 \pm 0.2$                                                        & $37.4 \pm 1.1$                                                    & $75.3 \pm 1.5$                  & $62.0 \pm 7.7$                  & $15.5 \pm 1.0$                  \\
			SAR-\notresetmodel          & $83.9 \pm 0.2$                                                        & $34.6 \pm 0.5$                                                    & $75.8 \pm 1.4$                  & $61.1 \pm 6.7$                  & $16.0 \pm 0.6$                  \\
			\bottomrule
		\end{tabular}
	}
	\vspace{-1em}
\end{table}

\paragraph{Spurious correlation shifts.}
To the best of our knowledge, this study represents the first examination of the efficacy of dominant TTA methods in addressing spurious correlation shifts as demonstrated in the ColoredMNIST and Waterbirds benchmarks.
As shown in~\autoref{tab:realistic_distribution_shifts}, while some TTA methods demonstrate a reduction in error rate compared to the baseline, \emph{none of TTA methods can improve performance on the ColoredMNIST benchmark}, as even a randomly initialized model exhibits a 50\% error rate on this dataset.
\emph{In terms of addressing the spurious correlation shift in the Waterbirds dataset, only T3A and TTT can consistently improve adaptation performance}, as measured by worst-group error.
TENT and SHOT may potentially improve performance on Waterbirds, but only through the utilization of impractical model selection techniques.
The adaptation results presented in~\cref{appendix:additional_results}, are obtained through the use of commonly accepted practices in terms of hyperparameter choices, and adhere to the evaluation protocol established in previous research.

\paragraph{Label shifts.}
\citet{boudiaf2022parameter} and \citet{gong2022note} have taken label shift into account in their research, but they paired it with co-variate shift on CIFAR10-C.
In contrast, our work solely examines the effectiveness of various TTA methods in addressing label shifts on the CIFAR10 dataset.
The experimental results indicate that \emph{all TTA methods, except the MEMO method, demonstrate a higher test error than the baseline under strong label shift conditions.}
Specifically, TTA methods that heavily rely on the test batch for recalculating Batch Normalization statistics, such as TENT and BN\_Adapt, experience the most significant performance degradation, with BN\_Adapt incurring a 77.8\% test error and TENT experiencing over 76.0\% error rate when the label shift parameter $\alpha$ is set to $0.01$.

% \begin{wraptable}{r}{0.25\textwidth}
\begin{table}[th]
	\centering
	\caption{\small
		\textbf{Adaptation performance (error in \%) of TTA methods on continual distribution shifts.} To make a fair comparison, we employ Batch Normalization (BN) layer and use the same checkpoint with the other methods in NOTE-\resetmodel and NOTE-\notresetmodel.
		We reproduce the original implementation (with Instand-aware BN) and pretrain another base model in NOTE-\notresetmodel $\star$.
	}
	\vspace{-0.5em}
	\resizebox{0.325\textwidth}{!}{
		\begin{tabular}{l c c}
			\toprule
			\multicolumn{1}{c}{}        & \multicolumn{2}{c}{\textbf{Cont. dist. shifts}}                           \\
			\cmidrule(lr){2-3}
			                            & CIFAR10-C                                       & ImageNet-C              \\
			% & \parbox{2.4cm}{\centering Cont. dist.\ shifts \\ (CIFAR10-C) }   \\
			\midrule
			Baseline                    & $44.3$                                          & $82.4$                  \\
			\midrule
			BN\_adapt                   & $79.9 \pm 0.5$                                  & $96.3 \pm 0.7$          \\
			\midrule
			SHOT-\resetmodel            & $41.3 \pm 0.1$                                  & $80.8 \pm 0.1$          \\
			SHOT-\notresetmodel         & $51.2 \pm 2.0$                                  & $93.5 \pm 2.1$          \\
			\midrule
			TTT-\resetmodel             & $27.8 \pm 0.1$                                  & -                       \\
			TTT-\notresetmodel          & $29.7 \pm 0.9$                                  & -                       \\
			\midrule
			TENT-\resetmodel            & $79.2 \pm 0.4$                                  & $95.5 \pm 0.6$          \\
			TENT-\notresetmodel         & $79.6 \pm 0.4$                                  & $97.5 \pm 0.6$          \\
			\midrule
			T3A                         & $43.2 \pm 0.3$                                  & $82.2 \pm 1.1$          \\
			\midrule
			CoTTA-\resetmodel           & $76.0 \pm 0.4$                                  & $97.8 \pm 0.6$          \\
			CoTTA-\notresetmodel        & $82.6 \pm 0.3$                                  & $98.5 \pm 0.8$          \\
			\midrule
			MEMO-\resetmodel            & $\mathbf{12.7} \pm 0.1$                         & $\mathbf{70.7} \pm 0.5$ \\
			\midrule
			NOTE-\resetmodel            & $39.2 \pm 0.1$                                  & $81.8 \pm 0.5$          \\
			NOTE-\notresetmodel         & $25.7 \pm 0.1$                                  & $72.2 \pm 1.3$          \\
			\midrule
			Conjugate PL-\resetmodel    & $79.3 \pm 0.4$                                  & $95.4 \pm 0.6$          \\
			Conjugate PL-\notresetmodel & $79.6 \pm 0.4$                                  & $98.5 \pm 0.5$          \\
			\midrule
			SAR-\resetmodel             & $77.2 \pm 0.5$                                  & $95.4 \pm 0.6$          \\
			SAR-\notresetmodel          & $79.6 \pm 0.4$                                  & $97.2 \pm 0.4$          \\
			\midrule
			NOTE-\notresetmodel $\star$ & $21.8 \pm 0.0$                                  & -                       \\
			\bottomrule
		\end{tabular}
	}
	\label{tab:continual_distribution_shifts}
	\vspace{-1em}
\end{table}
% \todo{update table}

\paragraph{Non-stationary shifts.}
In~\autoref{tab:continual_distribution_shifts} we report the adaptation performance of TTA methods on the temporally correlated CIFAR10-C dataset introduced in~\citet{gong2022note}. Additionally, we reproduce NOTE in TTAB, which is the current SOTA in the benchmark of temporal correlated shifts.
Our results indicate that, even with the appropriate model selection, TENT and BN\_Adapt still fail to improve adaptation performance in the presence of non-stationary shifts.
However, TTA methods (\textit{e.g.}, TTT and MEMO) demonstrate substantial performance gains when adapting to the temporally correlated test stream, likely due to their instance-aware adaptation strategies, which focus on individual test samples. Surprisingly, MEMO outperforms NOTE in our implementation, which demonstrates the necessity of proper model selection in the field.

\section{Conclusion}

We have presented TTAB, a large-scale open-sourced benchmark for test-time adaptation. Through thorough and systematic studies, we showed that current TTA methods fall short in three aspects critical for practical applications, namely the difficulty in selecting appropriate hyper-parameters due to batch dependency, significant variability in performance sensitive to the quality of the pre-trained model, and poor efficacy in the face of certain classes of distribution shifts. We hope the proposed benchmark will stimulate more rigorous and measurable progress in future test-time adaptation research.

\section*{Acknowledgement}
We thank anonymous reviewers for their constructive and helpful reviews.
This work was supported in part by the National Key R\&D Program of China (Project No.\ 2022ZD0115100), the Research Center for Industries of the Future (RCIF) at Westlake University, Westlake Education Foundation, and the Swiss National Science Foundation under Grant 2OOO21-L92326.

\clearpage

\bibliography{paper}
\bibliographystyle{configuration/icml2023}

% \end{document}

\clearpage
\appendix

% !TeX root = icml2023_ttab.tex
\onecolumn
{
	\hypersetup{linkcolor=black}
	\parskip=0em
	\renewcommand{\contentsname}{Contents of Appendix}
	\tableofcontents
	\addtocontents{toc}{\protect\setcounter{tocdepth}{3}}
}

\section{Messages} \label{appendix:messages}
We summarize some key messages of the manuscript here.

\begin{Summary}[title=\textbf{Limit 1: unfair evaluation in TTA}]{}{}
	\begin{itemize}[nosep,leftmargin=0pt]
		\item Methods are evaluated under distinct model statuses and experimental setups, e.g.,
		      \begin{enumerate}[nosep,leftmargin=12pt]
			      \item model quality used for the adaptation
			      \item pretraining procedure
			      \item optimizer used for the adaptation
			      \item learning rate
			      \item \# of the adaptation steps per test mini-batch
			      \item size of the test min-batch
			      \item online v.s.\ offline adaptation
			      \item w/ v.s.\ w/o resetting model (episodic v.s.\ online)
		      \end{enumerate}
		\item Methodology designs are biased to some specific neural architectures, and TTA methods cannot be fairly evaluated over various neural architectures;
	\end{itemize}
\end{Summary}

\begin{Summary}[title=\textbf{Limit 2: pitfalls of model selection in TTA}]{}{}
	\begin{itemize}[leftmargin=0pt]
		\item due to the lack of validation set and label information during test time.
		\item batch-dependency issue emerged in the streaming test mini-batches makes the oracle model selection method challenging\footnote{note that the domain generalization field only starts to examine the time-varying scenarios very recently~\citep{yao2022wild}}.
	\end{itemize}
\end{Summary}

\begin{Summary}[title=\textbf{Take-away messages}]{}{}
	\begin{itemize}[leftmargin=0pt]
		\item Improper evaluation in TTA methods.
		      Hyperparameters have a strong influence on the effectiveness of TTA, and yet they are exceedingly difficult to choose in practice without prior knowledge of the properties and structures of distribution shifts.
		      Even when the labels of test examples are available, selecting the TTA hyperparameters for model selection remains challenging, largely due to batch dependency during online adaptation.
		\item Batch dependency is a significant issue restricting the performance of \notresetmodel TTA methods.
		      Tackling the batch dependency issue of TTA methods or enabling effective model selection methods is beyond the scope of this manuscript and we leave it to the whole community for future work.
		\item Pre-trained model quality matters for TTA methods.
		      Even if hyperparameters are optimally selected given oracle information in the test domain, the effectiveness of TTA is not equal on different models.
		      The degree of improvement strongly depends on the quality of the pre-trained model, not only on its accuracy in the source domain but also on its feature properties.
		      Good practice in data augmentations~\citep{hendrycks2019augmix,hendrycks2022robustness} for out-of-distribution generalization leads to reverse effects for TTA.
		\item The community of TTA needs a comprehensive benchmark such as TTAB to guard effective progress.
		      For example, even under ideal conditions where optimal hyperparameters are used in conjunction with suitable pre-trained models, existing methods still perform poorly on certain classes of distribution shifts, such as correlation shifts~\citep{sagawa2019distributionally} and label shifts~\citep{sun2022beyond})
	\end{itemize}
\end{Summary}

\section{The Limits of Evaluation for TTA Methods}
\label{appendix:limits_of_evaluation}

\subsection{Recent Regularization Techniques Proposed to Resist Batch Dependency Problem} \label{appendix:batch_dependency_regularization}
\begin{figure}[!h]
	\centering
	\subfigure[\small Fisher regularizer]{
		\includegraphics[width=0.4\textwidth]{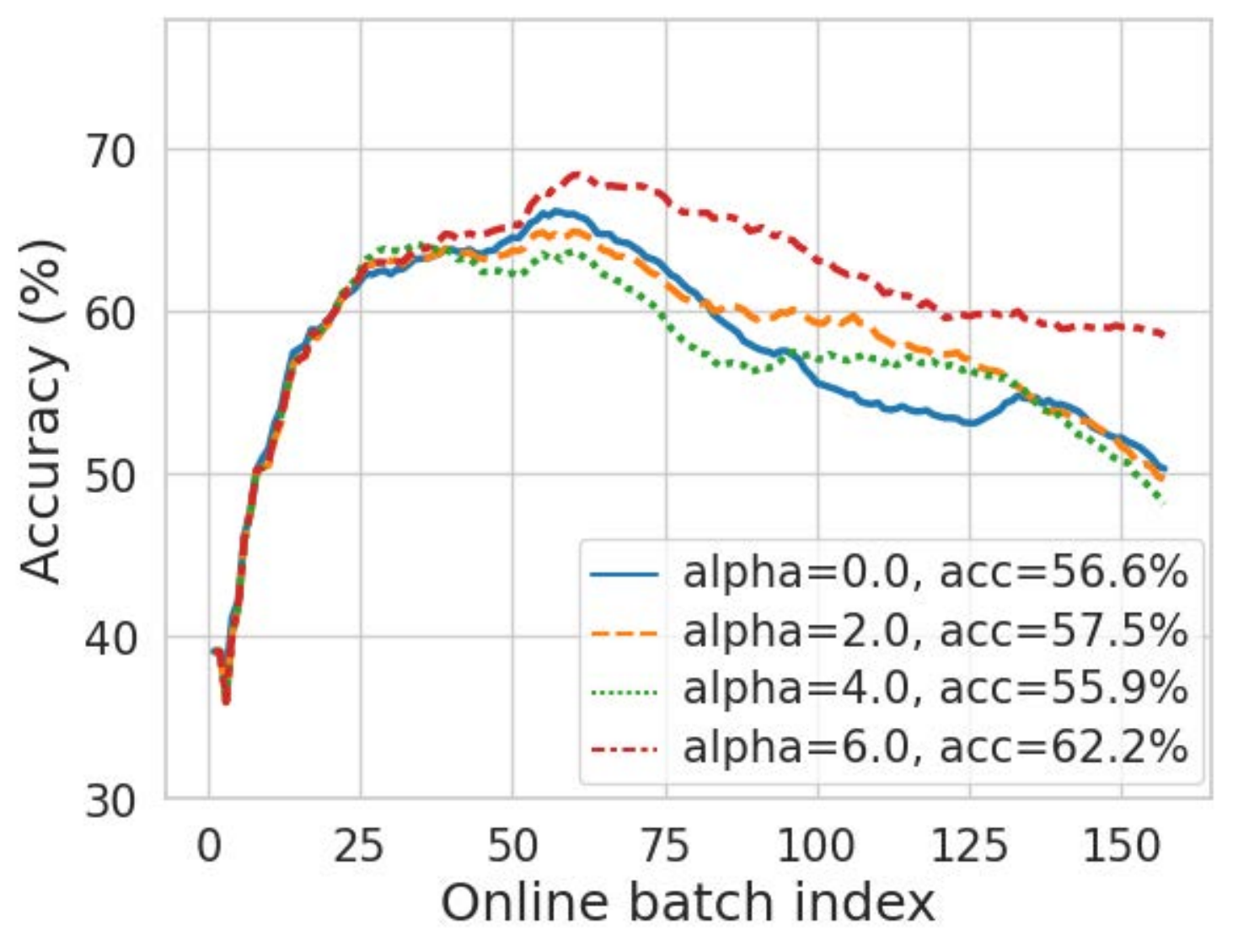}
		\label{fig:fisher_regularizer_lineplots}
	}
	\subfigure[\small Stochastic restoring]{
		\includegraphics[width=0.4\textwidth]{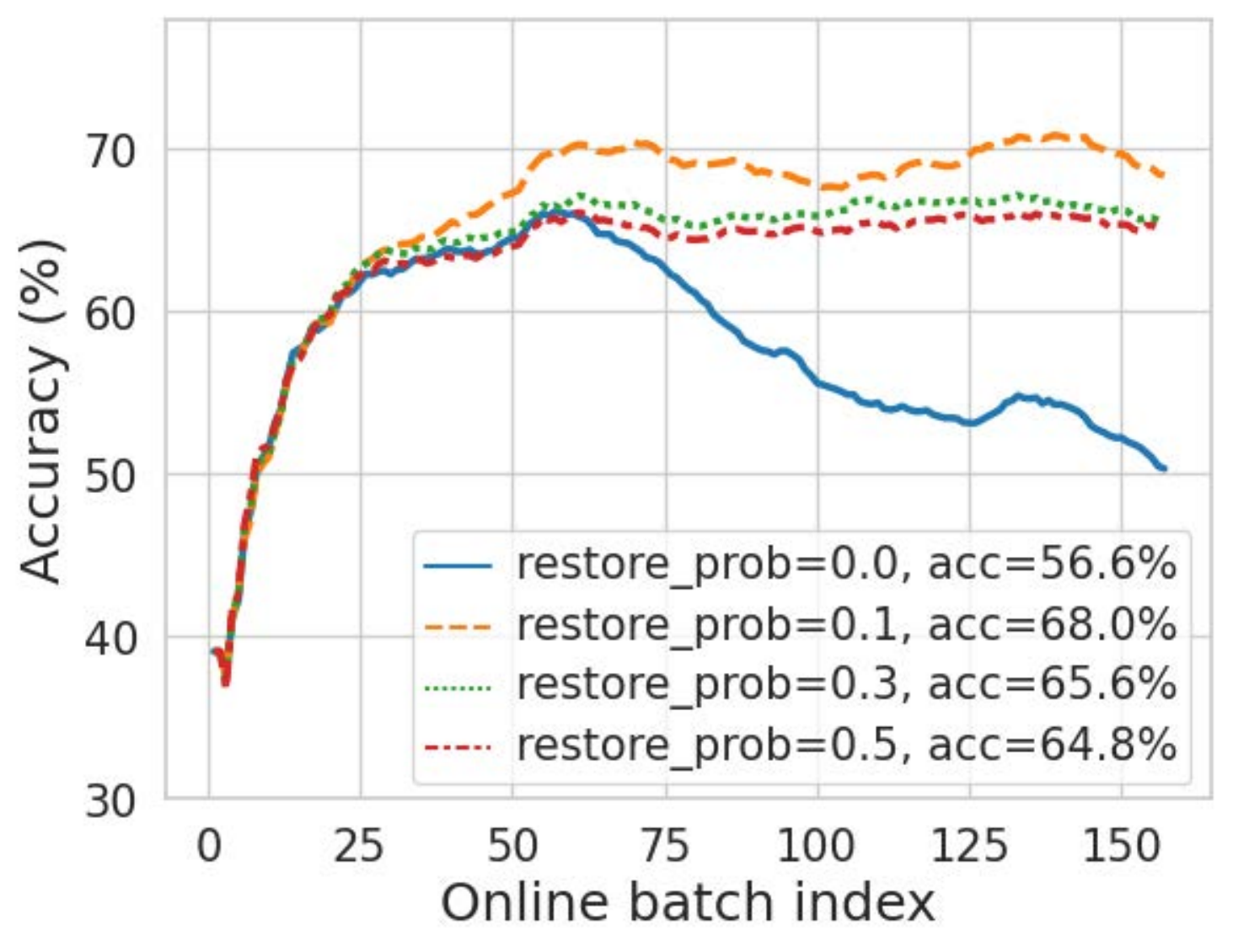}
		\label{fig:stochastic_restore_lineplots}
	}
	\caption{On the effect of fisher regularizer and stochastic restoring on batch dependency problem.}
	\label{fig:regularization_techniques}
\end{figure}

\begin{figure*}[!t]
	\centering
	\subfigure[\small Stochastic restoring: 1 steps]{
		\includegraphics[width=0.325\textwidth]{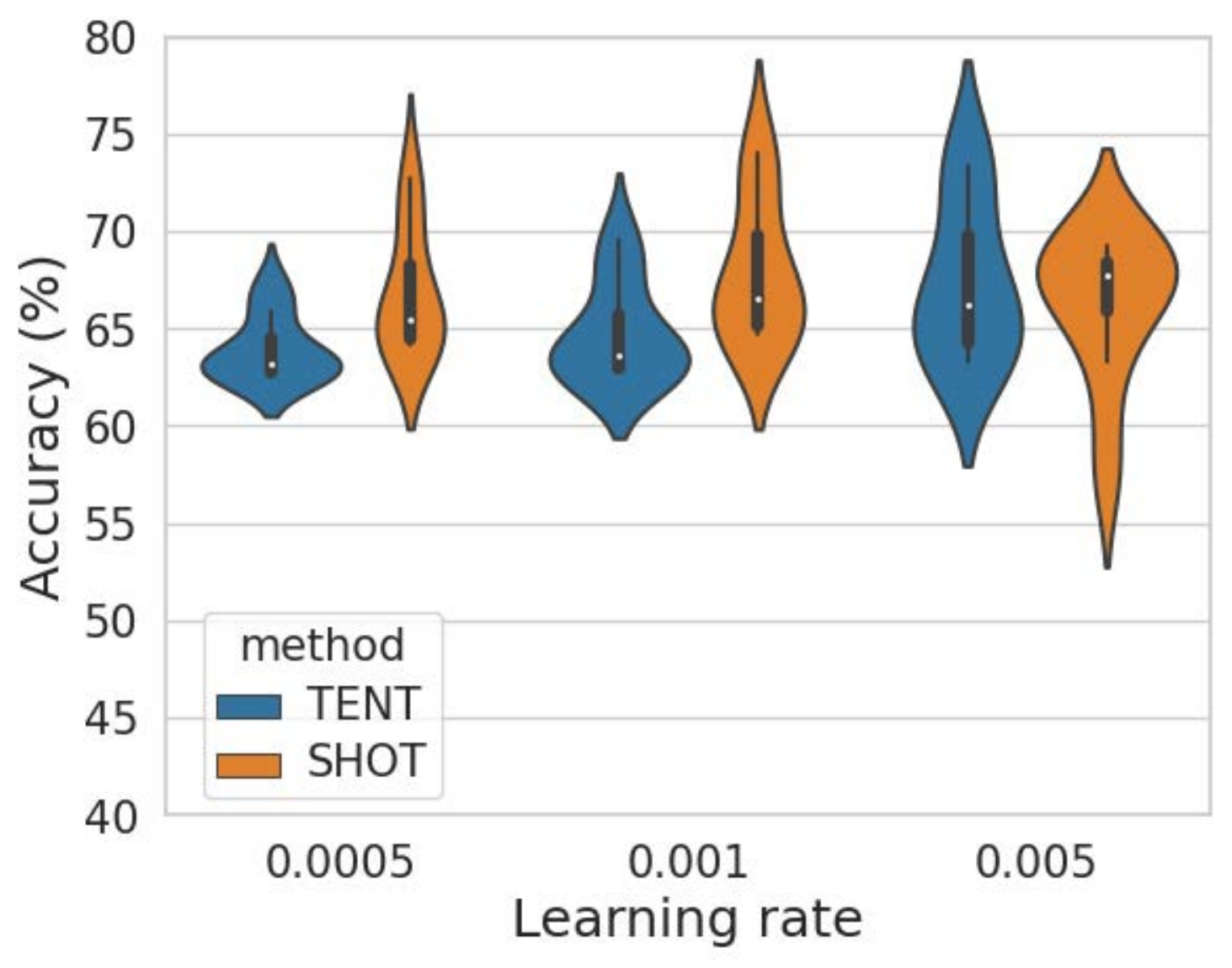}
		\label{fig:stochastic_restore}
	}
	\subfigure[\small Fisher regularizer: 1 steps]{
		\includegraphics[width=0.325\textwidth]{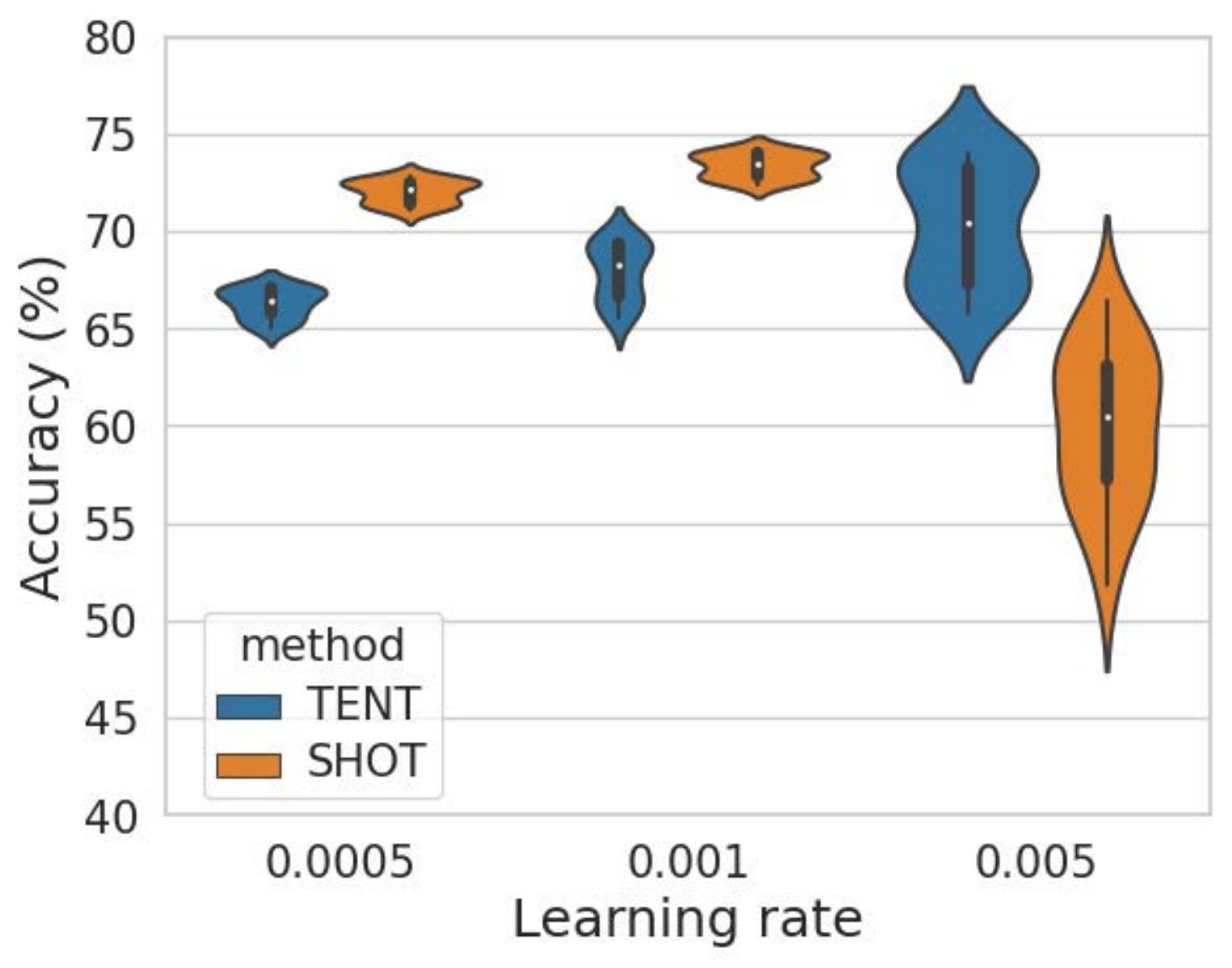}
		\label{fig:fisher_sregularizer}
	}
	\subfigure[\small Stochastic restoring: 2 steps]{
		\includegraphics[width=0.325\textwidth]{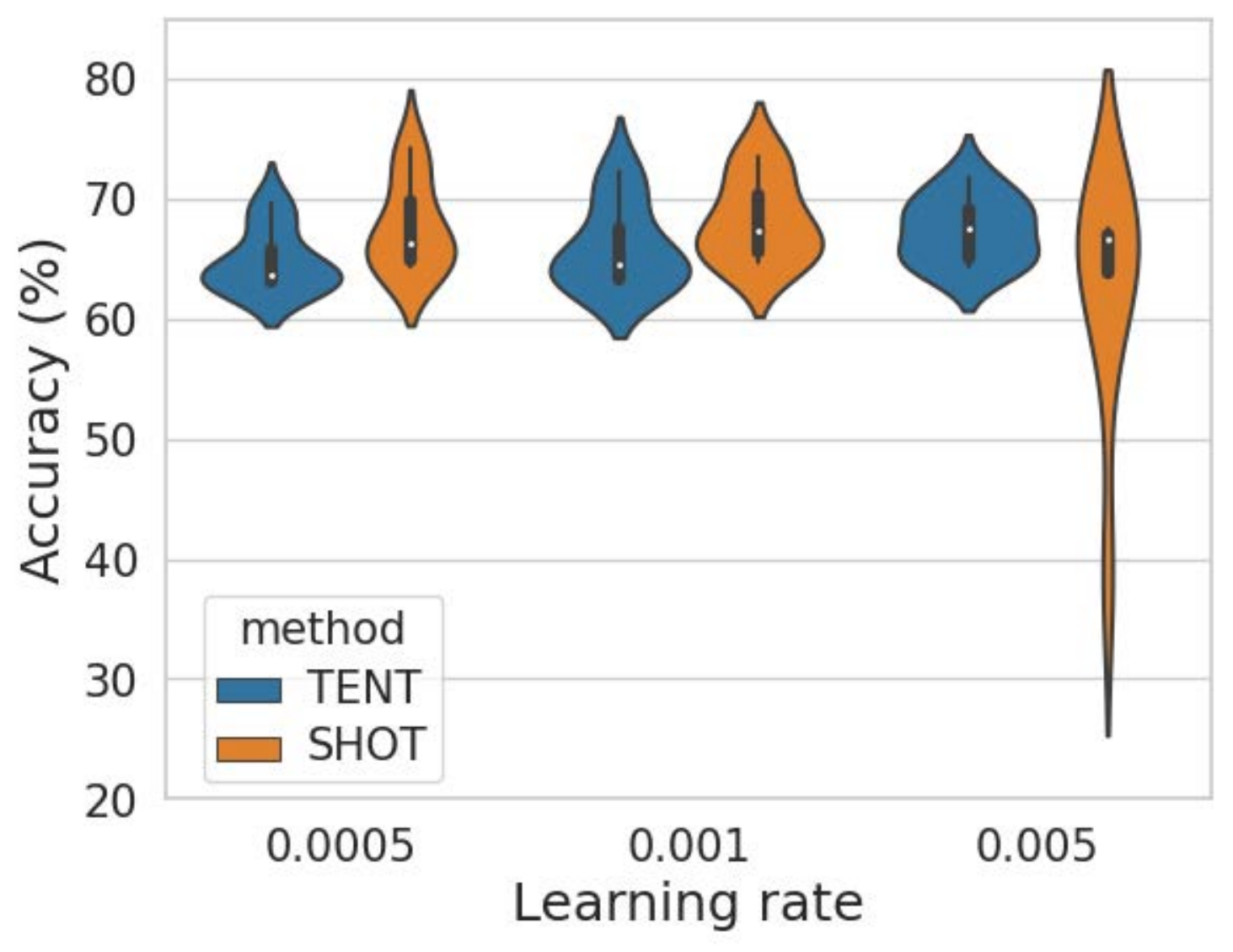}
		\label{fig:stochastic_restore_2steps}
	}
	\subfigure[\small Fisher regularizer: 2 steps]{
		\includegraphics[width=0.325\textwidth]{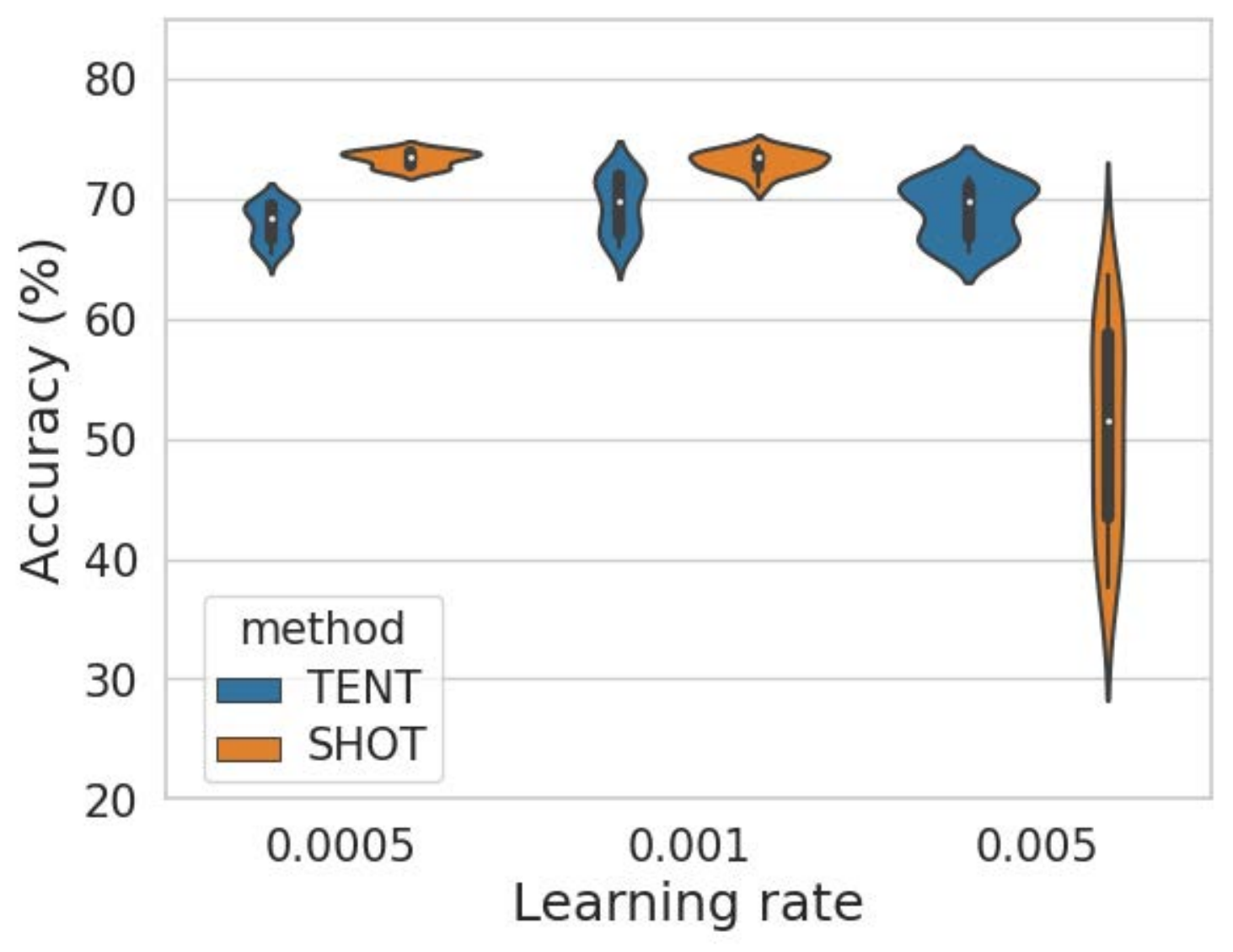}
		\label{fig:fisher_regularizer_2steps}
	}
	\subfigure[\small Stochastic restoring: 3 steps]{
		\includegraphics[width=0.325\textwidth]{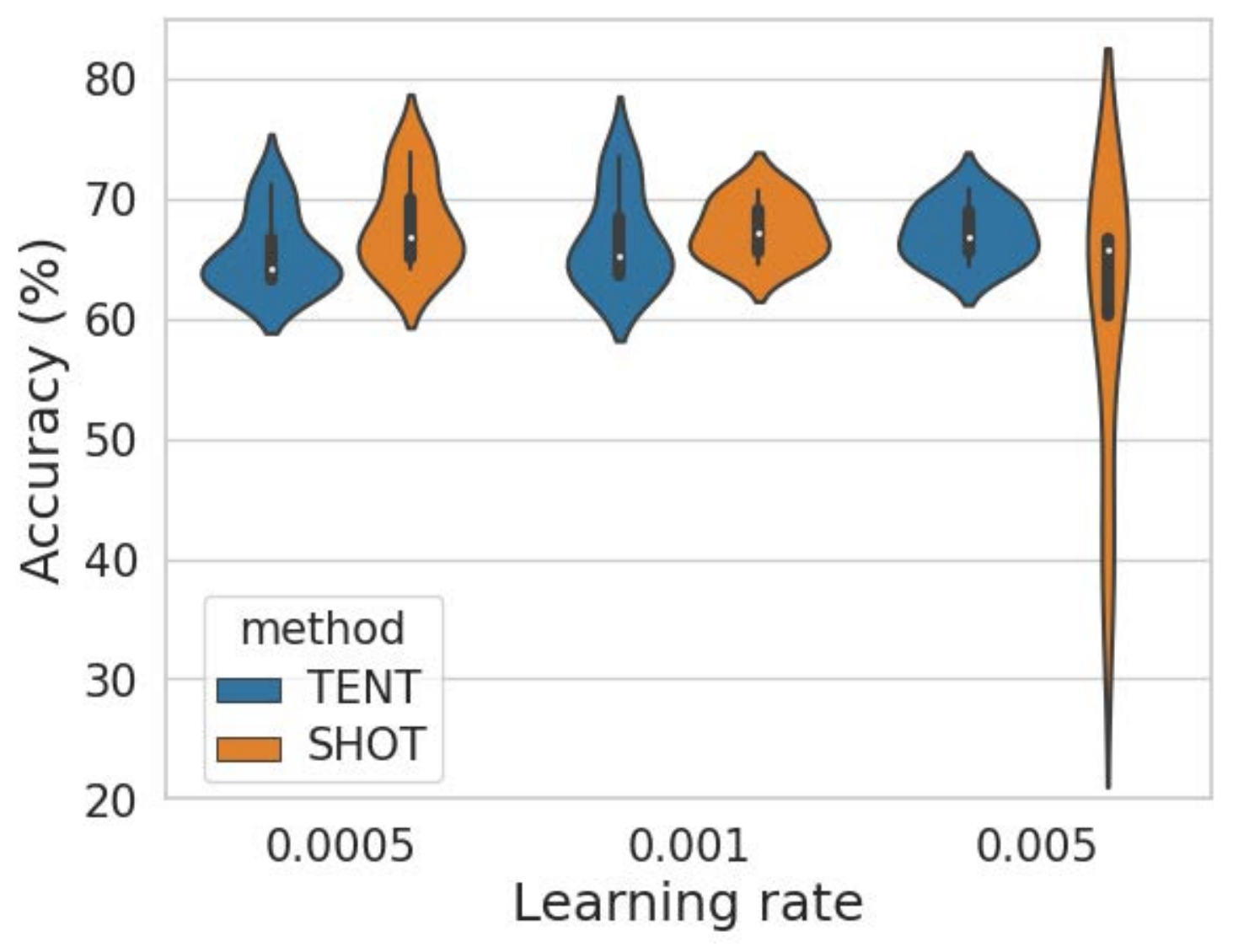}
		\label{fig:stochastic_restore_3steps}
	}
	\subfigure[\small Fisher regularizer: 3 steps]{
		\includegraphics[width=0.325\textwidth]{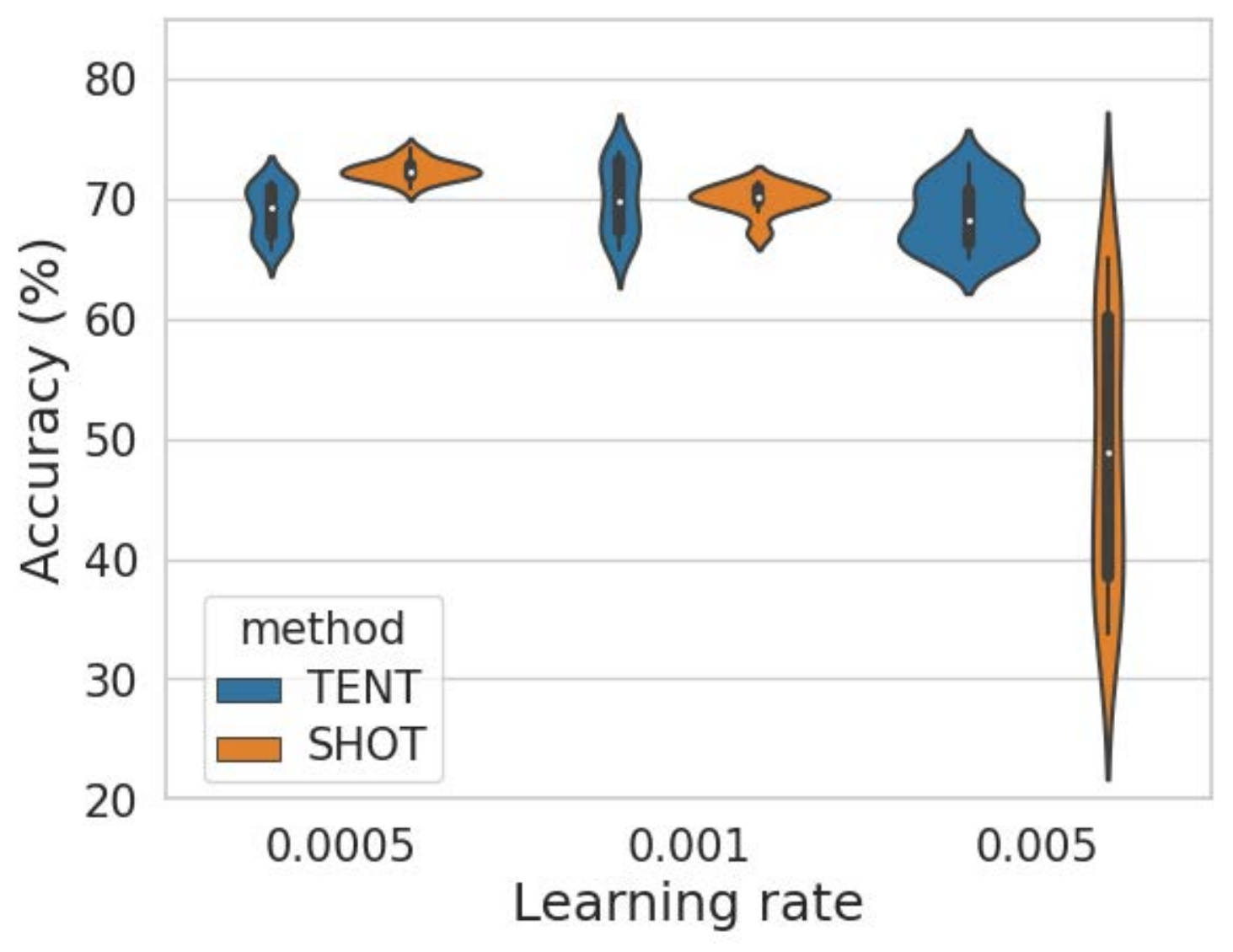}
		\label{fig:fisher_regularizer_3steps}
	}
	\subfigure[\small Stochastic restoring: 4 steps]{
		\includegraphics[width=0.325\textwidth]{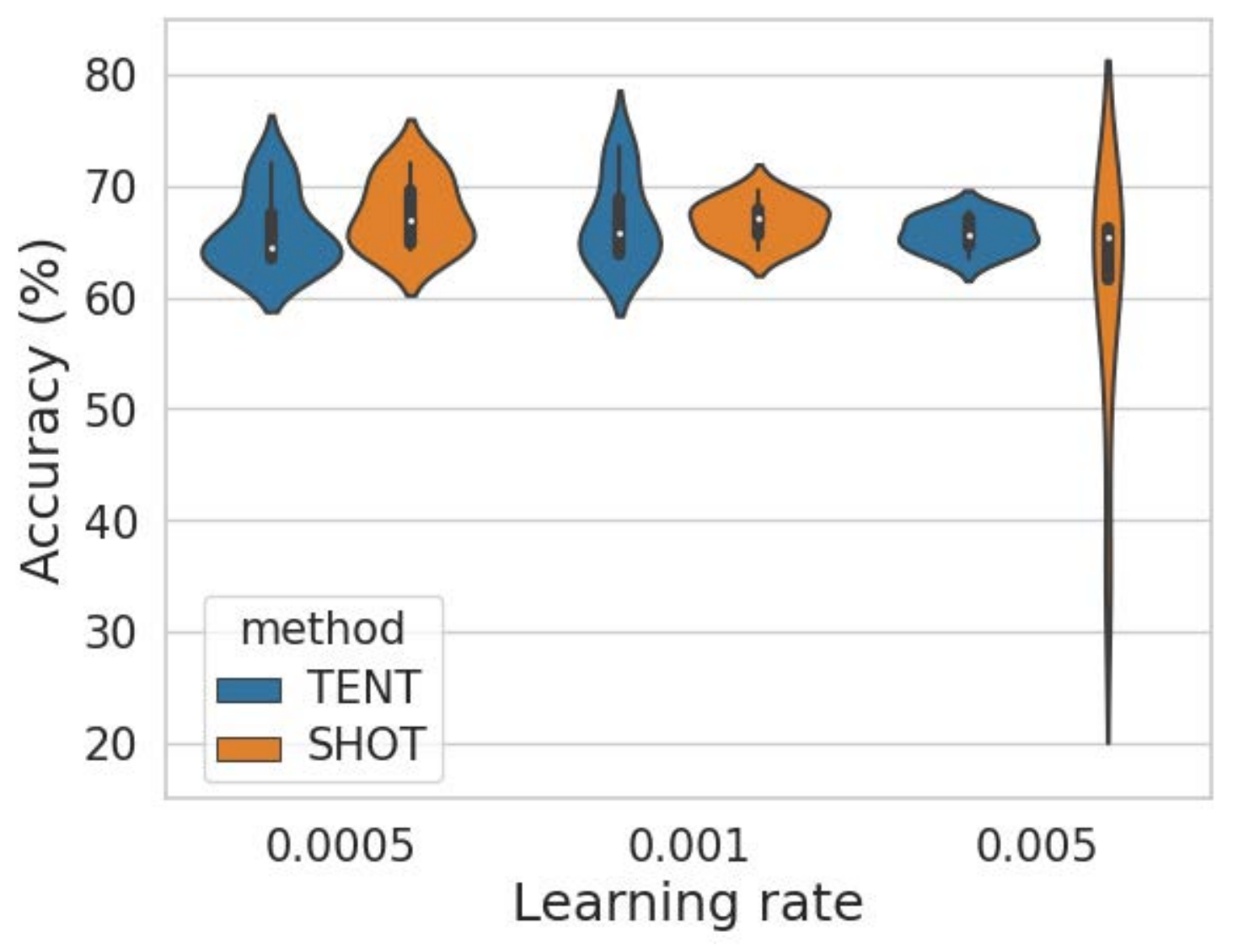}
		\label{fig:stochastic_restore_4steps}
	}
	\subfigure[\small Fisher regularizer: 4 steps]{
		\includegraphics[width=0.325\textwidth]{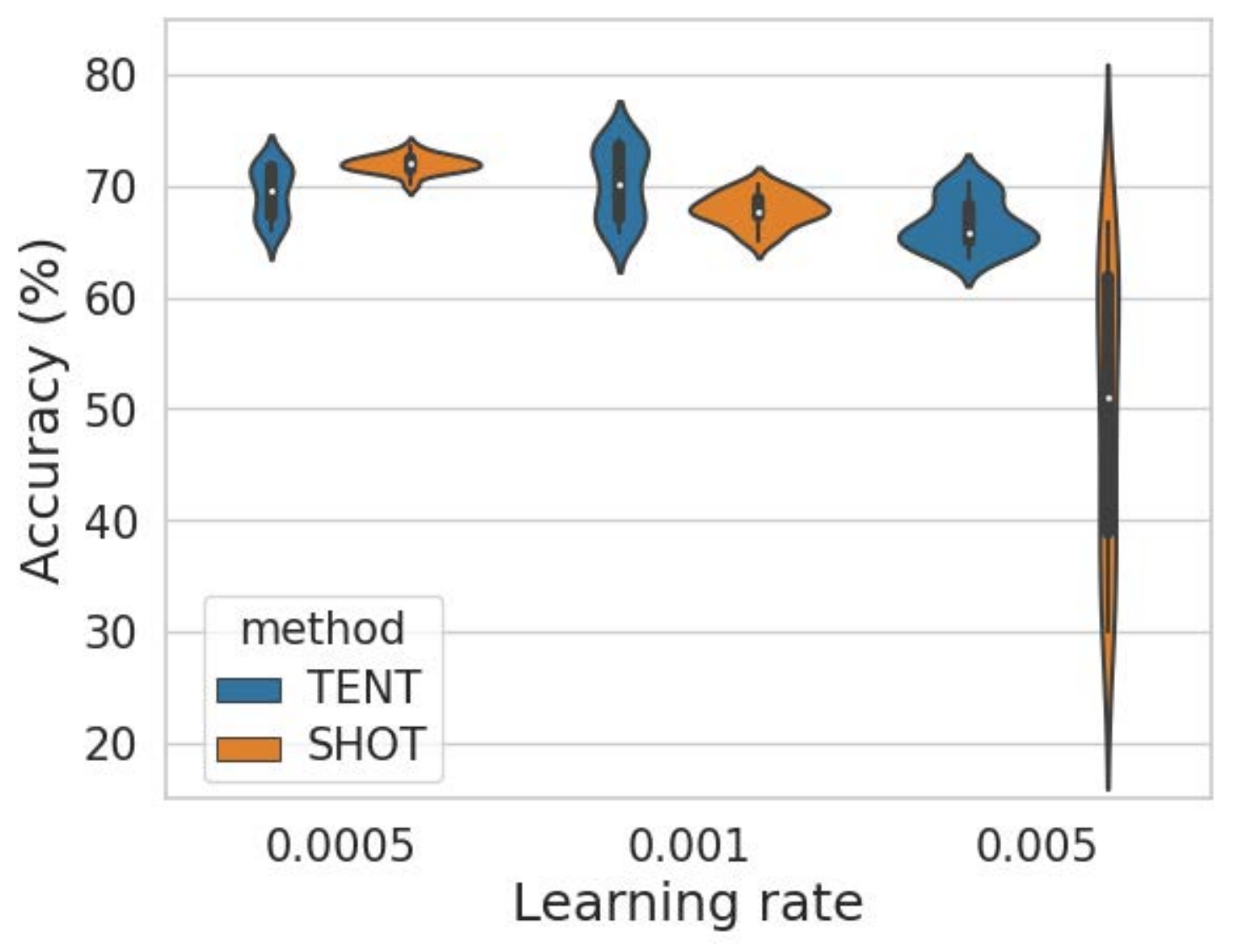}
		\label{fig:fisher_regularizer_4steps}
	}
	\vspace{-1em}
	\caption{
		\textbf{The standard deviation of stochastic restoring and Fisher regularizer when considering multiple adaptation steps}.
		Fisher regularizer~\citep{niu22efficient} aims to constrain important model parameters from drastic changes to alleviate the error accumulated due to batch dependency.
		Stochastically restoring~\citep{wang2022continual} involves a small portion of model parameters to their pre-trained values after adaptation on each test batch to prevent catastrophic forgetting.
		The hyperparameter tuning for these two techniques is challenging due to the high degree of variability inherent in these methods, which might impede their practical utility, particularly when compounded by the issue of batch dependency.
	}
	\label{fig:new_methods_under_multiple_steps}
\end{figure*}

\begin{itemize}
	\item On the influence of batch dependency problem as shown in~\autoref{fig:regularization_techniques}
	\item Stochastic restoring model parameters and Fisher regularizer still show large variance when considering multiple adaptation steps as shown in~\autoref{fig:new_methods_under_multiple_steps}.
\end{itemize}

\subsection{Optimal Model Selection for TTA is Non-trivial}\label{appendix:batch_dependency_in_tent_note}
\begin{figure}[!h]
	\centering
	\includegraphics[width=0.8\textwidth]{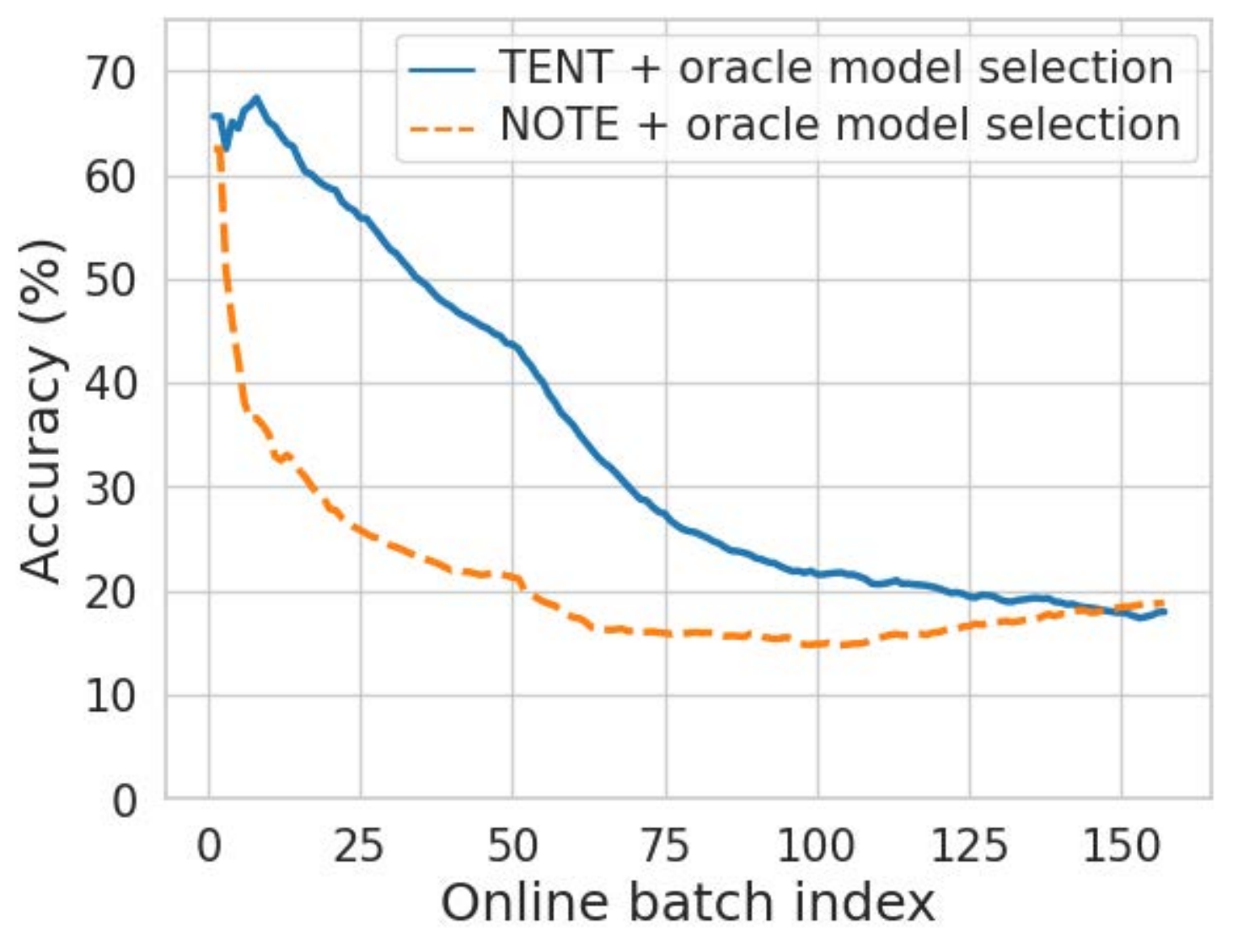}
	\caption{Oracle model selection also fails in TENT and NOTE under the online setting. Here we use ResNet-26 as the base model and learning rate is equal to 0.005.}
	\label{fig:batch_dependency_in_tent_note}
\end{figure}

Oracle model selection protocol also fails to solve the batch dependency issue in TENT and NOTE as shown in~\autoref{fig:batch_dependency_in_tent_note}
% ResNet26. learning rate = 0.005. Oracle model selection, maximum steps: 50.
% Given access to true labels, multiple TTA methods like TENT and NOTE still suffer from batch dependency problem as shown in~\autoref{fig:batch_dependency_in_tent_note}

\section{Implementation Details} \label{appendix:implementation_details}
\subsection{Implementation Details of TTA Methods} \label{appendix:tta_implementation_details}
Following prior work~\cite{gulrajani2021in,sun2020test,wang2022continual}, we use ResNet-18/ResNet-26/ResNet-50 as the base model on ColoredMNIST/CIFAR10-C/large-scale image datasets and always choose SGDm as the optimizer. We choose method-specific hyperparameters following prior work.
Following~\citet{iwasawa2021test}, we assign the pseudo label in SHOT if the predictions are over a threshold which is 0.9 in our experiment and utilize $\beta=0.3$ for all experiments except $\beta=0.1$ for ColoredMNIST just as \citet{liang2020we}.
We set the number of augmentations $B=32$ for small-scale images (e.g.\ CIFAR10-C, CIFAR100-C) and $B=64$ for large-scale image sets like ImageNet-C, becasue this is the default option in \citet{sun2020test} and \citet{zhang2022memo}.
We simply set $N=0$ that controls the trade-off between source and estimated target statistics because it achieves performance comparable to the best performance when using a batch size of 64 according to \citet{schneider2020improving}.
Training-domain validation data is used to determine the number of supports to store in T3A following \citet{iwasawa2021test}. We keep the average performance on the dataset if it has multiple test domains (e.g., CIFAR10-C, OfficeHome) and calculate the standard deviation over three different trials \{2022, 2023, 2024\}. We always examine the highest severity of corrupted data throughout our study.
% \tao{(need to add some refs), and maybe we can have some additional results in our appendix to argue that they are less sensitive than our considered hyper-parameters.}

\subsection{Implementation Details of TTAB Methods} \label{appendix:ttab_implementation_details}
To establish a consistent and realistic evaluation framework for TTA methods, we have implemented several key choices.
\circled{1} In contrast to the inconsistent pre-training strategies employed in previous studies, we have adopted a self-supervised learning approach utilizing the rotation prediction task as an auxiliary head, in conjunction with standard data augmentation techniques.
This allows us to include TTT variants and maintain a consistent level of model quality across different TTA methods.
\circled{2} For TTA methods that adapt a single image at a time (such as MEMO and TTT), we have modified the optimization procedure to accommodate larger batch sizes.
Specifically, we have fixed the model parameters and accumulated gradients computed for each sample in a batch, only updating the model parameters once all samples have been adapted in a batch.
Such a design excludes the unfairness caused by varied mini-batch sizes.
\circled{3} We have utilized Stochastic Gradient Descent with momentum for TTA throughout all experiments conducted in this work (see the discrepancy in~\autoref{tab:unfair_comparison}).
\looseness=-1

\section{Datasets} \label{appendix:datasets}
TTAB includes downloaders and loaders for all image classification tasks considered in our work:
\begin{itemize}
	\item \textbf{ColoredMNIST}~\cite{arjovsky2019invariant} is a variant of the MNIST handwritten digit classification
	      dataset. Domain $d \in \{0.1, 0.3, 0.9\}$ contains a disjoint set of digits colored either
	      red or blue. The label is a noisy function of the digit and color, such that color bears correlation $d$
	      with the label and the digit bears correlation 0.75 with the label. This dataset contains 70000
	      examples of dimension $(2, 28, 28)$ and 2 classes.
	\item \textbf{OfficeHome}~\cite{venkateswara2017deep} comprises four domains $d \in$ \{ art, clipart, product, real \}. This
	      dataset contains 15,588 examples of dimension $(3, 224, 224)$ and 65 classes.
	\item \textbf{PACS}~\cite{li2017deeper} comprises four domains $d \in$ \{ art, cartoons, photos, sketches \}. This
	      dataset contains 9,991 examples of dimension $(3, 224, 224)$ and 65 classes.
	\item \textbf{CIFAR10}~\cite{krizhevsky2009learning} consists of 60000 32x32 colour images in 10 classes, with 6000 images per class. There are 50000 training images and 10000 test images.
	\item \textbf{CIFAR10-C}~\cite{hendrycks2018benchmarking} is a dataset generated by adding 15 common corruptions + 4 extra corruptions to the test images in the Cifar10 dataset.
	\item \textbf{CIFAR10.1}~\cite{recht2018cifar10.1} contains roughly 2,000 new test images that were sampled after multiple years of research on the original CIFAR-10 dataset. The data collection for CIFAR-10.1 was designed to minimize distribution shift relative to the original dataset.
	\item \textbf{Waterbirds}~\cite{sagawa2019distributionally} is constructed by cropping out birds from photos in the Caltech-UCSD Birds-200-2011 (CUB) dataset and transferring them onto backgrounds from the Places dataset.
\end{itemize}

% \section{Multiple Adaptation Steps} \label{appendix:multiple_adaptation_steps}
% \subsection{The Effect of Considering Multiple Adaptation Steps}

% In~\autoref{fig:effect_of_adaptation_steps}, we show that it's necessary to consider tunning adaptation steps in TTA.

% \begin{figure}[!h]
% 	\centering
% 	\subfigure[]{
% 		\includegraphics[width=0.6\textwidth]{figures/effect_of_adaptation_steps.pdf}
% 	}
% 	\caption{
% 		\textbf{Adaptation performance on CIFAR10-C when using different combinations of learning rates and adaptation steps.}
% 	}
% 	\label{fig:effect_of_adaptation_steps}
% \end{figure}

% \subsection{Justify How Many Adaptation Steps are Sufficient for Oracle Model Selection}
% \begin{figure}[!h]
% 	\centering
% 	\subfigure[]{
% 		\includegraphics[width=0.8\textwidth]{figures/placeholder.pdf}
% 	}
% 	\caption{
% 		Maximum number of adaptation steps per test batch vs. overall adaptation performance in terms of top-1 accuracy (\%) on CIFAR10-C.
% 	}
% 	\label{fig:justity_maximum_adaptation_steps}
% \end{figure}

\section{Model Quality} \label{appendix:model_quality}

\subsection{On the Influence of Data Augmentation} \label{appendix:data_augmentation}
\begin{figure*}[!h]
	\centering
	\subfigure[\small BN\_Adapt]{
		\includegraphics[width=0.3\textwidth]{figures/model_quality/rn26_bn_adapt_ood_before_and_after_tta.pdf}
		\label{fig:data_aug_rn26_bn_adapt_appendix}
	}
	\subfigure[\small TENT]{
		\includegraphics[width=0.3\textwidth]{figures/model_quality/rn26_tent_episodic_ood_before_and_after_tta.pdf}
		\label{fig:data_aug_rn26_tent_appendix}
	}
	\subfigure[\small SHOT]{
		\includegraphics[width=0.3\textwidth]{figures/model_quality/rn26_shot_episodic_ood_before_and_after_tta.pdf}
		\label{fig:data_aug_rn26_shot_appendix}
	}
	\subfigure[\small BN\_Adapt]{
		\includegraphics[width=0.3\textwidth]{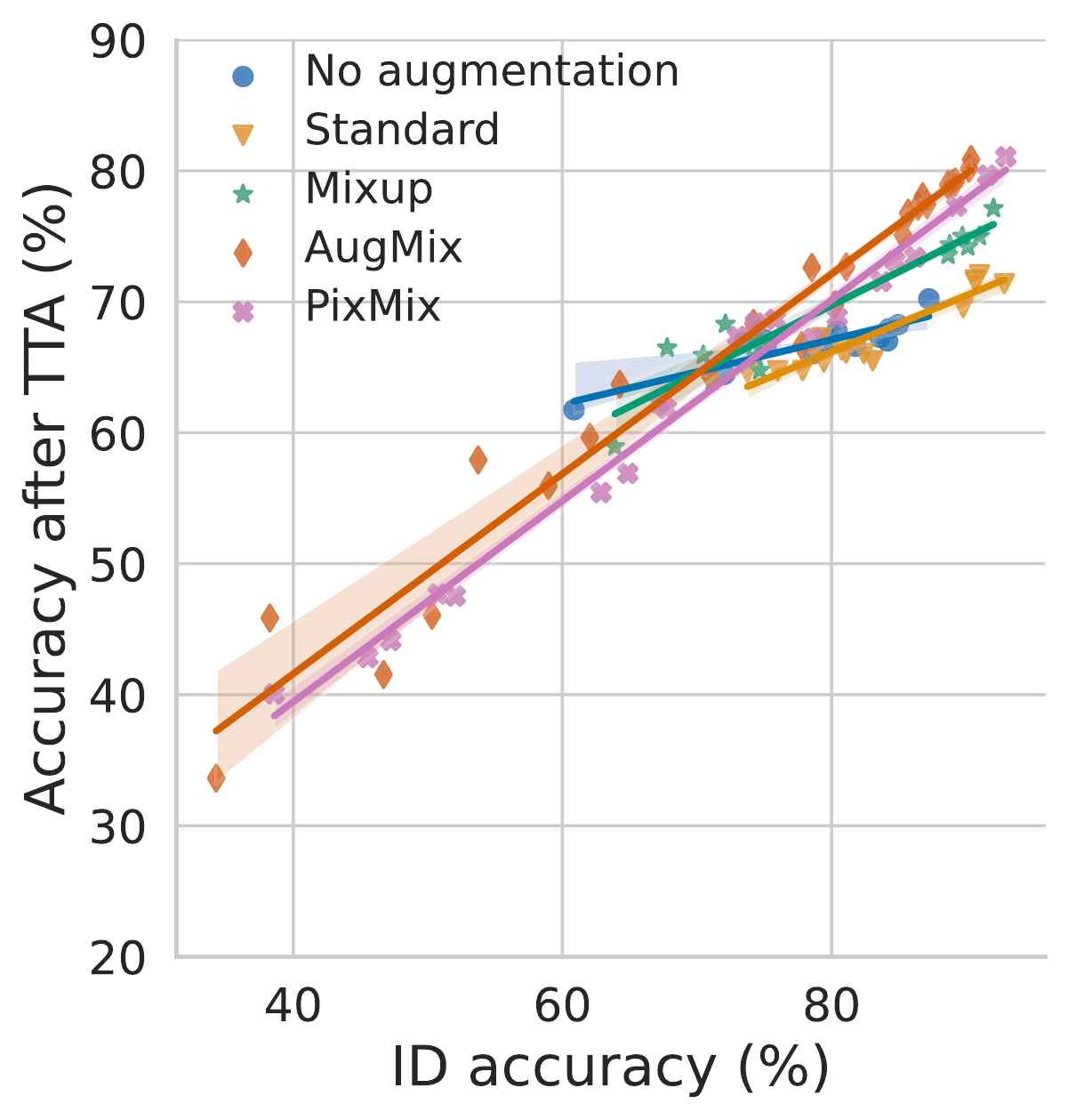}
		\label{fig:data_aug_bn_adapt_id_ood_appendix}
	}
	\subfigure[\small TENT]{
		\includegraphics[width=0.3\textwidth]{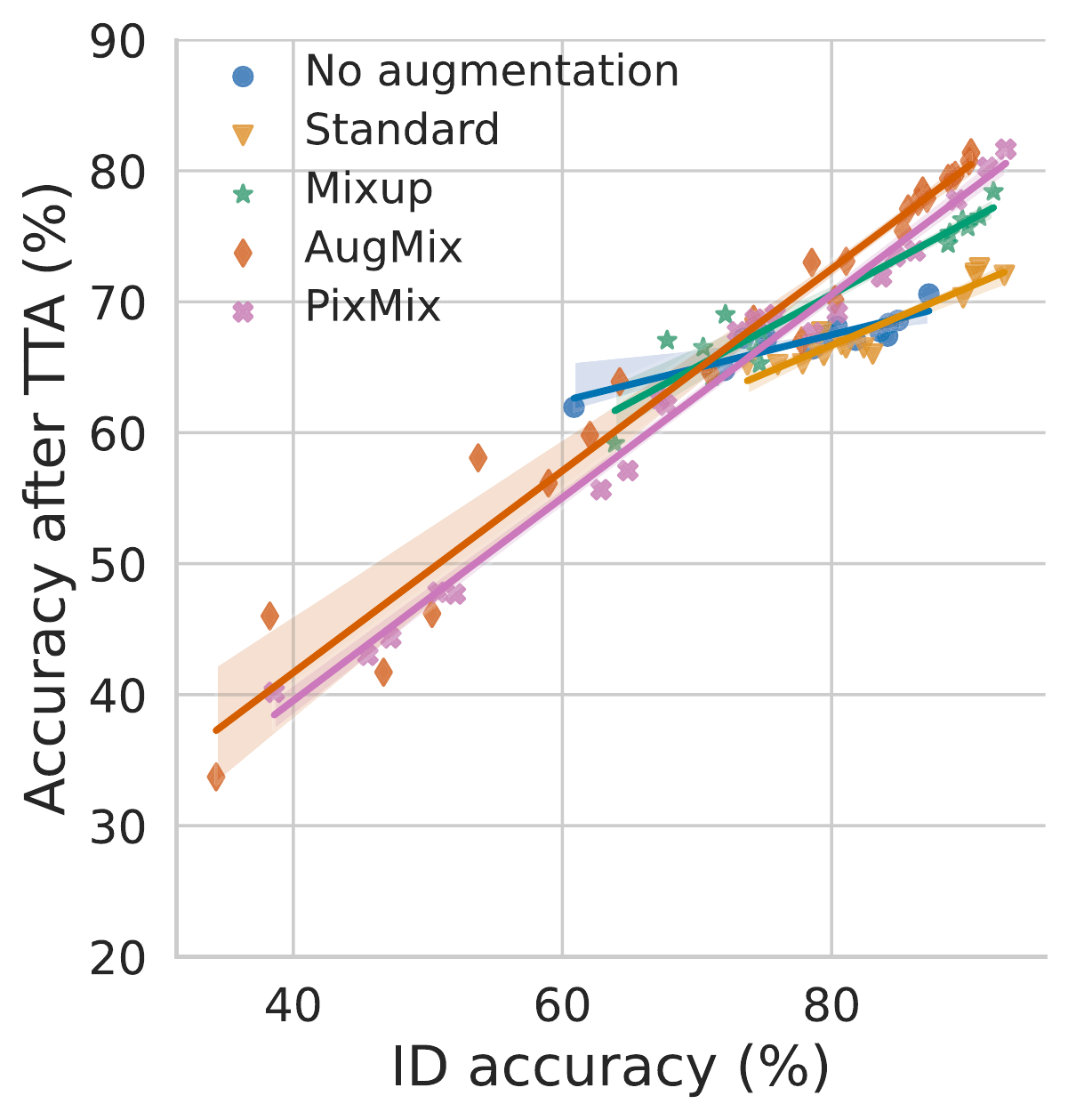}
		\label{fig:data_aug_tent_id_ood_appendix}
	}
	\subfigure[\small SHOT]{
		\includegraphics[width=0.3\textwidth]{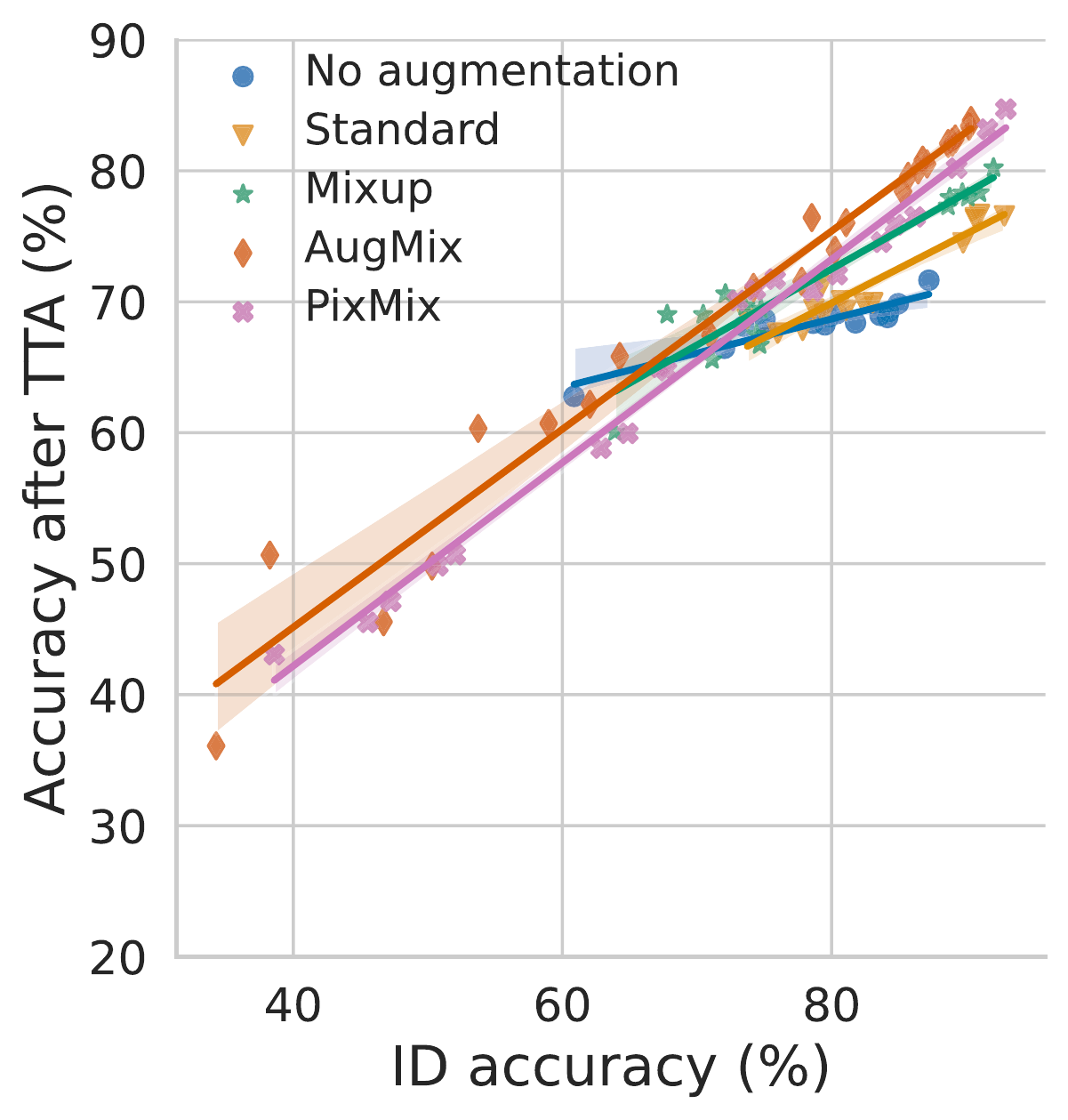}
		\label{fig:data_aug_shot_id_ood_appendix}
	}
	\subfigure[\small BN\_Adapt]{
		\includegraphics[width=0.3\textwidth]{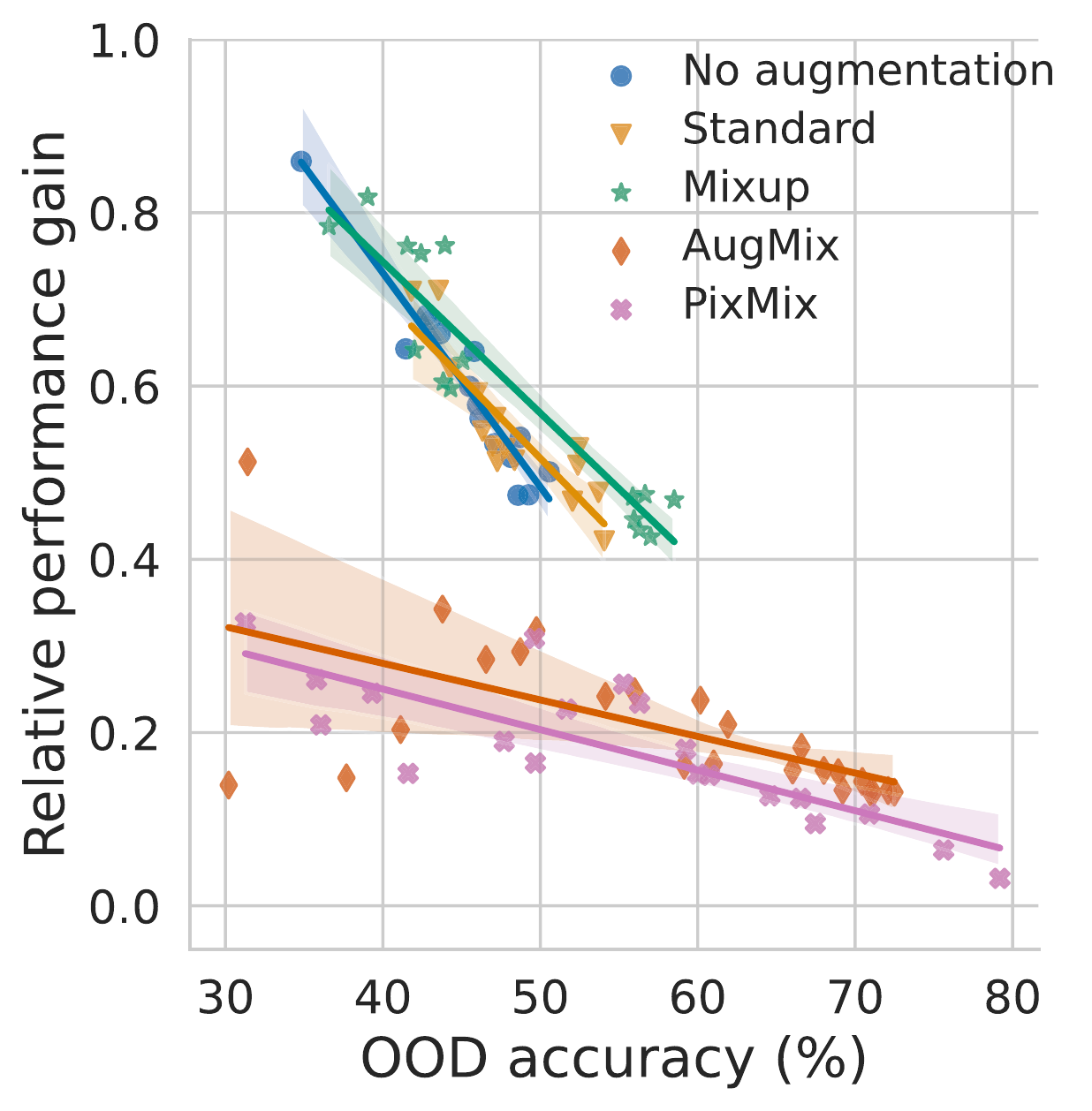}
		\label{fig:data_aug_bn_adapt_ood_gain_appendix}
	}
	\subfigure[\small TENT]{
		\includegraphics[width=0.3\textwidth]{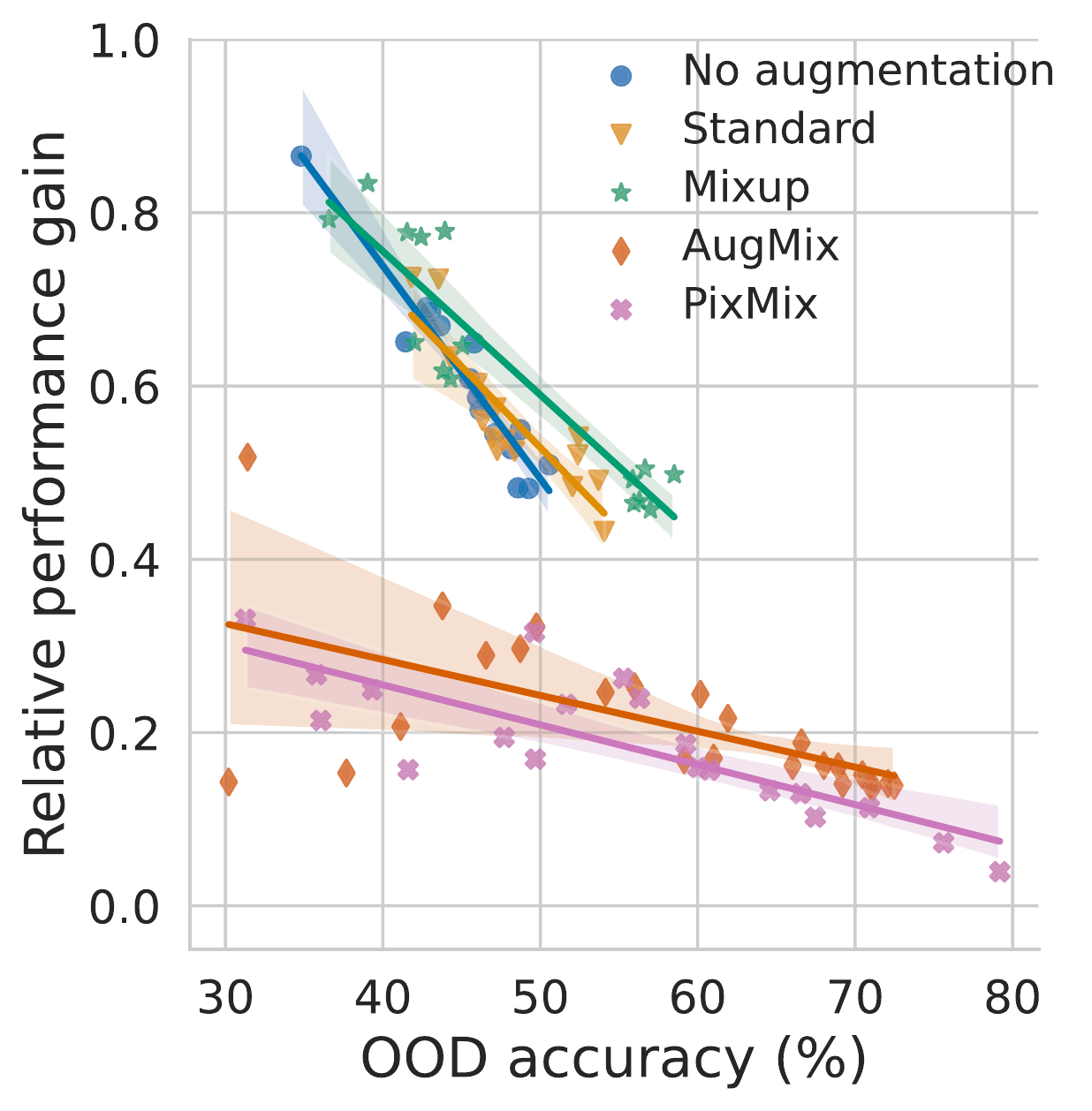}
		\label{fig:data_aug_tent_ood_gain_appendix}
	}
	\subfigure[\small SHOT]{
		\includegraphics[width=0.3\textwidth]{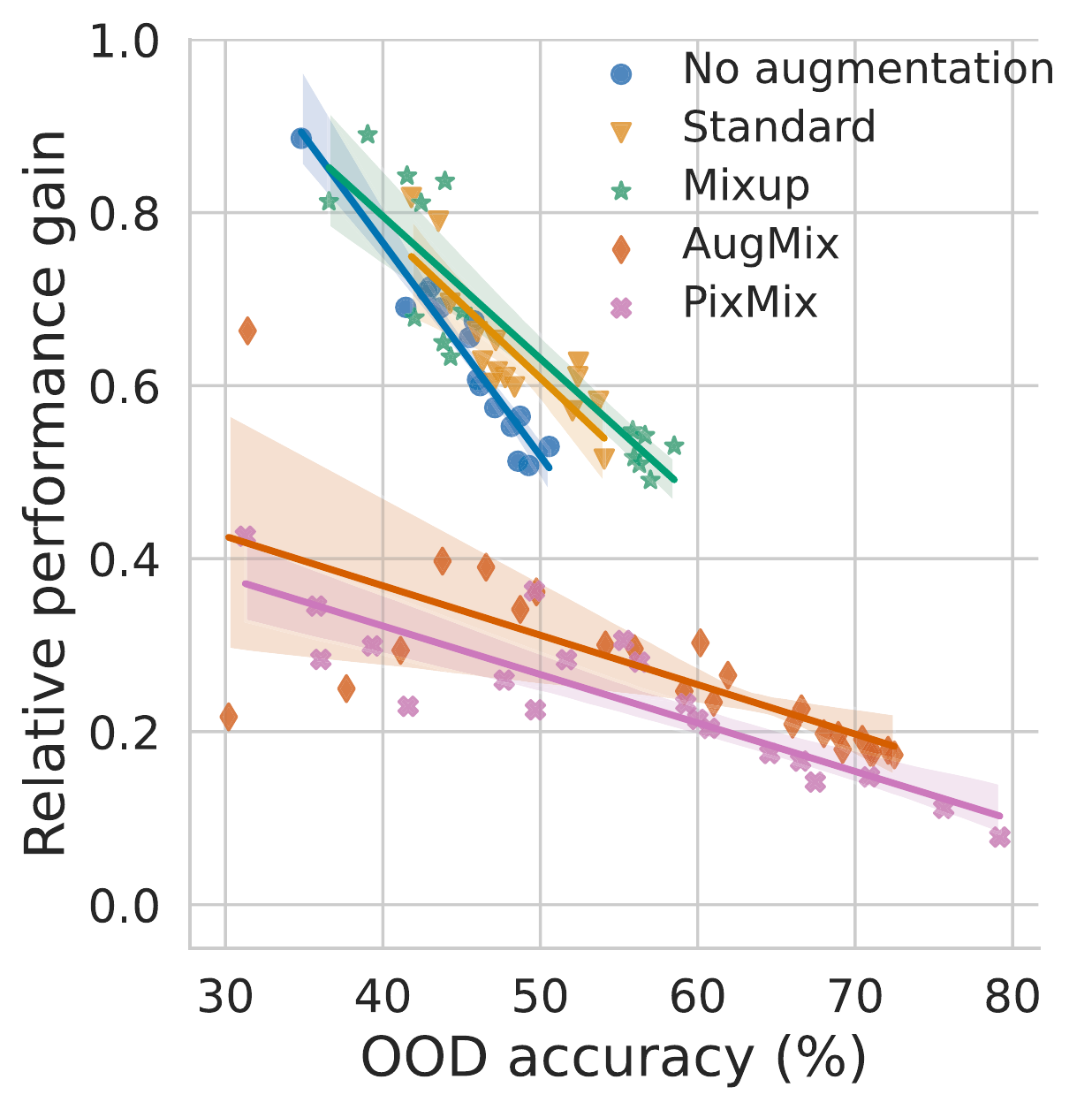}
		\label{fig:data_aug_shot_ood_gain_appendix}
	}
	\caption{\small
		\textbf{The effect of data augmentation on TTA performance in the target domain}.
		TENT and SHOT use episodic adaptation with oracle model selection and choose ResNet-26 as the base model.
	}
	\label{fig:data_augmentations_rn26_details}
\end{figure*}

\begin{figure*}[!h]
	\centering
	\subfigure[\small BN\_Adapt]{
		\includegraphics[width=0.3\textwidth]{figures/model_quality/rn26_bn_adapt_ood_before_and_after_tta.pdf}
		\label{fig:data_aug_rn26_bn_adapt_online_appendix}
	}
	\subfigure[\small TENT]{
		\includegraphics[width=0.3\textwidth]{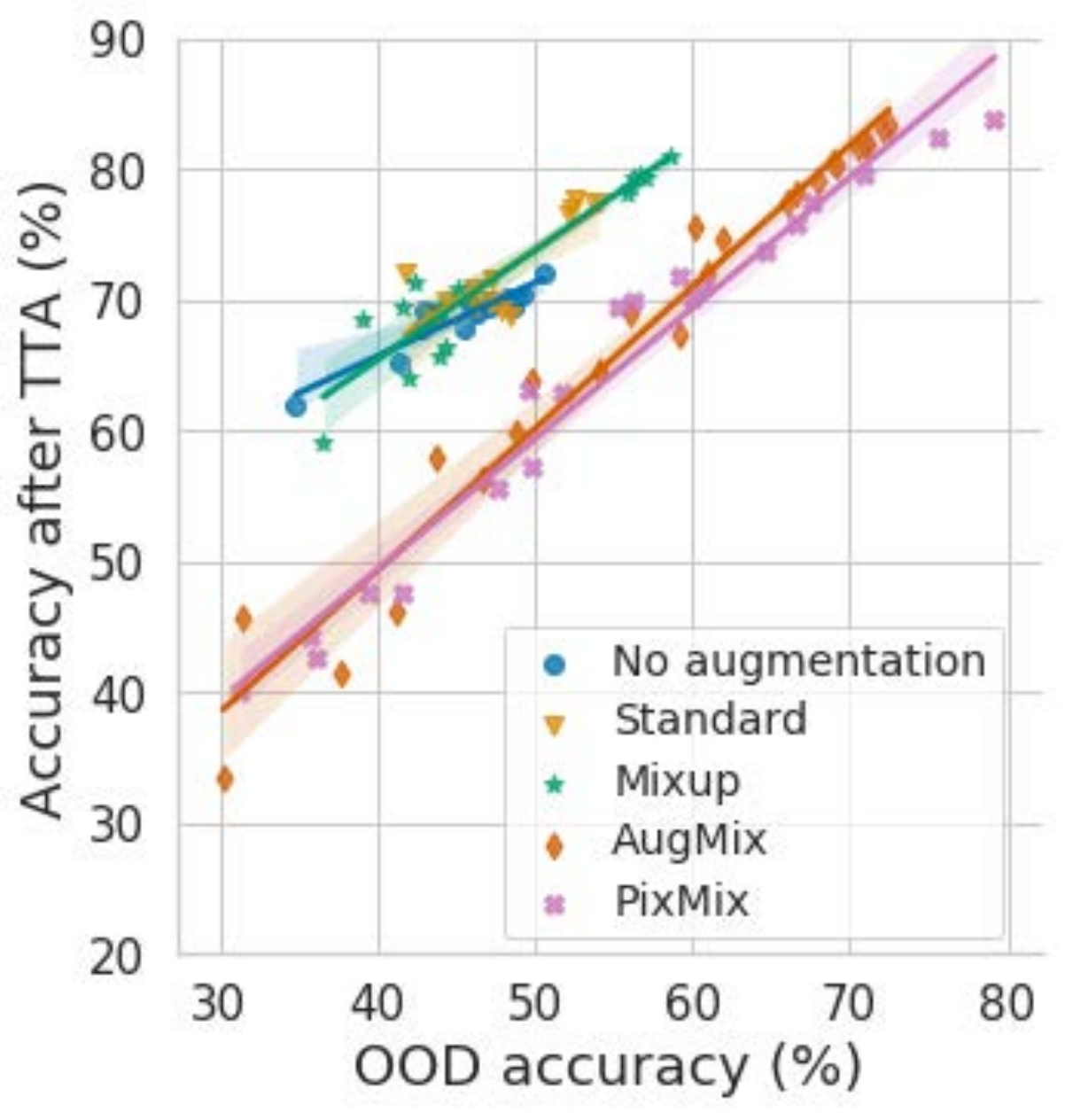}
		\label{fig:data_aug_rn26_tent_online_appendix}
	}
	\subfigure[\small SHOT]{
		\includegraphics[width=0.3\textwidth]{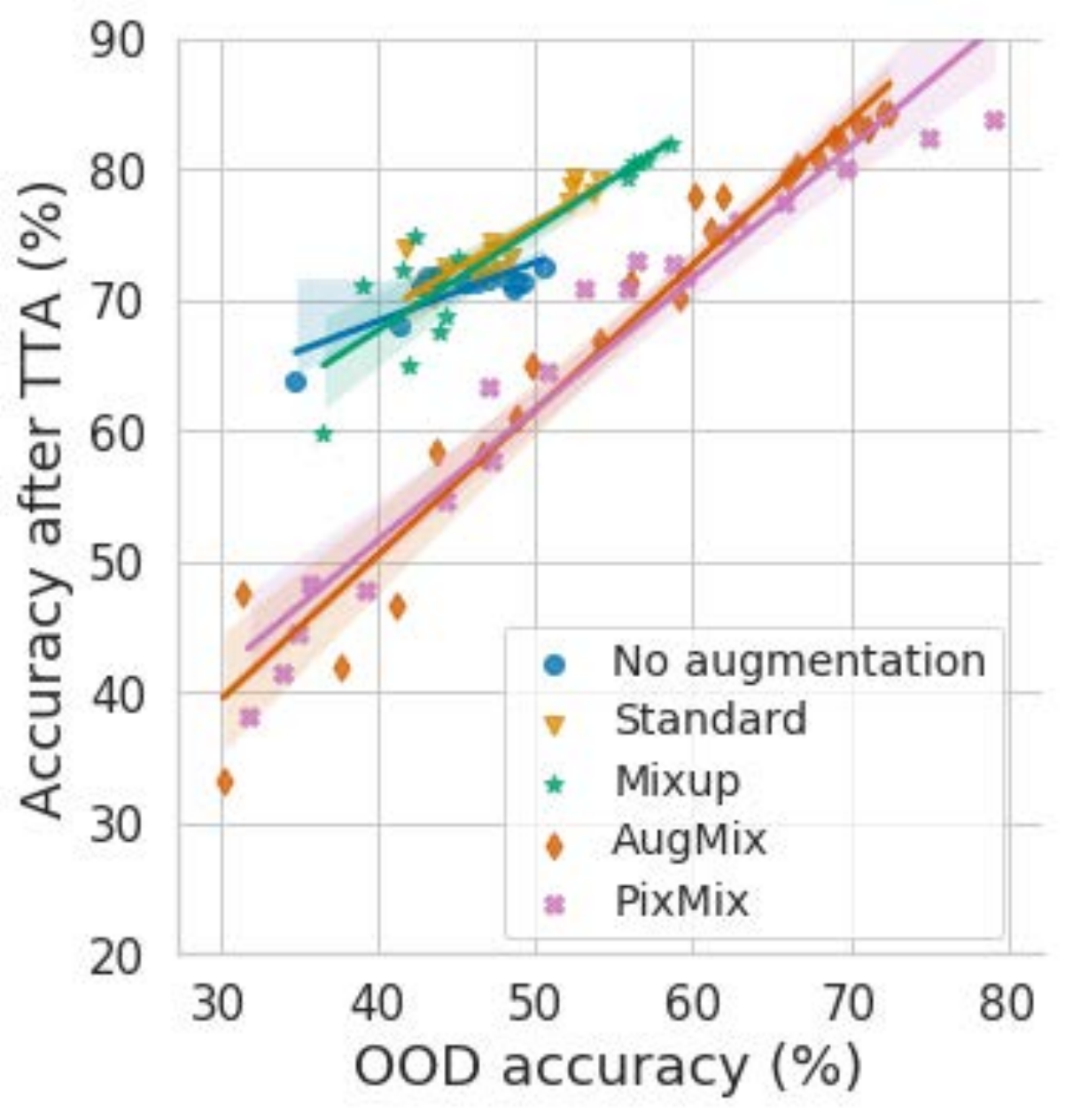}
		\label{fig:data_aug_rn26_shot_online_appendix}
	}
	\subfigure[\small BN\_Adapt]{
		\includegraphics[width=0.3\textwidth]{figures/model_quality/rn26_bn_adapt_id_ood_performance.pdf}
		\label{fig:data_aug_bn_adapt_online_id_ood_appendix}
	}
	\subfigure[\small TENT]{
		\includegraphics[width=0.3\textwidth]{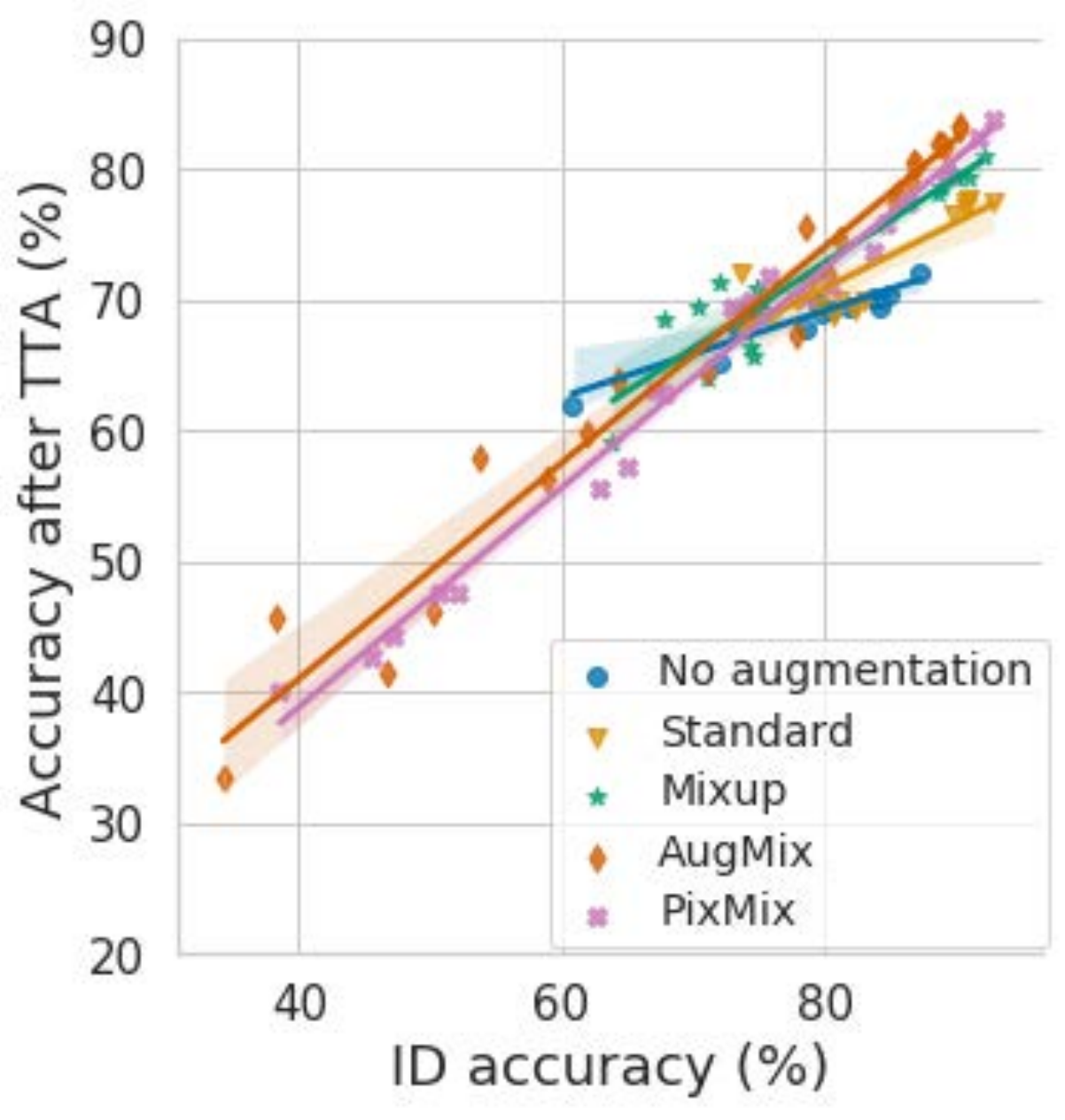}
		\label{fig:data_aug_tent_online_id_ood_appendix}
	}
	\subfigure[\small SHOT]{
		\includegraphics[width=0.3\textwidth]{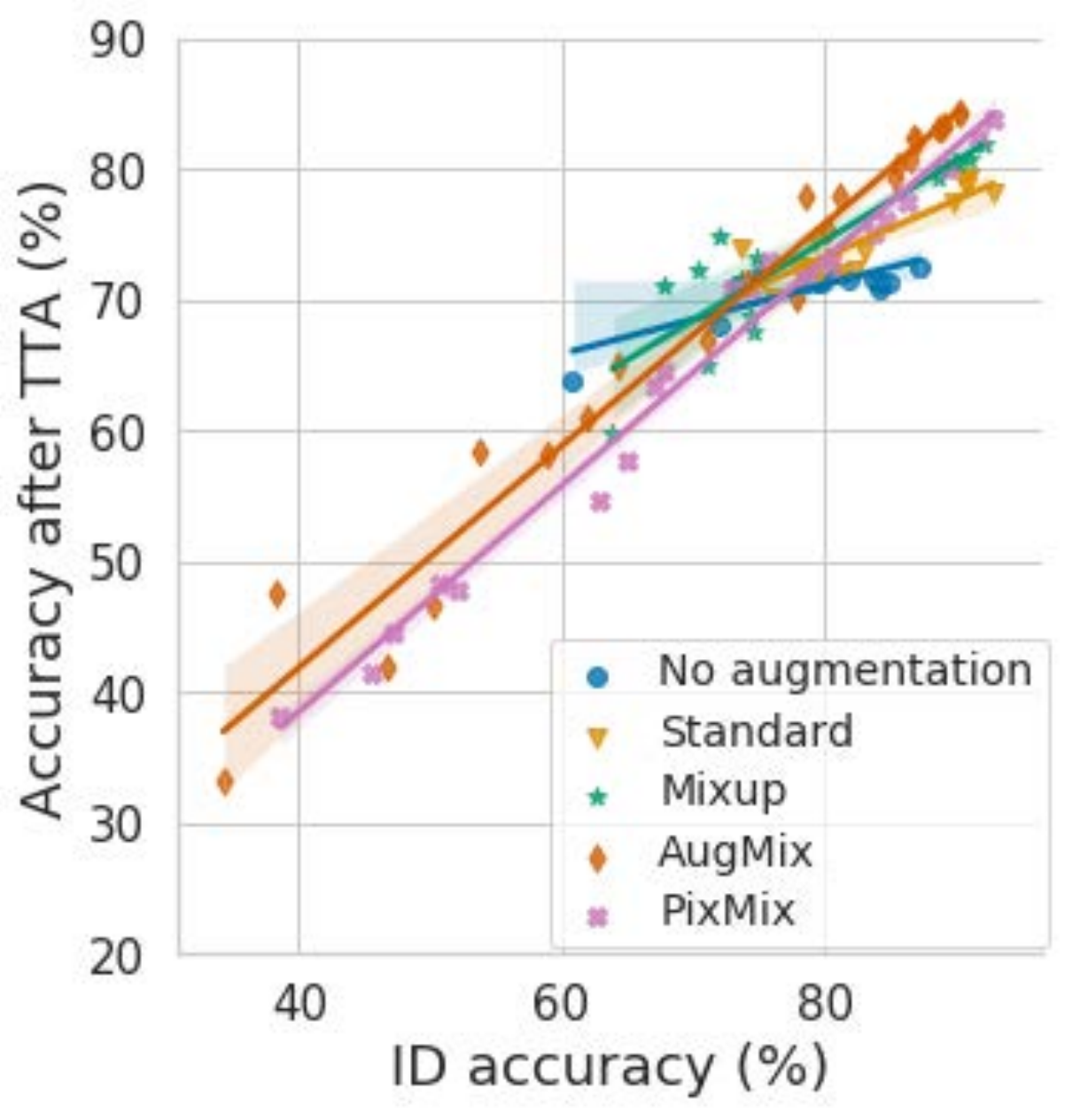}
		\label{fig:data_aug_shot_online_id_ood_appendix}
	}
	\subfigure[\small BN\_Adapt]{
		\includegraphics[width=0.3\textwidth]{figures/model_quality/rn26_bn_adapt_ood_performance_gain.pdf}
		\label{fig:data_aug_bn_adapt_online_ood_gain_appendix}
	}
	\subfigure[\small TENT]{
		\includegraphics[width=0.3\textwidth]{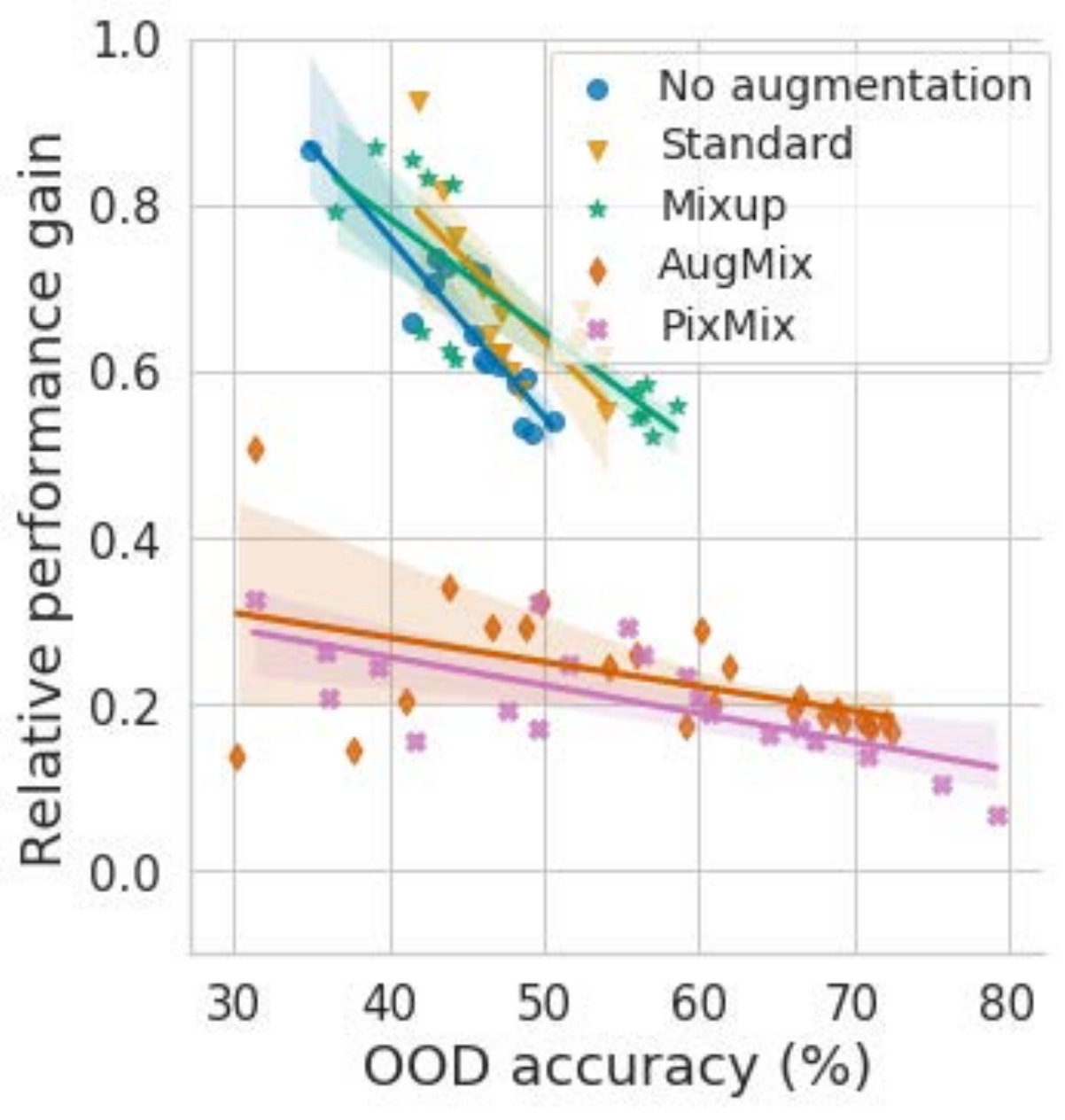}
		\label{fig:data_aug_tent_online_ood_gain_appendix}
	}
	\subfigure[\small SHOT]{
		\includegraphics[width=0.3\textwidth]{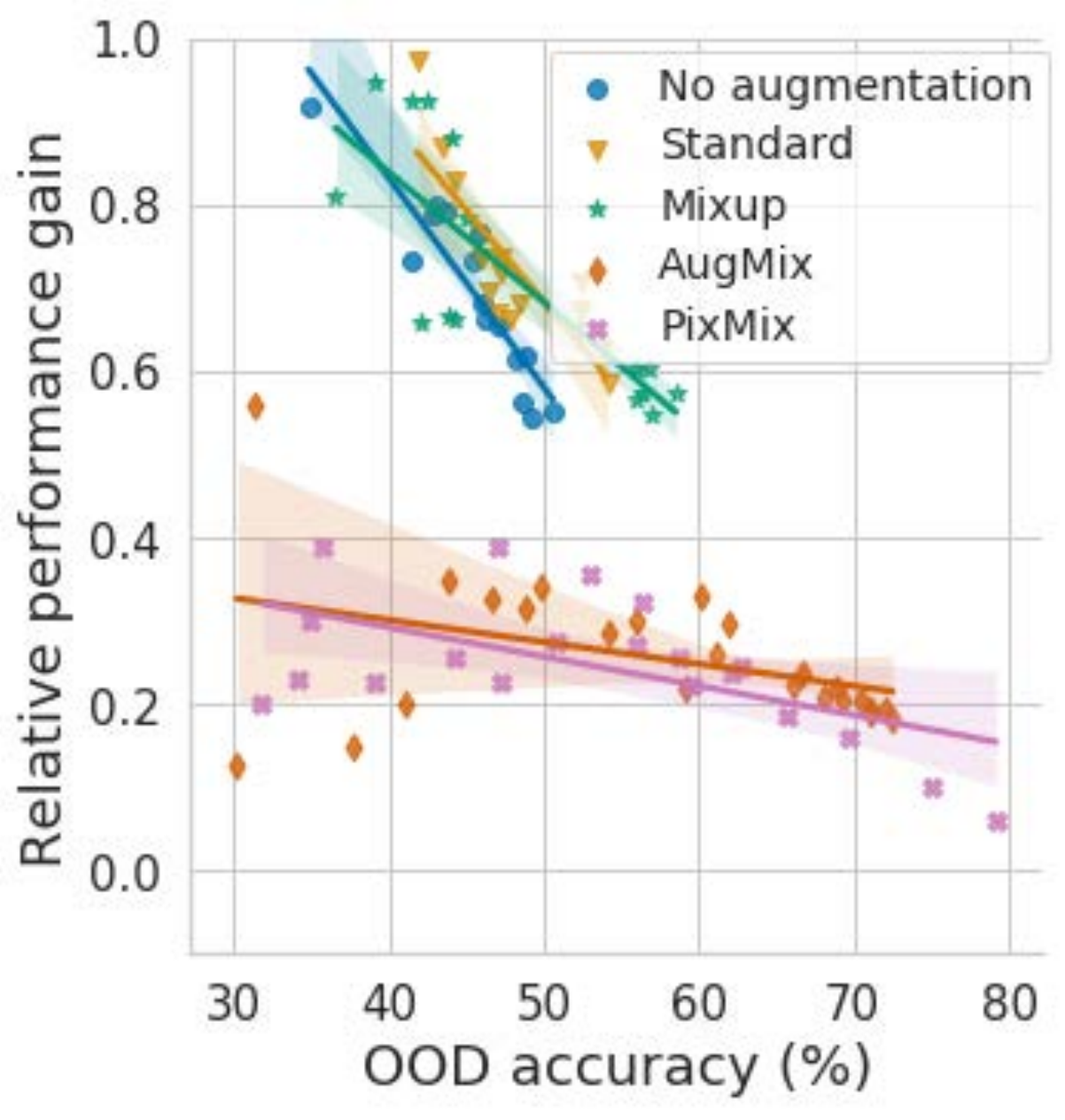}
		\label{fig:data_aug_shot_online_ood_gain_appendix}
	}
	\caption{\small
		\textbf{The effect of data augmentation on TTA performance in the target domain}.
		TENT and SHOT use online adaptation without oracle model selection and grid search the best performance. We use ResNet-26 as the base model here.
	}
	\label{fig:data_augmentations_rn26_online_addition}
\end{figure*}

\begin{figure*}[!h]
	\centering
	\subfigure[\small BN\_Adapt]{
		\includegraphics[width=0.3\textwidth]{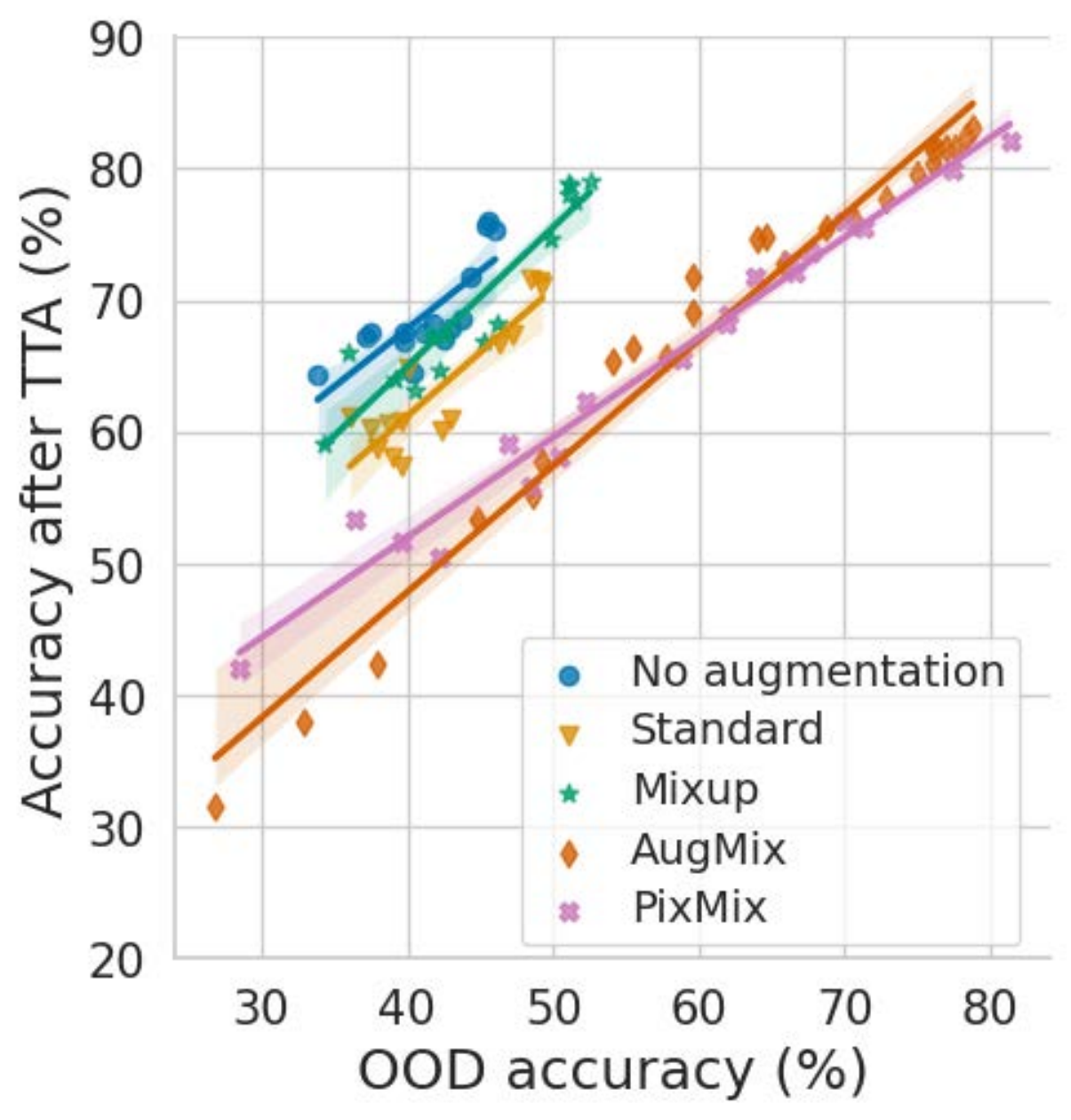}
		\label{fig:data_aug_wrn40_2_bn_adapt}
	}
	\subfigure[\small TENT]{
		\includegraphics[width=0.3\textwidth]{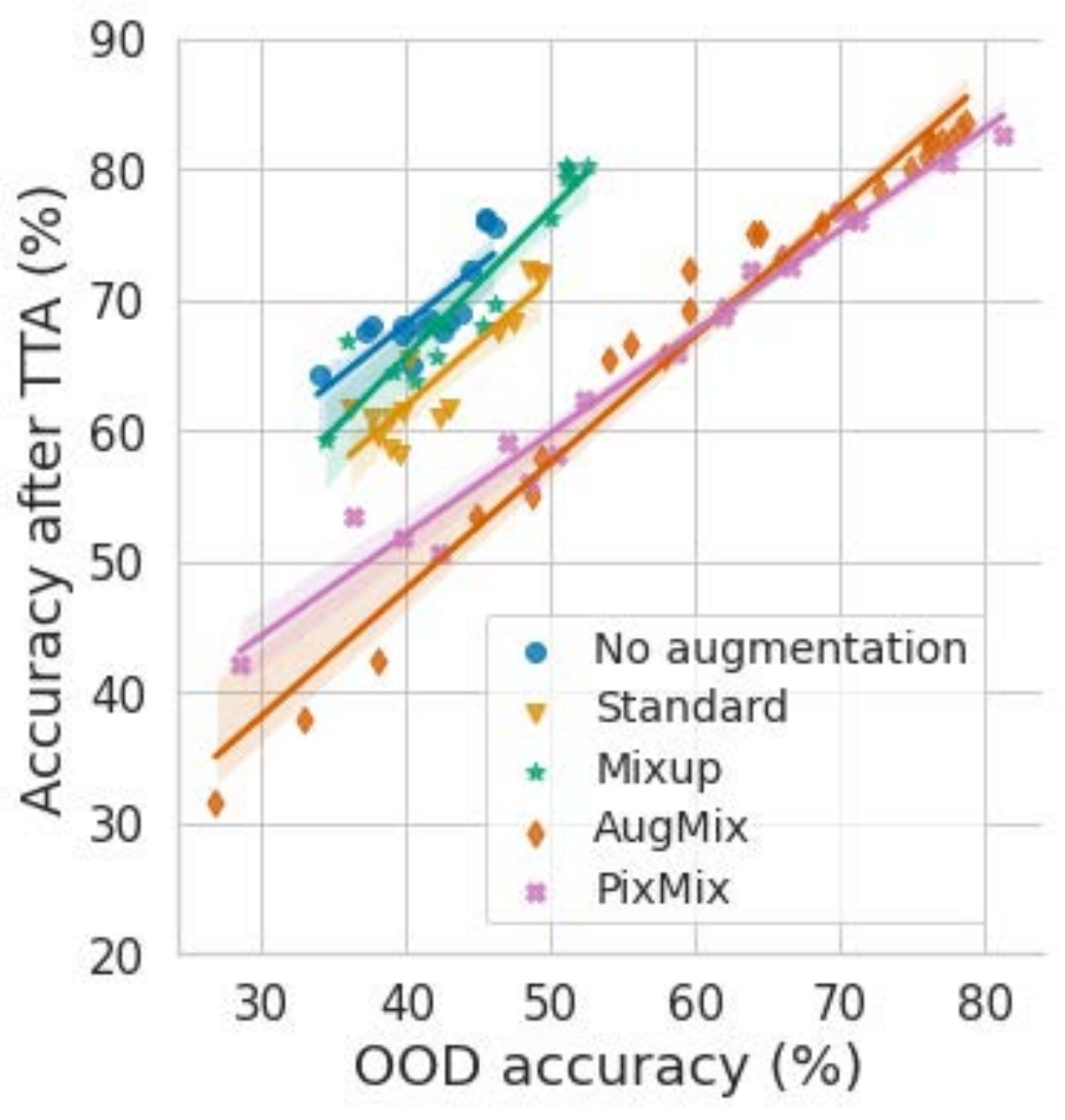}
		\label{fig:data_aug_wrn40_2_tent}
	}
	\subfigure[\small SHOT]{
		\includegraphics[width=0.3\textwidth]{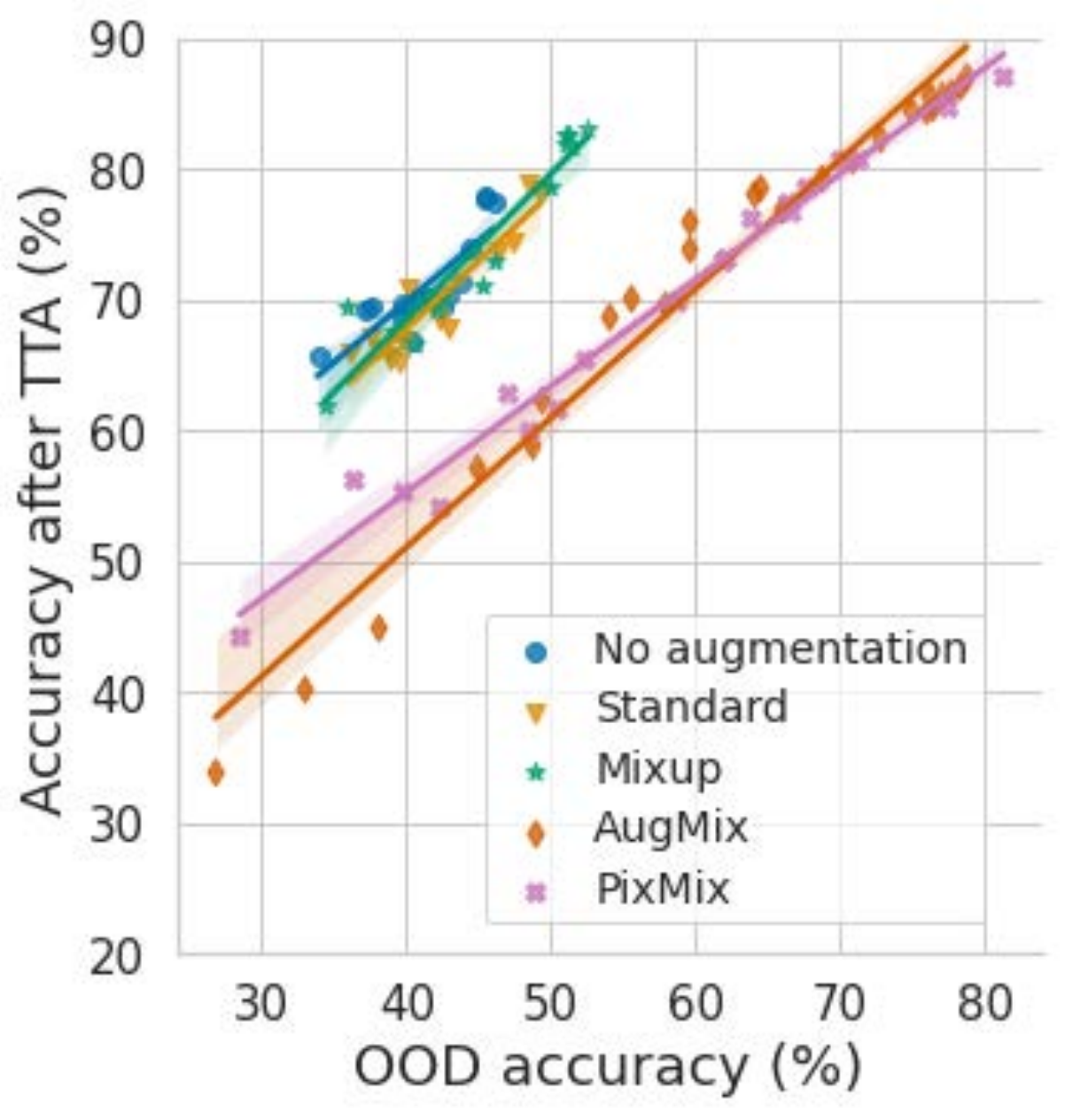}
		\label{fig:data_aug_wrn40_2_shot}
	}
	\subfigure[\small BN\_Adapt]{
		\includegraphics[width=0.3\textwidth]{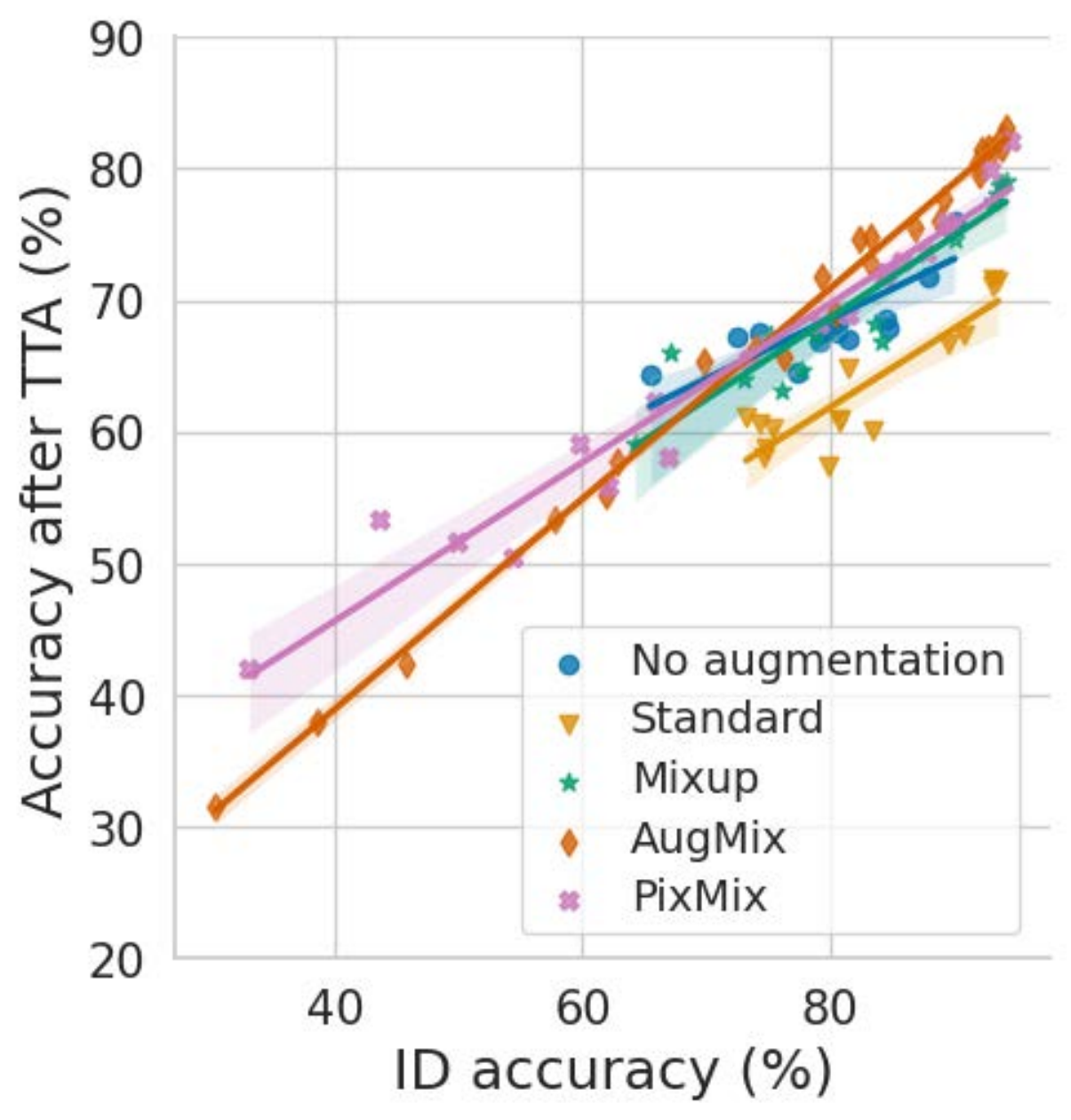}
		\label{fig:wrn40_2_data_aug_bn_adapt_id_ood}
	}
	\subfigure[\small TENT]{
		\includegraphics[width=0.3\textwidth]{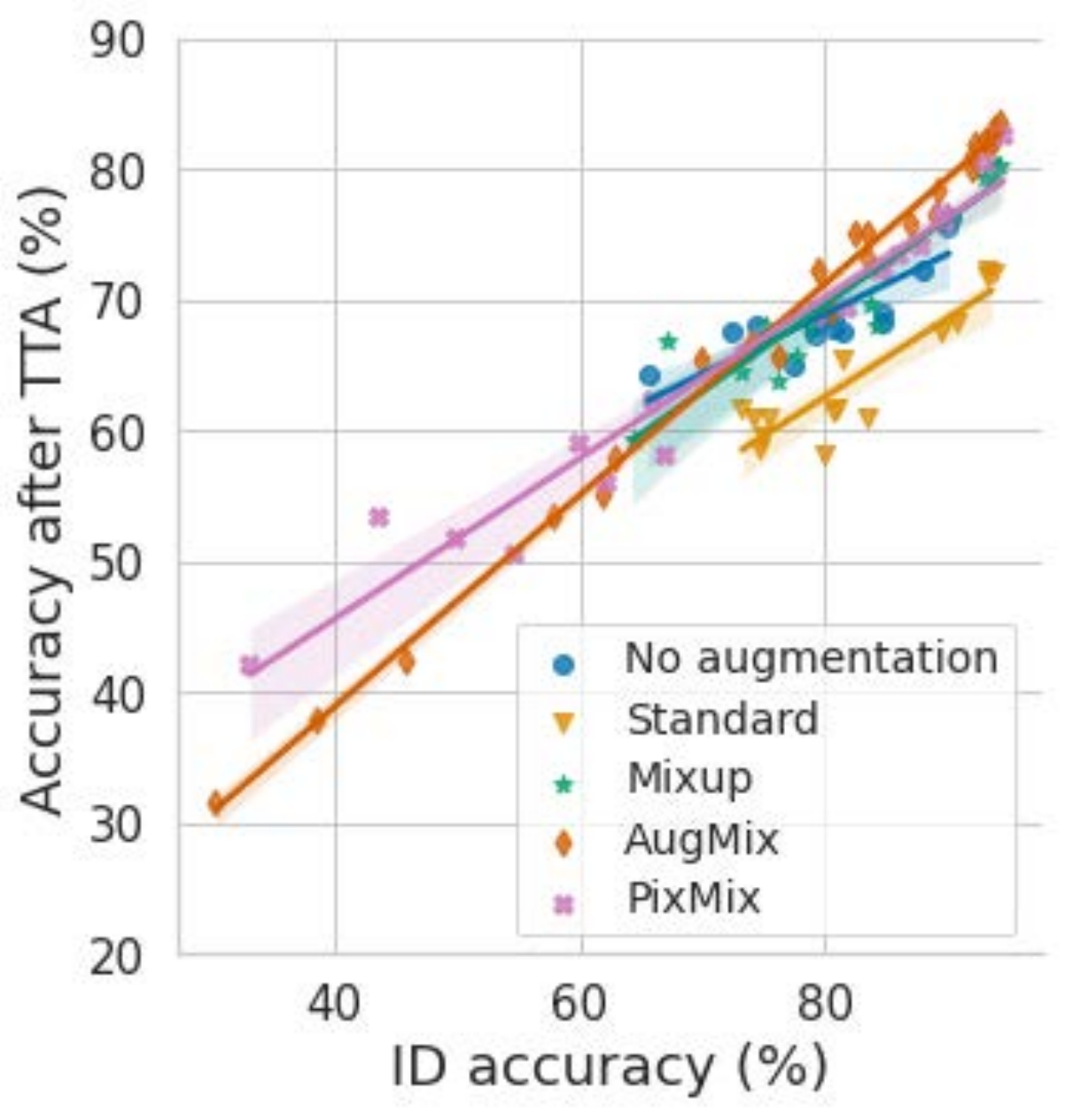}
		\label{fig:wrn40_2_data_aug_tent_id_ood}
	}
	\subfigure[\small SHOT]{
		\includegraphics[width=0.3\textwidth]{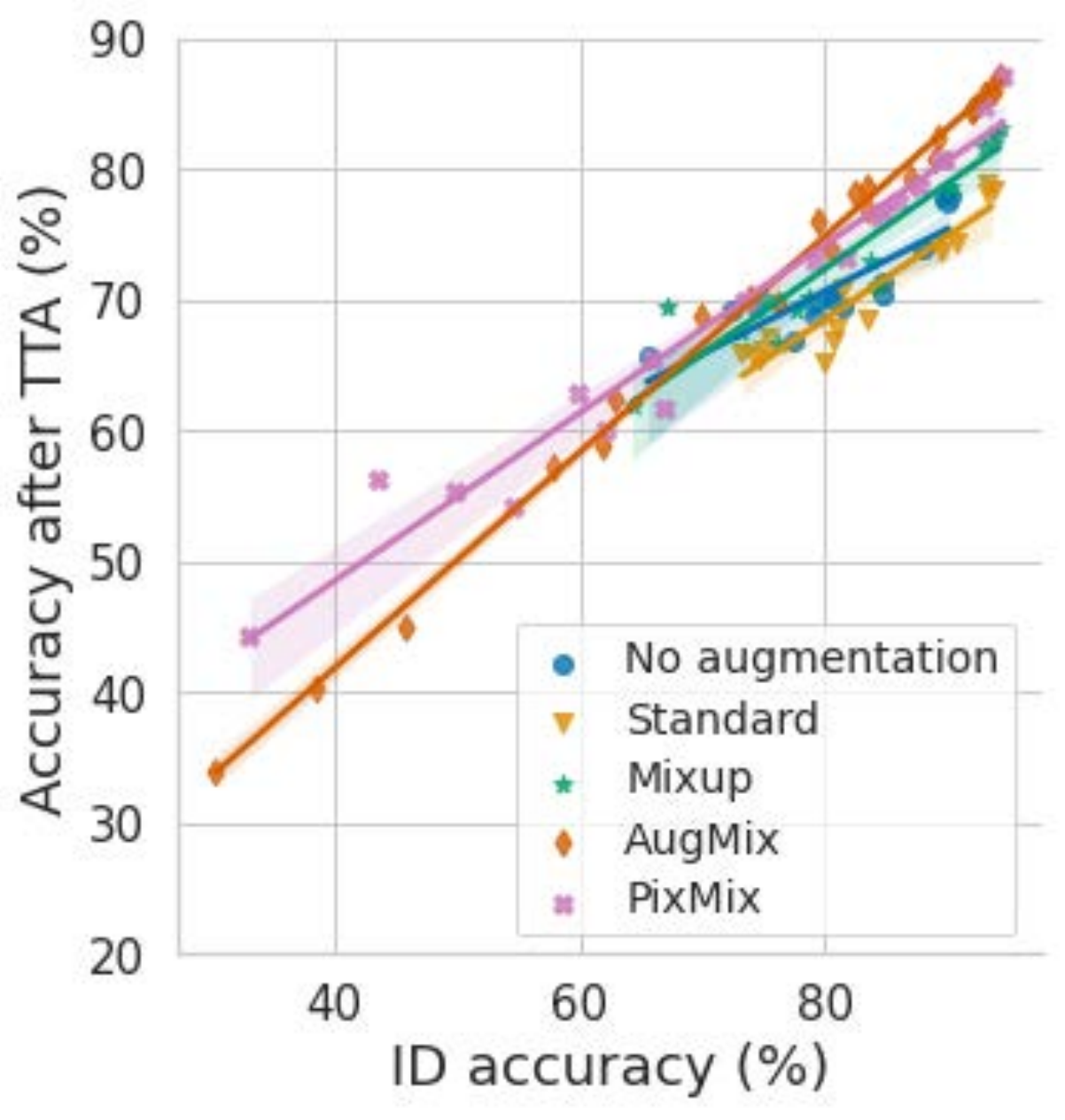}
		\label{fig:wrn40_2_data_aug_shot_id_ood}
	}
	\subfigure[\small BN\_Adapt]{
		\includegraphics[width=0.3\textwidth]{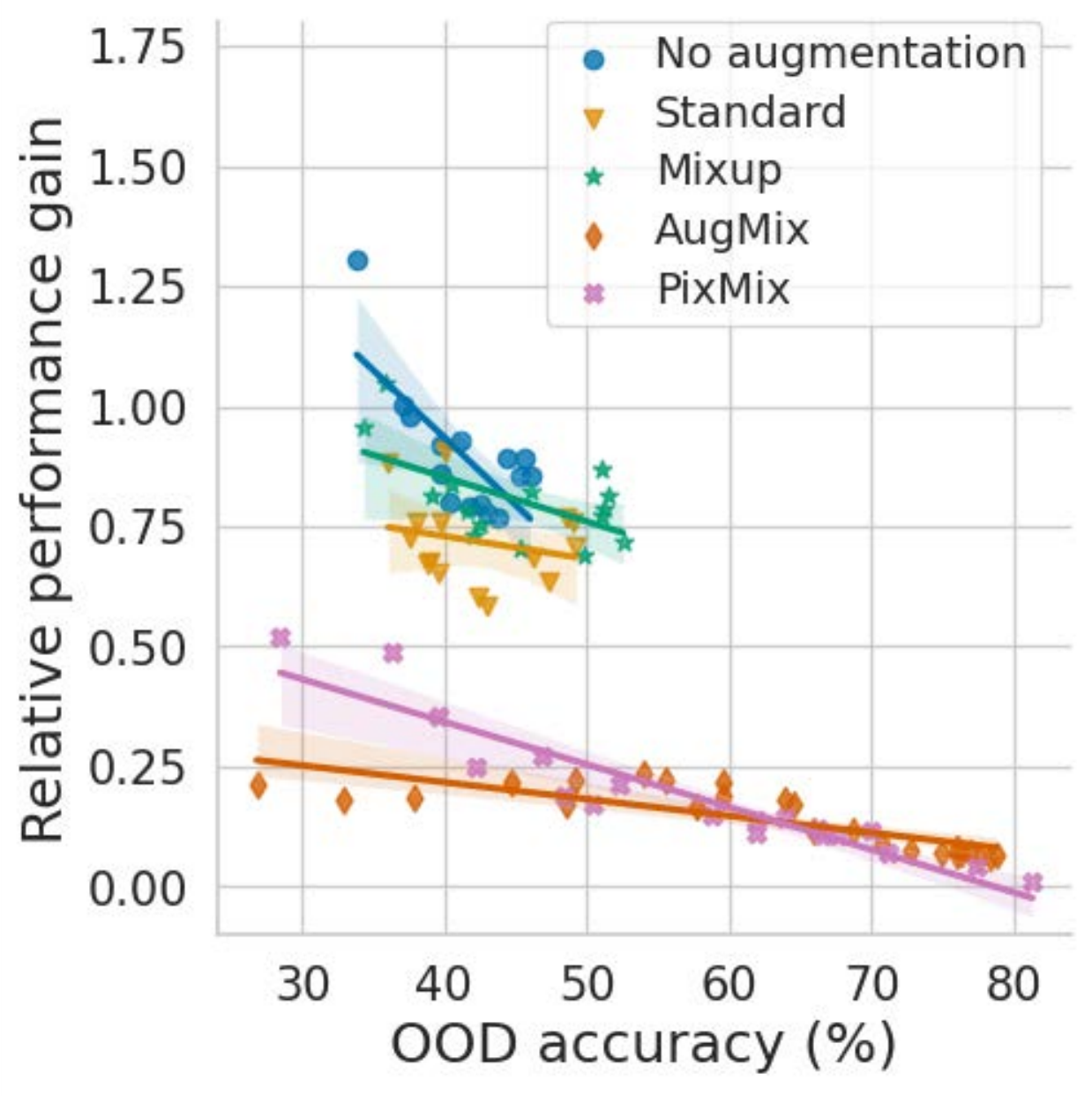}
		\label{fig:wrn40_2_data_aug_bn_adapt_ood_gain}
	}
	\subfigure[\small TENT]{
		\includegraphics[width=0.3\textwidth]{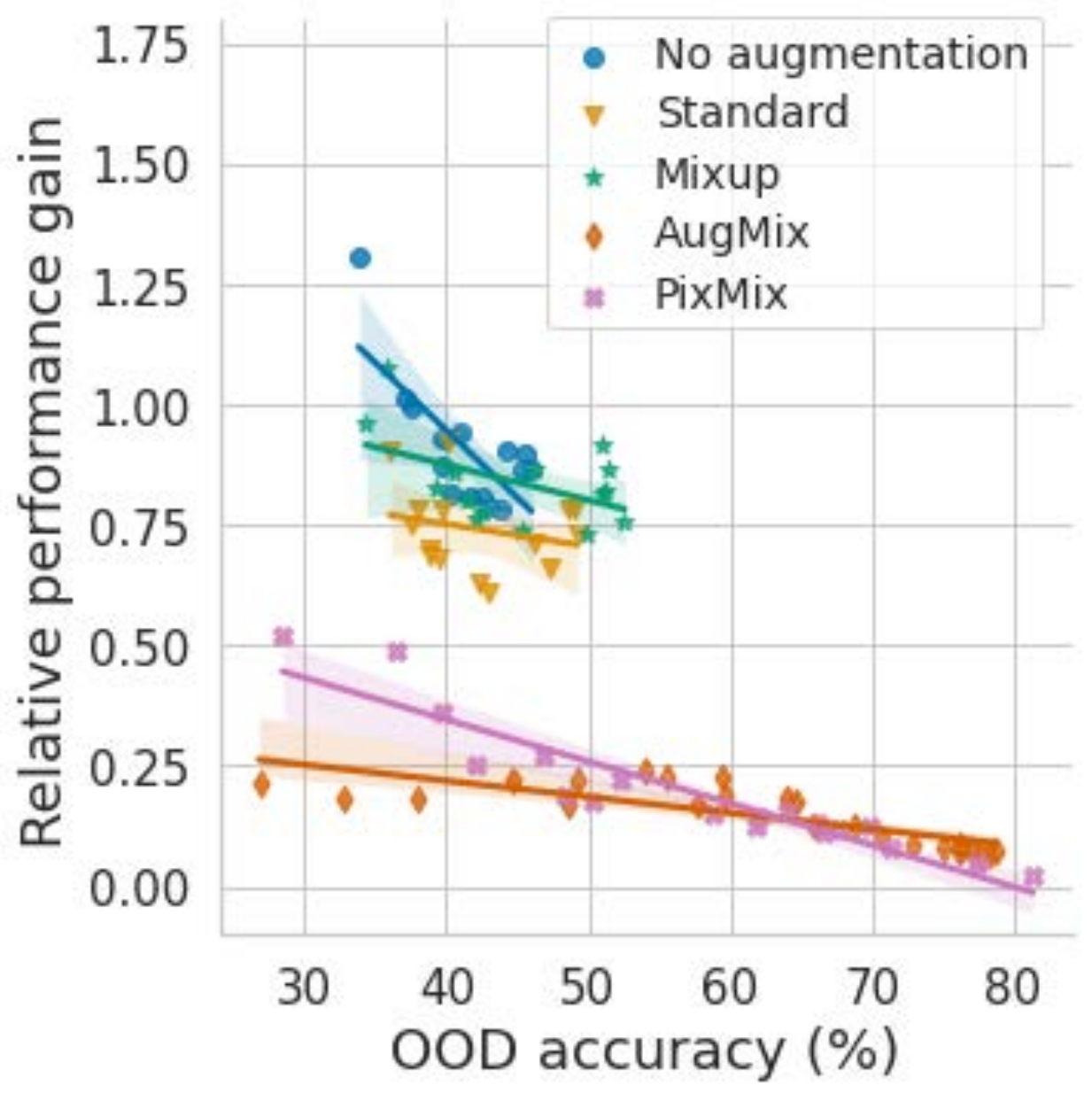}
		\label{fig:wrn40_2_data_aug_tent_ood_gain}
	}
	\subfigure[\small SHOT]{
		\includegraphics[width=0.3\textwidth]{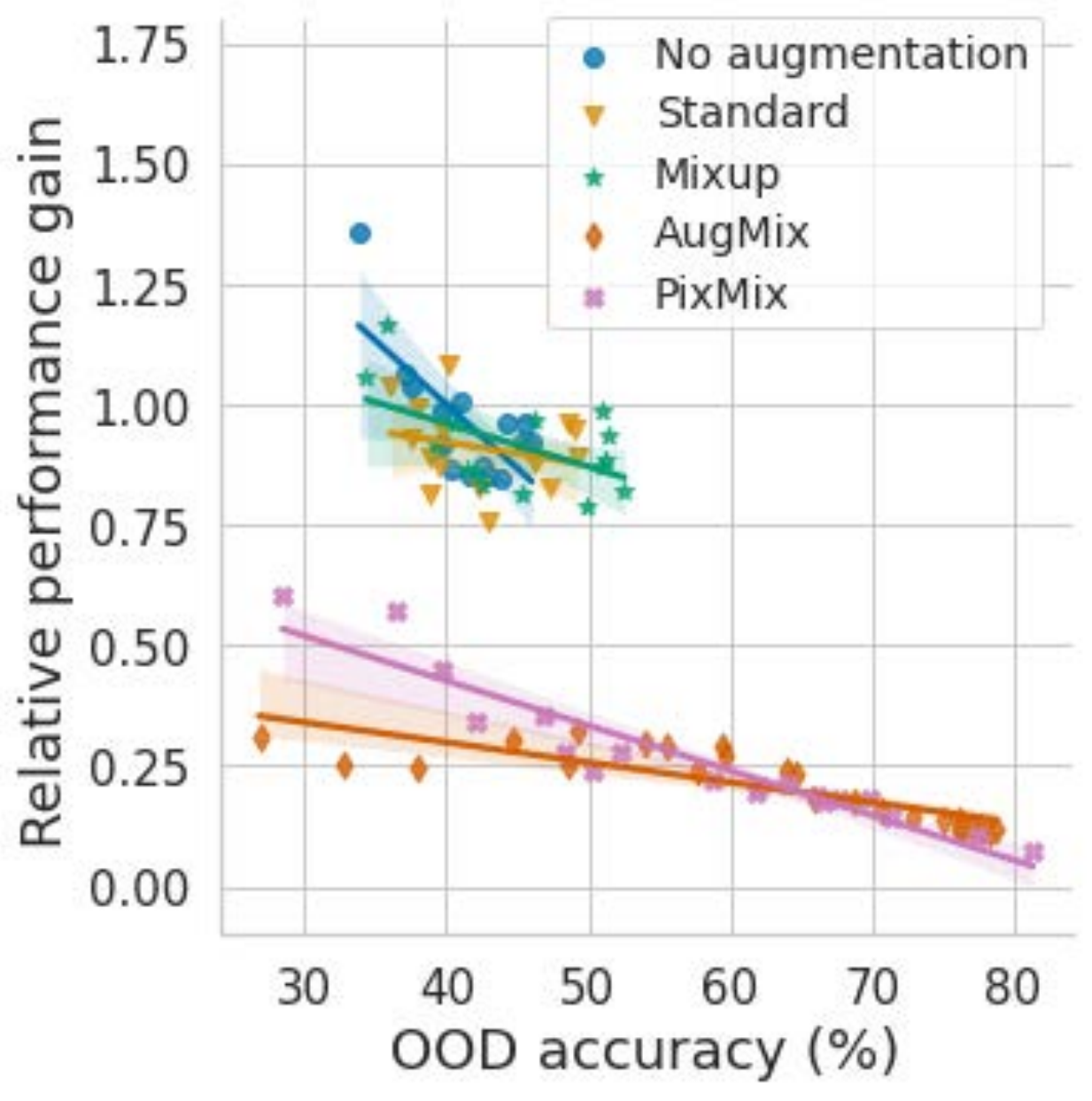}
		\label{fig:wrn40_2_data_aug_shot_ood_gain}
	}
	\caption{\small
		\textbf{The effect of data augmentation on TTA performance in the target domain}.
		TENT and SHOT use episodic adaptation with oracle model selection and choose WideResNet40-2 as the base model.
	}
	\label{fig:data_augmentations_wrn40_2_addition}
\end{figure*}

In~\autoref{fig:data_augmentations_rn26_details}, ~\autoref{fig:data_augmentations_rn26_online_addition}, and~\autoref{fig:data_augmentations_wrn40_2_addition}, we show more data augmentation results across different model architectures and different evaluation protocols.

\section{Additional Results} \label{appendix:additional_results}
\subsection{TTA on Label Shifts} \label{appendix:additional_results_on_label_shifts}
The efficacy of most TTA methods drops substantially when confronted with label shifts regardless of the data itself.
Here we report the additional results of TTA methods on the CIFAR10-100 dataset, which is another common benchmark for evaluating distribution shift problems.
The results are summarized in \cref{tab:additional_results_on_label_shifts}.

\begin{table}[!h]
	\caption{\small
		\textbf{Adaptation performance (error in \%) of TTA methods over label shifts on CIFAR100 with different severities. Optimal results are highlighted by \textbf{bold}.}
		\looseness=-1
	}
	\vspace{-0.5em}
	\label{tab:additional_results_on_label_shifts}
	\centering
	\setlength{\tabcolsep}{2pt}
	\def\arraystretch{1.25}
	\resizebox{.5\textwidth}{!}{
		\begin{tabular}{@{}l  c c  cc }
			\toprule
			\multicolumn{1}{c}{} & \multicolumn{2}{c}{\textbf{large label shift ($\alpha=1$)} $\downarrow$} & \multicolumn{2}{c}{\textbf{small label shift ($\alpha=10$)} $\downarrow$}                                                       \\
			\cmidrule(lr){2-3} \cmidrule(lr){4-5}
			                     & episodic                                                                 & online                                                                    & episodic                 & online                   \\
			\midrule
			Baseline             & $31.2(\pm 1.2)$                                                          & $\mathbf{31.2(\pm 1.2)}$                                                  & $29.3(\pm 1.4)$          & $29.3(\pm 1.4)$          \\
			\midrule
			BN\_adapt            & $41.9(\pm 1.3)$                                                          & $41.9(\pm 1.3)$                                                           & $37.7(\pm 2.0)$          & $37.7(\pm 2.0)$          \\
			\midrule
			SHOT                 & $30.4(\pm 1.0)$                                                          & $32.6(\pm 1.8)$                                                           & $27.7(\pm 1.5)$          & $29.5(\pm 1.7)$          \\
			\midrule
			TTT                  & $31.8(\pm 1.4)$                                                          & $32.9(\pm 1.5)$                                                           & $29.8(\pm 1.2)$          & $31.2(\pm 1.4)$          \\
			\midrule
			TENT                 & $40.2(\pm 1.2)$                                                          & $38.8(\pm 1.1)$                                                           & $36.0(\pm 1.5)$          & $35.4(\pm 1.9)$          \\
			\midrule
			T3A                  & $32.1(\pm 1.4)$                                                          & $32.1(\pm 1.4)$                                                           & $30.2(\pm 2.0)$          & $30.2(\pm 2.0)$          \\
			\midrule
			CoTTA                & $40.8(\pm 1.3)$                                                          & $68.4(\pm 0.7)$                                                           & $36.8(\pm 2.6)$          & $67.0(\pm 1.5)$          \\
			\midrule
			MEMO                 & $\mathbf{28.3(\pm 1.3)}$                                                 & $32.1(\pm 2.6)$                                                           & $\mathbf{26.2(\pm 2.0)}$ & $30.4(\pm 1.9)$          \\
			\midrule
			NOTE                 & $30.0(\pm 0.9)$                                                          & $\mathbf{31.2(\pm 1.1)}$                                                  & $28.2(\pm 1.7)$          & $\mathbf{29.2(\pm 1.4)}$ \\
			\midrule
			Conjugate PL         & $39.9(\pm 1.3)$                                                          & $37.9(\pm 1.9)$                                                           & $35.6(\pm 1.5)$          & $36.1(\pm 2.1)$          \\
			\midrule
			SAR                  & $40.1(\pm 1.1)$                                                          & $39.0(\pm 1.1)$                                                           & $36.3(\pm 1.5)$          & $35.8(\pm 1.5)$          \\
			\bottomrule
		\end{tabular}
	}
\end{table}

\subsection{Empirical Studies of Normalization Layers Effects in TTA} \label{appendix:additional_results_on_normalization_layers}
The most recent work~\citep{niu2023towards} dug into the effects of normalization layers on TTA performance and found that
TTA can perform more stably with batch-agnostic norm layers, i.e., group or layer norm. Here we revisit TTA performance on
all data scenarios we have discussed before when equipped with group or layer norm.

\begin{table*}[!h]
	\caption{
		Results of TTA performance on \textbf{ResNet26-GN}. We report the \textbf{error in (\%)} on CIFAR10-C severity level 5 \textbf{under uniformly distributed test streams}.
		Optimal results in \resetmodel \& \notresetmodel are highlighted by \textbf{bold} and \textcolor{blue}{blue} respectively.
		\looseness=-1
	}
	\label{tab:resnet26-gn}
	\newcommand{\tabincell}[2]{\begin{tabular}{@{}#1@{}}#2\end{tabular}}
	\begin{center}
		\begin{threeparttable}
			\large
			\resizebox{1.0\linewidth}{!}{
				\begin{tabular}{l|ccc|cccc|cccc|cccc|c}
					\toprule
					\multicolumn{1}{c}{}                    & \multicolumn{3}{c}{Noise} & \multicolumn{4}{c}{Blur} & \multicolumn{4}{c}{Weather} & \multicolumn{4}{c}{Digital} & \multirow{2}{*}{Avg.}                                                                                                                                                                                                                                                                                                             \\ \cmidrule(lr){2-4} \cmidrule(lr){5-8} \cmidrule(lr){9-12} \cmidrule(lr){13-16}
					Model + Method                          & Gauss.                    & Shot                     & Impul.                      & Defoc.                      & Glass                    & Motion                   & Zoom                     & Snow                     & Frost                    & Fog                      & Brit.                    & Contr.                   & Elastic                  & Pixel                    & JPEG                     &                          \\
					\midrule
					ResNet26 (GN)                           & $56.1$                    & $51.6$                   & $51.5$                      & $21.6$                      & $42.4$                   & $20.5$                   & $25.1$                   & $20.6$                   & $24.7$                   & $19.7$                   & $11.6$                   & $13.6$                   & $25.6$                   & $48.6$                   & $30.6$                   & $30.9$                   \\
					\midrule
					~~$\bullet~$SHOT-\resetmodel            & $40.8$                    & $38.6$                   & $45.7$                      & $19.9$                      & $40.0$                   & $19.1$                   & $23.3$                   & $19.9$                   & $23.1$                   & $18.8$                   & $11.1$                   & $13.0$                   & $24.2$                   & $38.8$                   & $29.2$                   & $27.0$                   \\
					~~$\bullet~$SHOT-\notresetmodel         & $29.9$                    & $27.5$                   & $35.4$                      & $14.1$                      & $34.0$                   & \textcolor{blue}{$14.6$} & \textcolor{blue}{$15.1$} & \textcolor{blue}{$18.2$} & $18.9$                   & $16.0$                   & \textcolor{blue}{$10.4$} & $11.7$                   & \textcolor{blue}{$22.3$} & $20.0$                   & $24.6$                   & $20.8$                   \\
					\midrule
					~~$\bullet~$TTT-\resetmodel             & $38.2$                    & $\mathbf{34.7}$          & $\mathbf{40.6}$             & $\mathbf{13.8}$             & $\mathbf{37.4}$          & $16.7$                   & $\mathbf{17.8}$          & $18.4$                   & $\mathbf{19.6}$          & $16.2$                   & $10.2$                   & $11.7$                   & $22.5$                   & $\mathbf{25.0}$          & $24.8$                   & $\mathbf{23.2}$          \\
					~~$\bullet~$TTT-\notresetmodel          & \textcolor{blue}{$27.9$}  & \textcolor{blue}{$24.5$} & \textcolor{blue}{$32.6$}    & \textcolor{blue}{$13.5$}    & \textcolor{blue}{$35.7$} & $16.3$                   & $16.9$                   & $18.8$                   & \textcolor{blue}{$17.5$} & \textcolor{blue}{$14.8$} & $10.6$                   & \textcolor{blue}{$11.5$} & $23.3$                   & \textcolor{blue}{$18.0$} & \textcolor{blue}{$22.0$} & \textcolor{blue}{$20.2$} \\
					\midrule
					~~$\bullet~$TENT-\resetmodel            & $55.5$                    & $50.3$                   & $50.3$                      & $20.5$                      & $41.6$                   & $19.5$                   & $24.1$                   & $20.0$                   & $23.6$                   & $19.0$                   & $11.2$                   & $13.1$                   & $24.7$                   & $47.4$                   & $29.5$                   & $30.0$                   \\
					~~$\bullet~$TENT-\notresetmodel         & $86.4$                    & $80.9$                   & $82.1$                      & $14.7$                      & $49.8$                   & $15.1$                   & $16.4$                   & $19.7$                   & $20.9$                   & $16.7$                   & $10.5$                   & $12.5$                   & $24.3$                   & $70.6$                   & $30.6$                   & $36.8$                   \\
					\midrule
					~~$\bullet~$T3A                         & $51.3$                    & $46.6$                   & $48.3$                      & $20.5$                      & $40.4$                   & $19.6$                   & $23.5$                   & $20.5$                   & $24.1$                   & $19.2$                   & $11.7$                   & $13.6$                   & $24.5$                   & $45.9$                   & $30.1$                   & $29.3$                   \\
					\midrule
					~~$\bullet~$CoTTA-\resetmodel           & $\mathbf{38.0}$           & $36.8$                   & $41.2$                      & $22.7$                      & $42.6$                   & $21.7$                   & $27.6$                   & $20.2$                   & $21.8$                   & $19.5$                   & $10.6$                   & $12.3$                   & $27.5$                   & $49.7$                   & $30.5$                   & $28.2$                   \\
					~~$\bullet~$CoTTA-\notresetmodel        & $55.3$                    & $57.3$                   & $50.1$                      & $47.5$                      & $73.6$                   & $44.2$                   & $54.2$                   & $37.8$                   & $41.9$                   & $44.0$                   & $12.9$                   & $15.2$                   & $62.7$                   & $67.7$                   & $56.4$                   & $48.1$                   \\
					\midrule
					~~$\bullet~$MEMO-\resetmodel            & $55.7$                    & $50.3$                   & $49.7$                      & $15.8$                      & $39.0$                   & $\mathbf{15.0}$          & $18.1$                   & $\mathbf{17.1}$          & $19.7$                   & $\mathbf{15.3}$          & $\mathbf{8.4}$           & $\mathbf{11.2}$          & $\mathbf{19.3}$          & $45.5$                   & $\mathbf{22.9}$          & $26.9$                   \\
					\midrule
					~~$\bullet~$NOTE-\resetmodel            & $46.7$                    & $42.9$                   & $46.4$                      & $20.3$                      & $40.4$                   & $19.4$                   & $23.6$                   & $20.0$                   & $23.2$                   & $18.7$                   & $11.3$                   & $13.2$                   & $24.6$                   & $42.4$                   & $29.1$                   & $28.1$                   \\
					~~$\bullet~$NOTE-\notresetmodel         & $34.5$                    & $31.3$                   & $39.8$                      & $15.2$                      & $36.4$                   & $16.0$                   & $16.8$                   & $19.6$                   & $19.8$                   & $17.1$                   & $10.7$                   & $12.5$                   & $23.4$                   & $23.7$                   & $26.6$                   & $22.9$                   \\
					\midrule
					~~$\bullet~$Conjugate PL-\resetmodel    & $55.6$                    & $50.6$                   & $50.5$                      & $20.6$                      & $41.7$                   & $19.7$                   & $24.2$                   & $20.1$                   & $23.8$                   & $19.2$                   & $11.3$                   & $13.2$                   & $24.7$                   & $47.7$                   & $29.6$                   & $30.2$                   \\
					~~$\bullet~$Conjugate PL-\notresetmodel & $86.9$                    & $75.3$                   & $82.4$                      & $15.0$                      & $76.9$                   & $15.5$                   & $16.2$                   & $20.1$                   & $19.8$                   & $17.4$                   & $10.5$                   & $13.1$                   & $27.0$                   & $76.7$                   & $31.9$                   & $39.0$                   \\
					\midrule
					~~$\bullet~$SAR-\resetmodel             & $51.6$                    & $47.7$                   & $48.0$                      & $19.6$                      & $39.8$                   & $18.3$                   & $22.6$                   & $18.6$                   & $22.5$                   & $17.7$                   & $10.7$                   & $12.5$                   & $22.9$                   & $46.0$                   & $28.3$                   & $28.4$                   \\
					~~$\bullet~$SAR-\notresetmodel          & $65.5$                    & $54.5$                   & $57.3$                      & $17.6$                      & $43.1$                   & $16.2$                   & $17.0$                   & $20.3$                   & $22.0$                   & $17.4$                   & $10.8$                   & $13.3$                   & $24.2$                   & $31.8$                   & $30.6$                   & $29.4$                   \\
					\bottomrule
				\end{tabular}
			}
		\end{threeparttable}
	\end{center}
\end{table*}

\begin{table*}[!h]
	\caption{
		Results of TTA performance on \textbf{ViTSmall (LN)}. We report the \textbf{error in (\%)} on CIFAR10-C severity level 5 \textbf{under uniformly distributed test streams}.
		Optimal results in \resetmodel \& \notresetmodel are highlighted by \textbf{bold} and \textcolor{blue}{blue} respectively.
		\looseness=-1
	}
	\label{tab:vit-small-ln}
	\newcommand{\tabincell}[2]{\begin{tabular}{@{}#1@{}}#2\end{tabular}}
	\begin{center}
		\begin{threeparttable}
			\large
			\resizebox{1.0\linewidth}{!}{
				\begin{tabular}{l|ccc|cccc|cccc|cccc|c}
					\toprule
					\multicolumn{1}{c}{}                    & \multicolumn{3}{c}{Noise} & \multicolumn{4}{c}{Blur} & \multicolumn{4}{c}{Weather} & \multicolumn{4}{c}{Digital} & \multirow{2}{*}{Avg.}                                                                                                                                                                                                                                                                                                   \\ \cmidrule(lr){2-4} \cmidrule(lr){5-8} \cmidrule(lr){9-12} \cmidrule(lr){13-16}
					Model + Method                          & Gauss.                    & Shot                     & Impul.                      & Defoc.                      & Glass                    & Motion                  & Zoom                    & Snow                    & Frost                   & Fog                     & Brit.                   & Contr.                  & Elastic                 & Pixel                   & JPEG                     &                         \\
					\midrule
					ViTSmall (LN)                           & $33.3$                    & $28.5$                   & $17.7$                      & $5.8$                       & $22.1$                   & $10.5$                  & $4.9$                   & $5.3$                   & $7.7$                   & $12.4$                  & $2.9$                   & $10.0$                  & $12.6$                  & $24.4$                  & $15.6$                   & $14.2$                  \\
					\midrule
					~~$\bullet~$SHOT-\resetmodel            & $23.6$                    & $21.2$                   & $14.0$                      & $5.3$                       & $19.7$                   & $9.5$                   & $4.6$                   & $5.0$                   & $7.1$                   & $10.9$                  & $2.6$                   & $8.1$                   & $11.5$                  & $12.2$                  & $14.6$                   & $11.3$                  \\
					~~$\bullet~$SHOT-\notresetmodel         & $14.7$                    & $14.5$                   & $10.2$                      & $4.3$                       & $12.6$                   & $5.4$                   & $3.3$                   & $4.4$                   & $5.1$                   & $5.8$                   & $2.3$                   & $3.3$                   & $8.8$                   & $5.3$                   & $11.6$                   & $7.4$                   \\
					\midrule
					~~$\bullet~$TTT-\resetmodel             & $\mathbf{14.4}$           & $\mathbf{12.2}$          & $\mathbf{8.7}$              & $\mathbf{3.8}$              & $\mathbf{13.6}$          & $\mathbf{6.2}$          & $\mathbf{3.0}$          & $\mathbf{3.7}$          & $\mathbf{4.8}$          & $\mathbf{7.3}$          & $2.1$                   & $\mathbf{4.2}$          & $8.3$                   & $5.4$                   & $\mathbf{11.4}$          & $\mathbf{7.3}$          \\
					~~$\bullet~$TTT-\notresetmodel          & \textcolor{blue}{$10.8$}  & \textcolor{blue}{$9.5$}  & \textcolor{blue}{$6.6$}     & \textcolor{blue}{$3.9$}     & \textcolor{blue}{$10.3$} & \textcolor{blue}{$5.3$} & \textcolor{blue}{$3.2$} & \textcolor{blue}{$3.8$} & \textcolor{blue}{$4.0$} & \textcolor{blue}{$4.8$} & \textcolor{blue}{$2.2$} & \textcolor{blue}{$2.8$} & \textcolor{blue}{$7.8$} & \textcolor{blue}{$4.3$} & \textcolor{blue}{$10.0$} & \textcolor{blue}{$5.9$} \\
					\midrule
					~~$\bullet~$TENT-\resetmodel            & $29.8$                    & $25.3$                   & $15.8$                      & $5.5$                       & $20.1$                   & $9.6$                   & $4.6$                   & $5.1$                   & $7.3$                   & $11.6$                  & $2.8$                   & $8.9$                   & $11.6$                  & $16.9$                  & $14.8$                   & $12.6$                  \\
					~~$\bullet~$TENT-\notresetmodel         & $18.7$                    & $16.9$                   & $10.5$                      & $4.3$                       & $12.8$                   & $6.6$                   & $3.4$                   & $4.6$                   & $5.3$                   & $5.8$                   & $2.4$                   & $3.5$                   & $8.9$                   & $5.8$                   & $12.1$                   & $8.1$                   \\
					\midrule
					~~$\bullet~$T3A                         & $29.4$                    & $24.8$                   & $17.1$                      & $6.0$                       & $21.3$                   & $10.2$                  & $4.8$                   & $5.4$                   & $7.1$                   & $10.7$                  & $2.9$                   & $9.0$                   & $12.1$                  & $21.1$                  & $16.0$                   & $13.2$                  \\
					\midrule
					~~$\bullet~$CoTTA-\resetmodel           & $81.7$                    & $82.2$                   & $75.5$                      & $4.8$                       & $76.0$                   & $22.9$                  & $3.8$                   & $4.0$                   & $6.4$                   & $32.3$                  & $\mathbf{1.9}$          & $14.2$                  & $44.4$                  & $68.0$                  & $51.2$                   & $38.0$                  \\
					~~$\bullet~$CoTTA-\notresetmodel        & $88.6$                    & $88.8$                   & $87.6$                      & $5.3$                       & $87.4$                   & $39.9$                  & $4.0$                   & $4.6$                   & $5.3$                   & $48.9$                  & $2.7$                   & $11.7$                  & $67.6$                  & $82.8$                  & $78.4$                   & $46.9$                  \\
					\midrule
					~~$\bullet~$MEMO-\resetmodel            & $20.9$                    & $17.5$                   & $13.2$                      & $4.1$                       & $15.2$                   & $6.9$                   & $3.4$                   & $3.9$                   & $5.3$                   & $8.0$                   & $\mathbf{1.9}$          & $4.4$                   & $\mathbf{7.9}$          & $\mathbf{4.9}$          & $11.7$                   & $8.6$                   \\
					\midrule
					~~$\bullet~$NOTE-\resetmodel            & $31.9$                    & $27.3$                   & $17.2$                      & $5.8$                       & $21.6$                   & $10.3$                  & $4.8$                   & $5.3$                   & $7.6$                   & $12.1$                  & $2.9$                   & $9.6$                   & $12.4$                  & $21.8$                  & $15.3$                   & $13.7$                  \\
					~~$\bullet~$NOTE-\notresetmodel         & $19.0$                    & $16.4$                   & $12.3$                      & $4.7$                       & $14.7$                   & $7.4$                   & $3.9$                   & $4.8$                   & $5.9$                   & $7.7$                   & $2.6$                   & $4.9$                   & $9.6$                   & $7.2$                   & $13.0$                   & $8.9$                   \\
					\midrule
					~~$\bullet~$Conjugate PL-\resetmodel    & $30.1$                    & $24.9$                   & $15.5$                      & $5.5$                       & $20.0$                   & $9.4$                   & $4.6$                   & $5.1$                   & $7.3$                   & $11.5$                  & $2.8$                   & $8.7$                   & $11.7$                  & $15.4$                  & $14.8$                   & $12.5$                  \\
					~~$\bullet~$Conjugate PL-\notresetmodel & $19.6$                    & $18.7$                   & $10.8$                      & $4.2$                       & $12.5$                   & $6.1$                   & \textcolor{blue}{$3.2$} & $4.6$                   & $5.2$                   & $6.1$                   & $2.6$                   & $3.2$                   & $8.9$                   & $5.9$                   & $12.1$                   & $8.2$                   \\
					\midrule
					~~$\bullet~$SAR-\resetmodel             & $29.2$                    & $24.8$                   & $15.5$                      & $5.7$                       & $19.6$                   & $9.4$                   & $4.8$                   & $5.3$                   & $7.6$                   & $11.2$                  & $2.9$                   & $8.6$                   & $11.6$                  & $17.5$                  & $14.3$                   & $12.5$                  \\
					~~$\bullet~$SAR-\notresetmodel          & $20.3$                    & $18.2$                   & $11.7$                      & $4.5$                       & $13.3$                   & $6.6$                   & $3.6$                   & $4.6$                   & $5.8$                   & $6.8$                   & $2.6$                   & $4.3$                   & $9.0$                   & $6.9$                   & $12.5$                   & $8.7$                   \\
					\bottomrule
				\end{tabular}
			}
		\end{threeparttable}
	\end{center}
\end{table*}

\begin{table*}[!h]
	\caption{
		Results of TTA performance on \textbf{ResNet50-GN}. We report the \textbf{error in (\%)} on ImageNet-C severity level 5 \textbf{under uniformly distributed test streams}.
		Optimal results in \resetmodel \& \notresetmodel are highlighted by \textbf{bold} and \textcolor{blue}{blue} respectively.
		\looseness=-1
	}
	\label{tab:resnet50-gn}
	\newcommand{\tabincell}[2]{\begin{tabular}{@{}#1@{}}#2\end{tabular}}
	\begin{center}
		\begin{threeparttable}
			\large
			\resizebox{1.0\linewidth}{!}{
				\begin{tabular}{l|ccc|cccc|cccc|cccc|c}
					\toprule
					\multicolumn{1}{c}{}                    & \multicolumn{3}{c}{Noise} & \multicolumn{4}{c}{Blur} & \multicolumn{4}{c}{Weather} & \multicolumn{4}{c}{Digital} & \multirow{2}{*}{Avg.}                                                                                                                                                                                                                                                                                                             \\ \cmidrule(lr){2-4} \cmidrule(lr){5-8} \cmidrule(lr){9-12} \cmidrule(lr){13-16}
					Model + Method                          & Gauss.                    & Shot                     & Impul.                      & Defoc.                      & Glass                    & Motion                   & Zoom                     & Snow                     & Frost                    & Fog                      & Brit.                    & Contr.                   & Elastic                  & Pixel                    & JPEG                     &                          \\
					\midrule
					ResNet50 (GN)                           & $78.3$                    & $78.7$                   & $78.0$                      & $83.4$                      & $91.3$                   & $81.2$                   & $74.6$                   & $64.5$                   & $57.7$                   & $66.1$                   & $34.1$                   & $69.1$                   & $83.9$                   & $65.4$                   & $50.0$                   & $70.4$                   \\
					\midrule
					~~$\bullet~$SHOT-\resetmodel            & $\mathbf{70.6}$           & $\mathbf{69.5}$          & $\mathbf{69.4}$             & $82.3$                      & $\mathbf{84.9}$          & $\mathbf{78.3}$          & $\mathbf{72.1}$          & $\mathbf{61.2}$          & $57.7$                   & $\mathbf{58.9}$          & $\mathbf{32.9}$          & $\mathbf{66.2}$          & $\mathbf{72.2}$          & $\mathbf{56.7}$          & $\mathbf{47.8}$          & $\mathbf{65.4}$          \\
					~~$\bullet~$SHOT-\notresetmodel         & $61.3$                    & $57.9$                   & $58.9$                      & $92.1$                      & \textcolor{blue}{$86.4$} & $82.2$                   & $73.3$                   & $54.9$                   & $59.2$                   & $56.2$                   & $34.8$                   & $89.1$                   & \textcolor{blue}{$55.7$} & $41.2$                   & $43.9$                   & $63.1$                   \\
					\midrule
					~~$\bullet~$TENT-\resetmodel            & $77.6$                    & $78.0$                   & $77.2$                      & $82.8$                      & $91.0$                   & $80.9$                   & $74.3$                   & $64.2$                   & $\mathbf{57.3}$          & $65.8$                   & $33.8$                   & $68.4$                   & $83.7$                   & $64.3$                   & $49.8$                   & $69.9$                   \\
					~~$\bullet~$TENT-\notresetmodel         & $86.6$                    & $78.9$                   & $83.5$                      & $90.5$                      & $98.5$                   & $88.5$                   & $80.1$                   & $83.3$                   & $81.6$                   & $83.4$                   & $33.2$                   & $63.9$                   & $96.2$                   & $54.7$                   & $49.9$                   & $76.8$                   \\
					\midrule
					~~$\bullet~$T3A                         & $84.7$                    & $84.3$                   & $84.9$                      & $85.0$                      & $92.1$                   & $83.6$                   & $75.4$                   & $64.5$                   & $58.8$                   & $66.2$                   & $34.1$                   & $71.6$                   & $83.2$                   & $66.5$                   & $51.0$                   & $72.4$                   \\
					\midrule
					~~$\bullet~$CoTTA-\resetmodel           & $91.9$                    & $92.5$                   & $90.9$                      & $93.7$                      & $97.2$                   & $89.8$                   & $83.9$                   & $73.7$                   & $66.0$                   & $72.0$                   & $47.5$                   & $82.3$                   & $90.7$                   & $83.2$                   & $61.9$                   & $81.2$                   \\
					~~$\bullet~$CoTTA-\notresetmodel        & $98.8$                    & $99.1$                   & $98.9$                      & $99.2$                      & $99.6$                   & $98.2$                   & $94.7$                   & $98.5$                   & $96.1$                   & $92.0$                   & $69.1$                   & $93.6$                   & $99.0$                   & $98.4$                   & $81.7$                   & $94.5$                   \\
					\midrule
					~~$\bullet~$MEMO-\resetmodel            & $77.0$                    & $77.5$                   & $76.3$                      & $83.0$                      & $86.6$                   & $79.0$                   & $72.7$                   & $63.0$                   & $57.6$                   & $62.9$                   & $\mathbf{32.9}$          & $67.8$                   & $82.1$                   & $58.1$                   & $48.1$                   & $68.3$                   \\
					\midrule
					~~$\bullet~$NOTE-\resetmodel            & $78.3$                    & $78.7$                   & $78.0$                      & $83.4$                      & $91.3$                   & $81.2$                   & $74.6$                   & $64.5$                   & $57.7$                   & $66.0$                   & $34.0$                   & $69.1$                   & $83.9$                   & $65.4$                   & $50.0$                   & $70.4$                   \\
					~~$\bullet~$NOTE-\notresetmodel         & $77.3$                    & $77.0$                   & $76.6$                      & \textcolor{blue}{$83.3$}    & $90.5$                   & $80.3$                   & $74.0$                   & $64.2$                   & $57.7$                   & $64.8$                   & $33.9$                   & $67.7$                   & $82.8$                   & $62.5$                   & $49.7$                   & $69.5$                   \\
					\midrule
					~~$\bullet~$Conjugate PL-\resetmodel    & $76.2$                    & $75.8$                   & $75.5$                      & $\mathbf{82.0}$             & $90.4$                   & $79.9$                   & $73.0$                   & $63.4$                   & $\mathbf{57.3}$          & $66.0$                   & $33.0$                   & $\mathbf{66.2}$          & $82.9$                   & $60.3$                   & $48.6$                   & $68.7$                   \\
					~~$\bullet~$Conjugate PL-\notresetmodel & $93.3$                    & $87.0$                   & $91.5$                      & $97.4$                      & $99.3$                   & $96.7$                   & $89.8$                   & $96.5$                   & $94.6$                   & $98.3$                   & $29.2$                   & $61.0$                   & $99.0$                   & $40.3$                   & $43.7$                   & $81.2$                   \\
					\midrule
					~~$\bullet~$SAR-\resetmodel             & $77.1$                    & $77.3$                   & $76.6$                      & $82.4$                      & $90.6$                   & $80.3$                   & $73.9$                   & $64.0$                   & $57.4$                   & $65.1$                   & $33.6$                   & $67.8$                   & $83.4$                   & $63.4$                   & $49.7$                   & $69.5$                   \\
					~~$\bullet~$SAR-\notresetmodel          & \textcolor{blue}{$60.1$}  & \textcolor{blue}{$57.1$} & \textcolor{blue}{$58.5$}    & $83.9$                      & $92.2$                   & \textcolor{blue}{$57.8$} & \textcolor{blue}{$55.3$} & \textcolor{blue}{$54.1$} & \textcolor{blue}{$55.7$} & \textcolor{blue}{$41.7$} & \textcolor{blue}{$28.8$} & \textcolor{blue}{$49.9$} & $94.0$                   & \textcolor{blue}{$38.8$} & \textcolor{blue}{$42.4$} & \textcolor{blue}{$58.0$} \\
					\bottomrule
				\end{tabular}
			}
		\end{threeparttable}
	\end{center}
\end{table*}

\begin{table*}[!h]
	\caption{
		Results of TTA performance on \textbf{ViTBase (LN)}. We report the \textbf{error in (\%)} on ImageNet-C severity level 5 \textbf{under uniformly distributed test streams}.
		Optimal results in \resetmodel \& \notresetmodel are highlighted by \textbf{bold} and \textcolor{blue}{blue} respectively.
		\looseness=-1
	}
	\label{tab:vit-base-ln}
	\newcommand{\tabincell}[2]{\begin{tabular}{@{}#1@{}}#2\end{tabular}}
	\begin{center}
		\begin{threeparttable}
			\large
			\resizebox{1.0\linewidth}{!}{
				\begin{tabular}{l|ccc|cccc|cccc|cccc|c}
					\toprule
					\multicolumn{1}{c}{}                    & \multicolumn{3}{c}{Noise} & \multicolumn{4}{c}{Blur} & \multicolumn{4}{c}{Weather} & \multicolumn{4}{c}{Digital} & \multirow{2}{*}{Avg.}                                                                                                                                                                                                                                                                                                             \\ \cmidrule(lr){2-4} \cmidrule(lr){5-8} \cmidrule(lr){9-12} \cmidrule(lr){13-16}
					Model + Method                          & Gauss.                    & Shot                     & Impul.                      & Defoc.                      & Glass                    & Motion                   & Zoom                     & Snow                     & Frost                    & Fog                      & Brit.                    & Contr.                   & Elastic                  & Pixel                    & JPEG                     &                          \\
					\midrule
					ViTBase (LN)                            & $74.1$                    & $78.2$                   & $75.4$                      & $70.1$                      & $78.6$                   & $67.5$                   & $73.1$                   & $84.2$                   & $75.3$                   & $52.8$                   & $46.4$                   & $56.8$                   & $70.3$                   & $52.1$                   & $48.8$                   & $66.9$                   \\
					\midrule
					~~$\bullet~$SHOT-\resetmodel            & $\mathbf{56.7}$           & $\mathbf{56.7}$          & $\mathbf{56.0}$             & $\mathbf{52.6}$             & $\mathbf{59.7}$          & $\mathbf{50.1}$          & $\mathbf{52.4}$          & $\mathbf{43.3}$          & $\mathbf{45.3}$          & $\mathbf{39.4}$          & $26.3$                   & $\mathbf{41.6}$          & $\mathbf{50.8}$          & $35.0$                   & $39.9$                   & $\mathbf{47.0}$          \\
					~~$\bullet~$SHOT-\notresetmodel         & $73.2$                    & $60.5$                   & $59.5$                      & $63.9$                      & $57.5$                   & $49.2$                   & $42.2$                   & $42.9$                   & \textcolor{blue}{$46.6$} & $34.0$                   & $24.8$                   & $60.8$                   & $34.4$                   & $29.1$                   & $34.7$                   & $47.6$                   \\
					\midrule
					~~$\bullet~$TENT-\resetmodel            & $73.4$                    & $77.3$                   & $74.7$                      & $69.0$                      & $78.0$                   & $66.7$                   & $72.3$                   & $83.4$                   & $74.2$                   & $52.1$                   & $45.1$                   & $55.8$                   & $69.7$                   & $51.1$                   & $48.3$                   & $66.1$                   \\
					~~$\bullet~$TENT-\notresetmodel         & $50.5$                    & $50.1$                   & $51.9$                      & $44.8$                      & $45.5$                   & $39.4$                   & $46.8$                   & $52.4$                   & $72.7$                   & $28.7$                   & $23.0$                   & $35.2$                   & $50.1$                   & $27.3$                   & $31.3$                   & $43.3$                   \\
					\midrule
					~~$\bullet~$T3A                         & $74.7$                    & $78.9$                   & $75.8$                      & $70.5$                      & $78.9$                   & $67.6$                   & $72.7$                   & $84.6$                   & $75.5$                   & $51.8$                   & $46.0$                   & $57.4$                   & $68.8$                   & $52.6$                   & $48.8$                   & $67.0$                   \\
					\midrule
					~~$\bullet~$CoTTA-\resetmodel           & $98.6$                    & $98.6$                   & $99.1$                      & $95.5$                      & $97.8$                   & $92.8$                   & $88.1$                   & $86.9$                   & $97.3$                   & $92.6$                   & $55.1$                   & $95.6$                   & $98.1$                   & $89.4$                   & $64.6$                   & $90.0$                   \\
					~~$\bullet~$CoTTA-\notresetmodel        & $99.4$                    & $99.5$                   & $99.5$                      & $99.6$                      & $99.7$                   & $99.4$                   & $99.3$                   & $99.2$                   & $99.3$                   & $99.3$                   & $96.3$                   & $99.5$                   & $99.5$                   & $99.0$                   & $92.5$                   & $98.7$                   \\
					\midrule
					~~$\bullet~$MEMO-\resetmodel            & $68.8$                    & $74.3$                   & $70.4$                      & $60.1$                      & $66.6$                   & $55.7$                   & $57.0$                   & $54.3$                   & $58.4$                   & $45.4$                   & $\mathbf{23.9}$          & $42.8$                   & $65.6$                   & $\mathbf{33.4}$          & $\mathbf{36.9}$          & $54.2$                   \\
					\midrule
					~~$\bullet~$NOTE-\resetmodel            & $74.1$                    & $78.1$                   & $75.4$                      & $70.0$                      & $78.6$                   & $67.5$                   & $73.1$                   & $84.2$                   & $75.3$                   & $52.7$                   & $46.4$                   & $56.8$                   & $70.3$                   & $52.1$                   & $48.8$                   & $66.9$                   \\
					~~$\bullet~$NOTE-\notresetmodel         & $72.0$                    & $75.4$                   & $73.2$                      & $67.2$                      & $76.7$                   & $64.9$                   & $70.7$                   & $79.4$                   & $71.4$                   & $50.9$                   & $41.9$                   & $54.3$                   & $68.5$                   & $49.2$                   & $47.7$                   & $64.2$                   \\
					\midrule
					~~$\bullet~$Conjugate PL-\resetmodel    & $69.1$                    & $72.8$                   & $70.2$                      & $65.1$                      & $74.3$                   & $63.3$                   & $68.7$                   & $80.7$                   & $71.7$                   & $48.7$                   & $41.4$                   & $50.4$                   & $67.3$                   & $46.4$                   & $45.7$                   & $62.4$                   \\
					~~$\bullet~$Conjugate PL-\notresetmodel & $80.7$                    & $75.5$                   & $84.0$                      & $45.3$                      & $48.3$                   & $40.3$                   & $68.9$                   & $91.4$                   & $96.0$                   & $29.3$                   & $23.7$                   & $35.1$                   & $96.5$                   & $27.7$                   & $31.9$                   & $58.3$                   \\
					\midrule
					~~$\bullet~$SAR-\resetmodel             & $73.6$                    & $77.6$                   & $75.0$                      & $69.4$                      & $78.3$                   & $67.1$                   & $72.7$                   & $83.6$                   & $74.7$                   & $52.4$                   & $45.8$                   & $56.3$                   & $70.0$                   & $51.7$                   & $48.6$                   & $66.5$                   \\
					~~$\bullet~$SAR-\notresetmodel          & \textcolor{blue}{$46.1$}  & \textcolor{blue}{$47.5$} & \textcolor{blue}{$44.5$}    & \textcolor{blue}{$43.5$}    & \textcolor{blue}{$43.8$} & \textcolor{blue}{$38.1$} & \textcolor{blue}{$39.9$} & \textcolor{blue}{$33.2$} & $54.2$                   & \textcolor{blue}{$28.1$} & \textcolor{blue}{$22.9$} & \textcolor{blue}{$34.7$} & \textcolor{blue}{$32.7$} & \textcolor{blue}{$27.1$} & \textcolor{blue}{$30.8$} & \textcolor{blue}{$37.8$} \\
					\bottomrule
				\end{tabular}
			}
		\end{threeparttable}
	\end{center}
\end{table*}

\section{Additional Related Work} \label{appendix:additional_related}

\paragraph{Unsupervised Domain Adaptation}
Unsupervised Domain Adaptation (UDA) is a technique aimed at enhancing the performance of a target model in scenarios where there is a shift in distribution between the labeled source domain and the unlabeled target domain.
UDA methods typically seek to align the feature distributions between the two domains through the utilization of discrepancy losses~\citep{long2015learning} or adversarial training~\citep{ganin2015unsupervised, tsai2018learning}.

\paragraph{Domain Generalization}
Our work is also related to DG~\citep{muandet2013domain, blanchard2011generalizing} in a broad sense, due to the shared goal of bridging the gap of distribution shifts between the source domain and the target domain.
Also, DG and TTA may share similar constraints on model selection for lacking label information in the target domain.
DomainBed~\cite{gulrajani2021in} highlights the necessity of considering model selection criterion in DG and concludes that ERM~\citep{vapnik1998statistical} outperforms the state-of-the-art in terms of average
performance after carefully tuning using model selection criteria. \looseness=-1

\paragraph{Distribution Shift Benchmarks.}
Distribution shift has been widely studied in the machine learning community.
Prior works have covered a wide range of distribution shifts.
The first line of such benchmarks applies different transformations to object recognition datasets to induce distribution shifts.
These benchmarks include: (1) CIFAR10-C \& ImageNet-C~\citep{hendrycks2018benchmarking}, ImageNet-A~\citep{hendrycks2021nae}, ImageNet-R~\citep{hendrycks2021many}, ImageNet-V2~\citep{recht2019imagenet}, and many others;
(2) ColoredMNIST~\citep{arjovsky2019invariant}, which makes the color of digits a confounder.
Most recent benchmarks collect sets of images with various styles and backgrounds, such as PACS~\citep{li2017deeper}, OfficeHome~\citep{venkateswara2017deep}, DomainNet~\citep{peng2019moment}, and Waterbirds~\citep{sagawa2019distributionally}.
Unlike most prior works that assume a specific stationary target domain, the study on continuous TTA that considers continually changing target data becomes more and more popular in the field.
Recently, a few works have constructed datasets and benchmarks for scenarios under temporal shifts.
\citet{gong2022robust} builds a temporally correlated test stream on CIFAR10-C sample by a Dirichlet distribution, where most existing TTA methods fail dramatically.
Wild-Time~\citep{yao2022wild} benchmark consists of 5 datasets that reflect temporal distribution shifts arising in a variety of real-world applications, including patient prognosis and news classification.
Studies on fairness and bias~\citep{mehrabi2021survey} have investigated the detrimental impact of spurious correlation in classification~\citep{geirhos2018imagenet} and conservation~\citep{beery2020synthetic}.
To our knowledge, there have been rare TTA work focused on tackling spurious correlation shifts.

\end{document}